%% file: forl_neurips_camera_ready.tex
\documentclass{article}
\PassOptionsToPackage{numbers,sort&compress}{natbib}
\usepackage[final]{neurips_2025}
\usepackage[utf8]{inputenc} 
\usepackage[T1]{fontenc}    
\usepackage{hyperref}      
\usepackage{url}           
\usepackage{booktabs}       
\usepackage{amsfonts}       
\usepackage{nicefrac}       
\usepackage{microtype}      
\usepackage{xcolor}         
\usepackage{siunitx}
\usepackage{longtable}
\usepackage{multirow}
\usepackage{tablefootnote}
\usepackage{natbib}
\usepackage{graphicx}
\usepackage{pgfplots}
\pgfplotsset{compat=1.17}
\usepgfplotslibrary{groupplots}
\usepackage{pgfplotstable}
\usetikzlibrary{patterns}
\usepackage{adjustbox}
\usepackage{algorithm}

\usepackage{algorithmic}
\input{header.tex}
\usepackage{amsmath}
\usepackage{amssymb}
\usepackage{mathtools}
\usepackage{amsthm}
\usepackage{wrapfig}
\usepackage{tcolorbox}
\usepackage{tikz}
\usetikzlibrary{shapes.arrows, arrows.meta}
\newcommand*{\FancyUpArrow}{\begin{tikzpicture}[baseline=-0.25em]
\node[single arrow,draw,rotate=90,single arrow head extend=0.2em,inner
ysep=0.1em,transform shape,line width=0.02em,color=ourred,top color=ourred!65!ourred,bottom color=ourred] (X){};
\end{tikzpicture}}
\newcommand*{\FancyDownArrow}{\begin{tikzpicture}[baseline=-0.25em]
\node[single arrow,draw,rotate=270,single arrow head extend=0.2em,inner
ysep=0.1em,transform shape,line width=0.02em,color=ourred,top color=ourred!65!ourred,bottom color=ourred] (X){};
\end{tikzpicture}}

\newcommand{\fmt}[2]{%
  \makebox[2em][r]{#1}\,%
 \ifthenelse{\equal{#2}{}}{\phantom{\scriptsize $\pm$}\makebox[.5em][l]{}}
  {\scriptsize $\pm$ \makebox[.5em][l]{#2}}}

\usepackage[capitalize]{cleveref}

\theoremstyle{plain}
\newtheorem{theorem}{Theorem}[section]

\theoremstyle{definition}
\newtheorem{definition}[theorem]{Definition}

\theoremstyle{remark}

\usepackage{threeparttable}

\usepackage[textsize=tiny]{todonotes}

\usepackage{subcaption}

\title{Forecasting in Offline Reinforcement Learning for Non-stationary Environments}

\author{%
  Suzan Ece Ada$^{1,2}$ \quad
  Georg Martius$^{2}$ \quad
  Emre Ugur$^{1}$ \quad
  Erhan Oztop$^{3,4}$ \\
  \vspace{1ex} 
  $^{1}$Bogazici University, Türkiye \quad
  $^{2}$University of Tübingen, Germany \\
  $^{3}$Ozyegin University, Türkiye \quad
  $^{4}$Osaka University, Japan \\
  \texttt{ece.ada@bogazici.edu.tr}
}

\begin{document}

\maketitle

\begin{abstract}
Offline Reinforcement Learning (RL) provides a promising avenue for training policies from pre-collected datasets when gathering additional interaction data is infeasible. However, existing offline RL methods often assume stationarity or only consider synthetic perturbations at test time, assumptions that often fail in real-world scenarios characterized by abrupt, time-varying offsets. These offsets can lead to partial observability, causing agents to misperceive their true state and degrade performance. To overcome this challenge, we introduce \textbf{F}orecasting in Non-stationary \textbf{O}ffline \textbf{RL} (\forl), a framework that unifies (i) conditional diffusion-based candidate state generation, trained without presupposing any specific pattern of future non-stationarity, and (ii) zero-shot time-series foundation models. \forl targets environments prone to unexpected, potentially non-Markovian offsets, requiring robust agent performance from the onset of each episode. Empirical evaluations on offline RL benchmarks, augmented with real-world time-series data to simulate realistic non-stationarity, demonstrate that \forl consistently improves performance compared to competitive baselines. By integrating zero-shot forecasting with the agent's experience, we aim to bridge the gap between offline RL and the complexities of real-world, non-stationary environments.
\end{abstract}
\begin{wrapfigure}[18]{r}{0.45\linewidth}
    \centering
    \vspace{-2.65em}
    \includegraphics[width=0.9\linewidth]{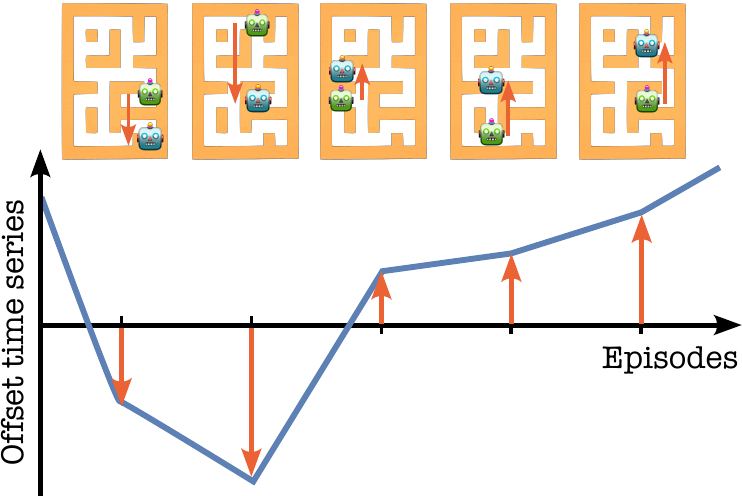}
 \caption{\label{fig:maze_offset_plots_ts}
 \textbf{Setting.} The agent does not know its location in the environment because its perception is offset every episode $j$ by an unknown offset $b^j$ (only vertical offsets are illustrated). \forl leverages historical offset data and offline RL data (from a stationary phase) to forecast and correct for new offsets at test time. Ground-truth offsets (\protect\FancyDownArrow,\protect\FancyUpArrow) are hidden throughout the evaluation episodes.} 
\end{wrapfigure}
\section{Introduction}
Offline Reinforcement Learning (RL) leverages static datasets to avoid costly or risky online interactions \citep{levineofflinerltutorial_20202,fujimoto2021a}. Yet, agents trained on fully observable states often fail when deployed with noisy or corrupted observations. While robust offline RL methods address test-time perturbations, such as sensor noise or adversarial attacks \citep{zhihedmbp}, a critical gap persists in addressing non-stationarity within the observation function—a challenge that fundamentally alters the agent's perception of the environment over time.

Prior \textit{online algorithms} that consider the scope of non-stationarity as the observation function focus on learning agent morphologies \cite{trabucco2022anymorph} and generalization in Block MDPs \cite{zhang2020invariant}. While this scope of non-stationarity holds significant potential for real-world applications \citep{khetarpal2022towards}, it remains underexplored. We focus on the episodic evolution of the observation function at test-time in offline RL. In our setup, each dimension of an agent's state is influenced by an unknown, time-dependent constant additive offset that remains fixed within a single operational interval (an ``episode''). This leads to a stream of evolving observation functions \cite{chandak2022reinforcement}, extending across multiple future episodes, \textbf{where the offsets remain hidden throughout the prediction window}. For instance, industrial robots might apply a daily calibration offset to each joint, while sensors can exhibit a deviation until the next scheduled recalibration. Similarly, in healthcare or finance, data may be partially aggregated or withheld to comply with regulations, effectively obscuring finer-grained variations and leaving a single offset as the dominant factor per episode. By only storing these representative offsets, we circumvent the challenges of continuous interaction buffers in bandwidth-constrained or privacy-sensitive environments. Because the offset can differ across state dimensions (e.g., different sensor or actuation channels), each state dimension can be affected by a different unknown bias that stays constant for that episode but evolves differently across episodes. Approaches that assume predefined perturbations can struggle with these abrupt, episodic shifts because such offsets violate the typical assumption of smoothly varying observation functions. Frequent retraining, hyperparameter optimization, or extensive online adaptation to new observation function evolution patterns are costly, risky (due to trial-and-error in safety-critical settings), and may be infeasible if these patterns no longer reflect assumptions made during training. By separating offset data (episodic calibration values) from the massive replay buffers, a zero-shot forecasting-based approach can anticipate each new offset from the beginning of the episode without requiring policy updates or making assumptions on task evolution at test time \cite{xie2021deep}. Modeling these multidimensional additive offsets as stable, per-episode constants presents a robust and efficient way to handle time-varying conditions in non-stationary environments where the evolution of tasks follows a non-Markovian, time-series pattern \cite{lee2023tempo}, mitigating the risks of online exploration. 

\begin{wrapfigure}[23]{r}{0.5\linewidth}
\centering
    \vspace{-\baselineskip}
    \includegraphics[width=.45\columnwidth]{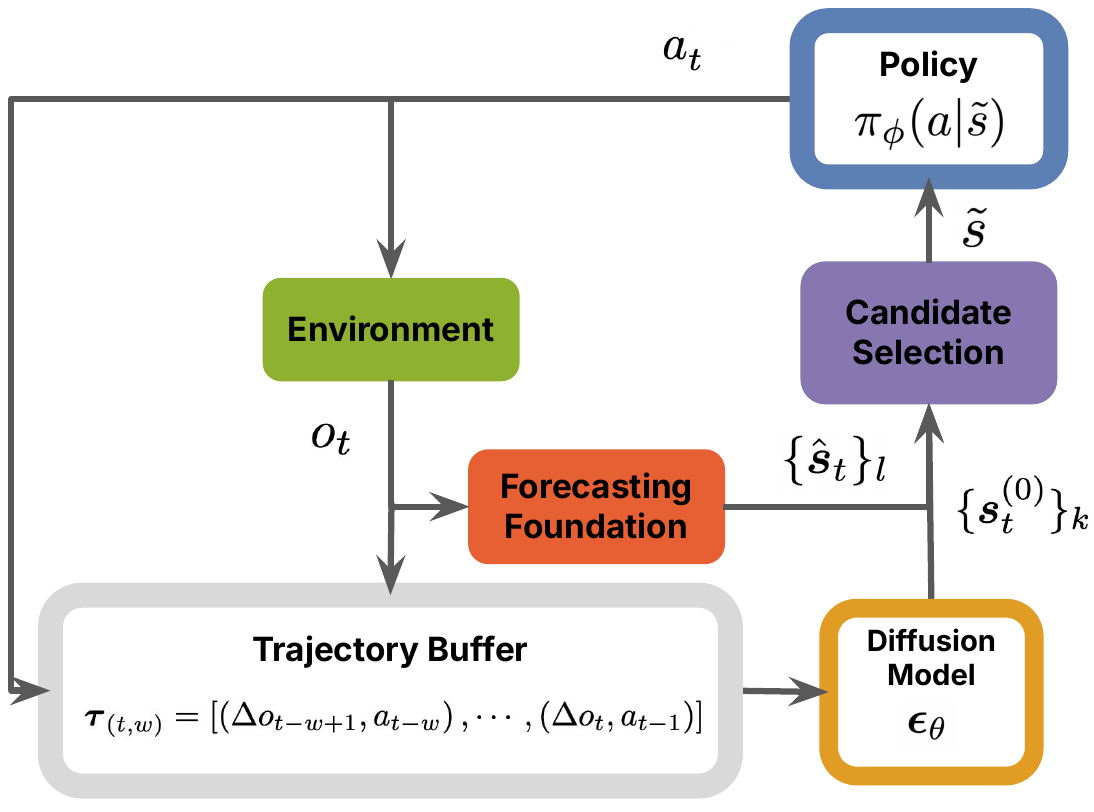}
 \caption{Overview of the proposed \forl framework at test-time. The observations are processed by both the trajectory buffer and the time-series \textcolor{ourred}{\bf forecasting foundation} module \cite{rasul2023lag}. Observation changes and actions sampled from the \textcolor{ourblue}{\bf policy} $(\Delta o, a)$, are stored in the trajectory buffer. The \textcolor{ourorange}{\bf diffusion model} generates candidate states $\{\s_{t}^{(0)}\}_k$ conditioned on $\boldsymbol{\tau}_{(t,w)}$. The \textcolor{ourviolet}{\bf candidate selection} module then generates the estimated $\tilde{s}_{t}$. }
\label{fig:porlframwork}
\end{wrapfigure}
We consider an offline RL setting during training where we have only access to a standard offline RL dataset collected from a stationary environment \cite{zhihedmbp} with fully observable states. Initial data may be collected under near-ideal conditions and then gradually affected by wear, tear, or other natural shifts—even as the underlying physical laws (dynamics) remain unchanged. At test time, however, we evaluate in a non-stationary environment where both the observation function and the observation space change due to time-dependent external factors. This setup can be interpreted as environments shifting along observation space dimensions while the initial state of the agent is sampled from a uniform distribution over the state space. A simplified version of this setup for an offset affecting only one dimension of the state is illustrated in Figure~\ref{fig:maze_offset_plots_ts}. Here, the agent “knows” it is in a maze but does not know where it is in the maze. Furthermore, \textbf{it will remain uncertain of its location across all episodes at test-time}, as in every episode, a new offset leads to a systematic misalignment between perceived and actual positions. Importantly, these offsets may not conform to Gaussian or Markovian assumptions; instead, they may stem directly from complex, real-world time-series data \cite{lee2023tempo} and remain constant throughout each episode. As a result, standard noise-driven or parametric state-estimation techniques, which typically rely on smoothly varying or randomly perturbed functions, cannot reliably identify these persistent, episode-wide offsets that are not available after the episode terminates. While zero-shot forecasting can adjust observation offsets, its performance depends on the forecaster's accuracy. Similarly, integrating zero-shot forecasting into a model-based offline RL approach \cite{zhihedmbp} can underperform when real-world offsets deviate from predefined assumptions about future observation functions. Our approach uses the insight that the belief of the true states can be refined from a sequence of actions and effects. For instance, in maze navigation, if an agent misjudges its location and hits a wall, analyzing its actions and delta observations leading to the collision can provide evidence for likely locations within the maze.

We propose the \textbf{F}orecasting in Non-stationary \textbf{O}ffline \textbf{RL} (\forl) framework (Figure~\ref{fig:porlframwork}) for test-time adaptation in non-stationary environments where the observation function is perturbed by an arbitrary time-series.
Our framework has two main ingredients: 
forecasting offsets with a zero-shot time-series forecasting model \cite{rasul2023lag} from past episode offsets (ground truth offsets after the episode terminates are not accessible at test-time) and
a within-episode update of the state estimation using a conditional diffusion model \cite{ho2020ddpm_neurips} trained on offline stationary data. 

\noindent\textbf{Contributions.} We unify the strengths of probabilistic forecasting and decision-making under uncertainty to enable continuous adaptation when the environment diverges from predictions. Consequently, our framework: \emph{(1)}~accommodates future offsets \textit{without assuming specific non-stationarity patterns during training}, eliminating the need for retraining and hyperparameter tuning when the agent encounters new, unseen non-stationary patterns at test time, and \emph{(2)}~\textit{targets non-trivial non-stationarities at test time without requiring environment interaction or knowledge of POMDPs during training}. \emph{(3)}~\forl introduces a novel, modular framework combining a conditional diffusion model (\forl-\dm) for multimodal belief generation with a lightweight Dimension-wise Closest Match (\candidsel) fusion strategy, validated by extensive experiments on no-access to past offsets, policy-agnostic plug-and-play, offset magnitude-scaling, and inter/intra-episode drifts. \emph{(4)} We propose a novel benchmark that integrates offsets from real-world time-series datasets with standard offline RL benchmarks and demonstrate that \forl consistently outperforms baseline methods.

\textbf{Background: Diffusion Models}
Denoising diffusion models \cite{DBLP:journals/corr/Sohl-DicksteinW15,luo2022understanding} aim to model the data distribution with $p_\theta(\boldsymbol{x}_0):=\int p_\theta(\boldsymbol{x}_{0:N}) \, d\boldsymbol{x}_{1:N}$ from samples $x_0$ in the dataset. The joint distribution follows the Markov Chain
$p_\theta\left(\boldsymbol{x}_{0:N}\right):=\mathcal{N}(\boldsymbol{x}_N;\mathbf{0},\mathbf{I}) \prod_{n=1}^N p_\theta\left(\boldsymbol{x}_{n-1} \mid \boldsymbol{x}_n\right) \quad$ where $\boldsymbol{x}_n$ is the noisy sample for diffusion timestep $n$ and $p_\theta\left(\boldsymbol{x}_{n-1} \mid \boldsymbol{x}_n\right):=\mathcal{N}\left(\boldsymbol{x}_{n-1} ; \boldsymbol{\mu}_\theta\left(\boldsymbol{x}_n, n\right), \mathbf{\Sigma}_\theta\left(\boldsymbol{x}_n, n\right)\right)$. During training, we use the samples from the distribution $q\left(\boldsymbol{x}_n \mid \boldsymbol{x}_0\right)=\mathcal{N}\left(\boldsymbol{x}_n ; \sqrt{\bar{\alpha}_n} \boldsymbol{x}_0,\left(1-\bar{\alpha}_n\right) \mathbf{I}\right)$ where $\bar{\alpha}_n=\prod_{i=1}^n \alpha_i$ \cite{ho2020ddpm_neurips}. General information on diffusion models is given in \cref{app:background:diffusion}.

\section{Method}
In this section, we formulate our problem statement and describe our \forl diffusion model trained on the offline RL dataset to predict plausible states. Then, we introduce our online state estimation method,
\candidselmethodname that uses plausible states predicted by the multimodal \forl diffusion model (\dm) and the states predicted from past episodes prior to evaluation by a probabilistic unimodal zero-shot time-series foundation model.

\subsection{Problem Statement}
\paragraph{Training (Offline Stationary MDP)}
We begin with an episodic, stationary Markov Decision Process (MDP) $\mathcal{M}_{\text{train}} = (\mathcal{S}, \mathcal{A}, \mathcal{T}, \mathcal{R}, \rho_{0})$, where the initial state distribution $\rho_{0}$ is a uniform distribution over the state space $\mathcal{S}$. We only have access to an offline RL dataset $\mathcal{D}=\{(\s^k_t,\boldsymbol{a}^k_t,\s^k_{t+1},r^k_t)\}$ with $k$ transitions collected from this MDP. Crucially, our \forl diffusion model and a diffusion policy \cite{wang2023diffusion} are trained offline using this dataset, such as the standard D4RL benchmark \cite{fu2020d4rl}, without making any assumptions about how the environment might become non‐stationary at test time. 

\paragraph{Test Environment (Sequence of POMDPs)}
At test time, the agent faces an infinite sequence of POMDPs $\{\hat{\mathcal{M}}_j\}_{j=1}^\infty$. Each POMDP is described by a 7-tuple \cite{kaelbling1998planning} $\hat{\mathcal{M}}_j = \bigl(\mathcal{S}, \mathcal{A}, \mathcal{O}_j, \mathcal{T}, \mathcal{R}, \rho_{0}, \x_j\bigr)$, where the state space \(\mathcal{S}\), action space \(\mathcal{A}\), transition function \(\mathcal{T}\), and the reward function \(\mathcal{R}\) remain identical to the training MDP. \(\x\) is the observation function, where we restrict ourselves to deterministic versions (\(\x: \mathcal{S} \rightarrow \mathcal{O}\)) \cite{bonet2012deterministic,khetarpal2022towards}.
Non-stationary environments can be formulated in different ways.
In \citet{khetarpal2022towards}, a \emph{general non-stationary RL} formulation is put forward,
which allows each component of the underlying MDP or POMDP to evolve over time, \ie $\bigl( \mathcal{S}(t), \mathcal{A}(t), \mathcal{T}(t), \mathcal{R}(t), \x(t), \mathcal{O}(t) \bigr)$.
A set \(\kappa\) specifies which of these components vary, and a \emph{driver} determines how they evolve. In particular, \emph{passive} drivers of non-stationarity imply that exogenous factors alone govern the evolution of the environment, independent of the agent's actions.
In this work, we consider the scope of non-stationarity \cite{khetarpal2022towards} (\cref{app:background:nonstationary}) to be the observation function and the observation space, \ie
$\kappa = \{\x,\mathcal{O}\}$.

We consider the case where non-stationarity unfolds over episodes and where the observation function \(\x_j\) is different in each episode $j$.
The change in the observation function is assumed to have an additive structure and is independent of the agent's actions (passive non‐stationarity \cite{khetarpal2022towards}).
Concretely, the function \(\x_j\) offsets states \(s_t\) by a fixed offset \(b^j\in\mathbb{R}^n\):
   \[
   \mathcal{O}_j
   \;=\;
   \{\,s + b^j :\, s\in\mathcal{S}\},
   \quad
   \x_j(s)
   \;=\;
   s + b^j.
   \]
Importantly, the sequence $(b^j)$ can evolve under arbitrary real‐world time‐series data, and the agent \textbf{does not have access to the ground-truth information throughout the evaluation}—similar to scenarios where observations are only available periodically and shifts occur between these intervals. Thus, the episodes have a temporal order, relating to Non-Stationary Decision Processes (NSDP), defining a sequence of POMDPs \cite{chandak2022reinforcement} (\cref{app:background:nonstationary}).

\paragraph{Partial Observability and Historical Context}
Since $b^j$ is \textbf{never directly observed} for $P$ episodes into the future, each $\hat{\mathcal{M}}_j$ is a POMDP. The agent receives only the offset‐shifted observations $(o_t)$, where $o_t = s_t + b^j$ without any noise. Moreover, the agent may have access to a limited historical context of previous offsets $(b^{j-C}, \dots, b^{j-1})$ at discrete intervals $P$, but \textbf{no direct information} about future offsets $(b^j,b^{j+1},\dots b^{j+P-1})$.
Hence, the agent must forecast and/or adapt to unknown future offsets without prior non‐stationary training.

\paragraph{Partial Identifiability}
Despite observing $o_t = s_t + b^j$, the agent cannot generally disentangle $s_t$ from $b^j$.
For any single observation, there are infinite possible pairs of state and offset that yield $o_t = s' + b'$. Additionally, the initial state distribution $\rho$ is uniform and does not provide information about $b$.
Thus, we can only form a belief over $s_t$ and refine that belief based on two sources of information: \textsf{a)} the sequence of actions and effects observed within an episode and \textsf{b)} the sequence of past identified offsets.
To exploit source \textsf{a}, we utilize a predictive model of commonly expected outcomes based on a diffusion model, which will be explained next. To make use of source \textsf{b}, we use a zero-shot forecasting model (see \cref{sec:lagllama} for details). Afterwards, both pieces of information are fused (\cref{sec:fusion}). 
\begin{wrapfigure}[14]{r}{0.4\linewidth}
\includegraphics[width=\linewidth]{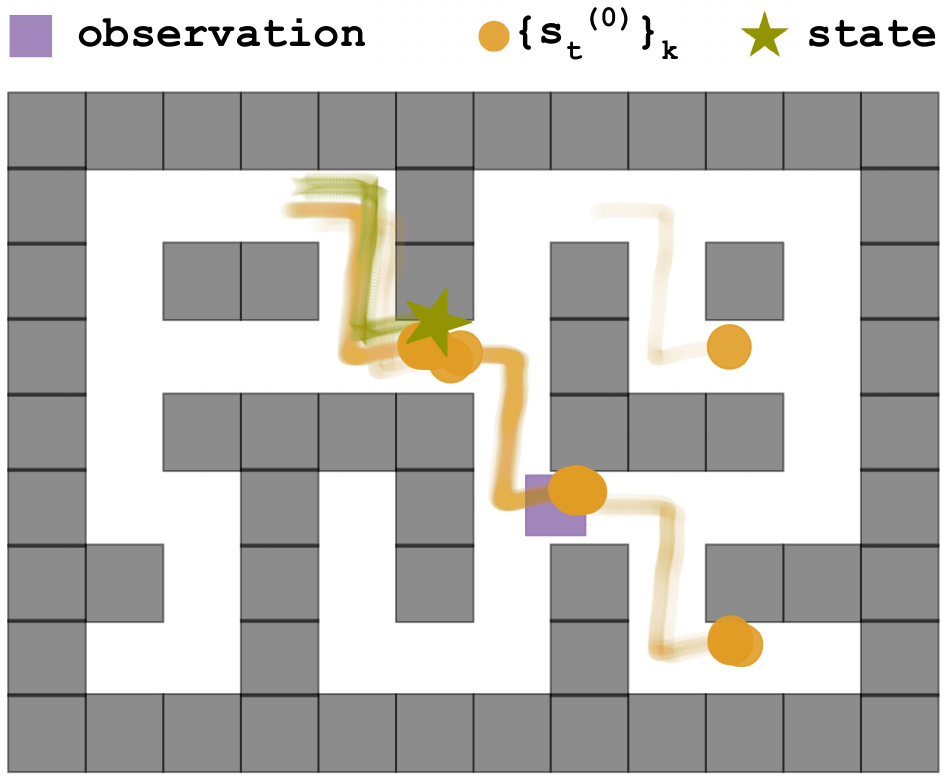}
\caption{Candidate states generated by \forl Diffusion Model.}\label{fig:hypomaze}
\end{wrapfigure}
\subsection{\forl Diffusion Model}
\label{subsec:porldm}

In our setting, we assume that the offsets added to the states are unobservable at test time, while the transition dynamics of the evaluation environment remain unchanged.
To eventually reduce the uncertainty about the underlying state,
we perform filtering or belief updates using the sequence of past interactions.
To understand why the history of interactions is indicative of a particular ground truth state, consider the following example in a maze environment illustrated in \cref{fig:hypomaze}. When the agent moves north for three steps and then bumps into a wall, the possible ground truth states can only be those three steps south of any wall. The agent cannot observe the hidden green trajectory of ground-truth states; it only has access to the sequence of observation changes $(\Delta o)$ and action vectors, which narrows down the possible positions to four candidate regions—exactly those identified by our model.
Clearly, the distribution of possible states is highly multimodal,
such that we propose using a diffusion model as a flexible predictive model of plausible states given the observed actions and outcomes. 
Diffusion models excel at capturing multimodal distributions \cite{wang2023diffusion}, making them well-suited for our task.
We train the diffusion model on offline data without offsets ($b_j=0$) which we detail below.

To distinguish between the trajectory timesteps in reinforcement learning (RL) and the timesteps in the diffusion process, we use the subscript $t\in \{0,\dots,T\}$ to refer to RL timesteps and $n\in \{0,\dots,N\}$ for diffusion timesteps. We first begin by defining a subsequence of a trajectory
\begin{equation}
\boldsymbol{\tau}_{(t,w)} = \left[ \left(\Delta o_{t-w+1}, a_{t-w}\right), \cdots, \left(\Delta o_{t}, a_{t-1}\right) \right].
\label{eq:trajeq}
\end{equation}
where delta observations $\Delta o_{t}=o_{t}-o_{t-1} = s_{t}-s_{t-1}$ denote the state changes (effects), $w$ is the window size. Using a conditional diffusion model, we aim to model the distribution $p\left(\s_t \mid \boldsymbol{\tau}_{(t,w)}\right)$. For that, we define the reverse process (denoising) as
\begin{equation}
p\big(s^{(N)}_t\big) \prod_{n=1}^{N} p_\theta\left(\s_t^{(n-1)} \mid \s_t^{(n)},\boldsymbol{\tau}_{(t,w)} \right), \quad p\big(s^{(N)}_t\big)=\mathcal{N}(\s^{(N)}_t;\mathbf{0},\mathbf{I})
\end{equation}
and $p_\theta$ is modeled as the distribution $\mathcal{N}(\s^{(n-1)}_t
;\mu_\theta(\s^{(n)}_t,\boldsymbol{\tau}_{(t,w)}, n),\Sigma_\theta(\s^{(n)}_t,\boldsymbol{\tau}_{(t,w)}, n))$ with a learnable mean and variance.
We could directly supervise the training of $\mu_\theta$ using the forward (diffusion) process. Following \citet{song2020score,ho2020ddpm_neurips}, we
compute a noisy sample $s_t^{(n)}$ based on the true sample $s_t = s_t^{(0)}$:
\begin{equation}
s_t^{(n)}= \sqrt{\bar \alpha(n)}s_t + \sqrt{1-\bar \alpha(n)}\eps
 \label{eq:sn}
\end{equation}
where $\eps \sim \mathcal{N}(\mathbf{0}, \boldsymbol{I})$ is the noise, $\bar \alpha(n) = \prod_{i=1}^n \alpha(i)$ and the weighting factors $\alpha{(n)} = e^{-\left(\beta_{\text{min}}\left(\frac{1}{N}\right) + (\beta_{\text{max}} - \beta_{\text{min}}) \frac{2n-1}{2N^2}\right)}$
where $\beta_{\text{max}}=10$ and $\beta_{\text{min}}=0.1$ are parameters introduced for empirical reasons \cite{xiao2021tackling}.

We can equally learn to predict the true samples by learning a noise model \cite{luo2022understandingdiffusionmodelsunified}. Hence, we train a noise model $\eps_\theta(\s^{(n)}_t,\boldsymbol{\tau}_{(t,w)}, n)$ that learns to predict the noise vector $\eps$. By using the conditional version of the simplified surrogate objective from \cite{ho2020ddpm_neurips}, we minimize 
\begin{equation}
\mathcal{L}_p(\theta)=\mathop{\mathbb{E}}_{n,\boldsymbol{\tau},\boldsymbol{s_t},\eps}\left[ \left\| \eps - \eps_\theta \left(\s^{(n)}, \boldsymbol{\tau}_{(t,w)} , n \right) \right\|^2 \right]
\label{eq:porlloss}
\end{equation}
where $\boldsymbol{s_t}$ is the state sampled from the dataset $D$ for $t\sim \textit{U}_T(\{w,\dots,T-1\})$, $\s^{(n)}$ is computed according to \cref{eq:sn},
$\eps$ is the noise, and $n\sim \textit{U}_D(\left\{1,\dots,N\right\})$ is the uniform distribution used for sampling the diffusion timestep.

We use the true data sample $\boldsymbol{s_t}$ from the offline RL dataset to obtain the noisy sample in \cref{eq:sn}. Leveraging our model's capacity to learn multimodal distributions, we generate a set of $k$ samples $\{\s_t^{(0)}\}$ as our \textbf{predicted state candidates} in parallel from the reverse diffusion chain.
We use the noise prediction model \cite{ho2020ddpm_neurips}
with the reverse diffusion chain $\s_t^{(n-1)} \mid \s_t^{(n)} $ formulated as
\begin{equation}
\frac{\s_t^{(n)}}{\sqrt{\alpha_{(n)}}} - \frac{1-\alpha_{(n)}}{\sqrt{\alpha_{(n)} (1 - \bar{\alpha}_{(n)})}} \eps_\theta(\s_t^{(n)}, \boldsymbol{\tau}_{(t,w)}, n) + \sqrt{1-\alpha_{(n)}} \eps
\label{eq:revchainsample}
\end{equation}
where $\eps \sim \mathcal{N}(0, \mathbf{I})$ for $n = N, \dots, 1$, and $\eps=0$ for $n=1$ \cite{ho2020ddpm_neurips}. Below, we detail how the state candidates are used during an episode.

\begin{wrapfigure}[22]{R}{0.60\linewidth}
  \vspace{-2.5em}
  \begin{minipage}{\linewidth}
    \begin{algorithm}[H] 
    \caption{Candidate Selection}
    \label{alg:porlcandselection}
    \begin{algorithmic}[1]
    \STATE  Sample $\{\{\hat{b}\}^1_l, \dots, \{\hat{b}\}^P_l\}\sim \textit{Zero-Shot FM}$
    \STATE \textbf{for} each episode $p = 1,\cdots, P$ \textbf{do}
    \STATE \hspace{1em} $t=0$
    \STATE \hspace{1em} Reset environment $o_0 \sim \mathcal{E}$
    \STATE \hspace{1em} $\tilde{s}$ $\gets$ $o_0- \overline{\{\hat{b}\}^p_l}$ 
    \STATE \hspace{1em} Initialize $\boldsymbol{\tau}_{(t,w)}$
    \STATE \hspace{1em} \textbf{while} not \textit{done} \textbf{do}
    \STATE \hspace{2em} Sample $a^{(0)} \sim \pi_\phi(a|\tilde{s})$
    \STATE \hspace{2em} Take action $a^{(0)}$ in $\mathcal{E}$, observe $o_{t+1}$
    \STATE \hspace{2em} $\{\hat{s}^b_{t+1}\}_l \gets$ $o_{t+1} - \{\hat{b}\}^p_l$
    \STATE \hspace{2em} $\boldsymbol{\tau}_{(t+1,w)}=\operatorname{PUSH}\left(\boldsymbol{\tau}_{(t,w)},\left(\Delta o_{t+1}, a^{(0)}\right)\right)$
    \STATE \hspace{2em} $\boldsymbol{\tau}_{(t+1,w)}=\operatorname{POP}\left(\boldsymbol{\tau}_{(t+1,w)},\left(\Delta o_{t-w+1}, a_{t-w}\right)\right)$
    \STATE \hspace{2em} \textbf{if} $t>w$ \textbf{then}
    \STATE \hspace{3em} Sample $\{\s_{t+1}^{(0)}\}_k$ from \forl by Eq. \ref{eq:revchainsample}
    \STATE \hspace{3em} $\tilde{s}$ $\gets$ \textbf{\candidsel}($\{\s_{t+1}^{(0)}\}_k,\{\hat{s}^b_{t+1}\}_l$)
    \STATE \hspace{2em} \textbf{else}
    \STATE \hspace{3em} $\tilde{s}$ $\gets$ $o_{t+1} - \overline{\{\hat{b}\}^p_l}$
    \STATE \hspace{2em} \textbf{end if}
    \STATE \hspace{2em} $t\gets t+1$
    \STATE \hspace{1em} \textbf{end while}
    \STATE \textbf{end for}
    \end{algorithmic}
    \end{algorithm}
  \end{minipage}
  \vspace{-1em}
\end{wrapfigure}

\subsection{Forecasting using Zero-Shot Foundation Model}\label{sec:lagllama}
Because we assume that the offsets $b^j$ originate from a time series, we propose using a probabilistic zero-shot forecasting foundation model \textit{(Zero-Shot FM)} such as Lag-Llama \cite{rasul2023lag}, to forecast future offsets from past ones.
We assume that after $P$ episodes, the true offsets are revealed, and we predict the offsets for the following $P$ episodes. Using the probabilistic \zeroshotfm we generate (${\hat b^j_l,\dots,\hat b_l^{j+P-1}}$), where ($l$) denotes the number of samples generated for each episode (timestamp). Since Lag-Llama is a probabilistic model, it can generate multiple samples per timestamp, conditioned on $C$ number of past contexts (${b^{j-C},\dots, b^{j-1}}$). In practice, we forecast every dimension of $b$ independently since the \textit{Zero-Shot FM} (Lag-Llama \cite{rasul2023lag}) is a univariate probabilistic model.

\subsection{\forl State Estimation}\label{sec:fusion}
The next step in our method is to fuse the information from the forecaster and the diffusion model into a state estimate used for control at test time.

At the beginning of an episode, no information can be obtained from the diffusion model, so for the first $w$ steps we only rely on the forecaster's mean prediction, \ie ${\tilde{s}_t} = o_t - \overline{\hat{b}^j}$ where the mean is taken over the $l$ samples.

\begin{wrapfigure}[12]{r}{0.5\linewidth}
    \vspace{-1.5em}
    \includegraphics[width=\linewidth]{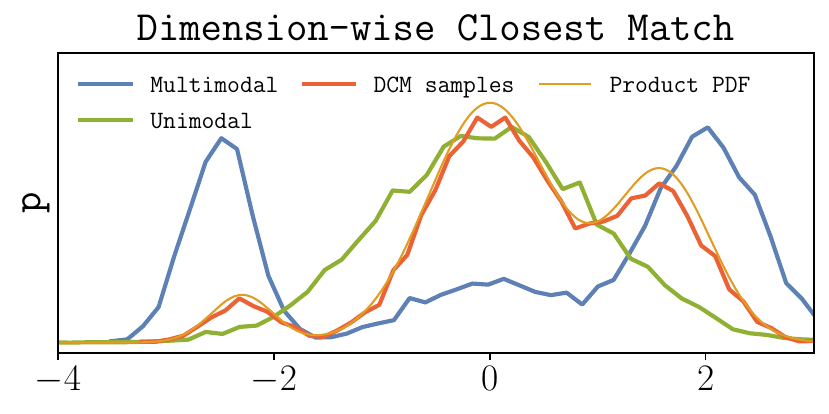}
    \vspace{-1.2em}
 \caption{\label{fig:dcm} Distribution of samples produced by \candidsel (histograms for 10k samples for illustration).}
 \vskip1em
\end{wrapfigure}

As soon as $w$ steps are taken, our \emph{\forl State Estimation} improves on the inferred state as detailed below.
Figure~\ref{fig:porlframwork} offers an overview of the entire system and Algorithm~\ref{alg:porlcandselection} provides a detailed pseudocode. 

To recap, the diffusion model generates samples \(\{\s_t^{(0)}\}_k\) from the in-episode history $\tau$, \cref{eq:trajeq}. These samples represent a multimodal distribution of plausible state regions.
The \textit{Zero-Shot FM} generates $l$ samples of offsets \(\{\hat{b}\}_l\) from which we compute forecasted states using \(\{\hat{\s}_t\}_l = o_t - \{\hat{b}\}_l\).

\paragraph{\forl: \candidselmethodname (\candidsel)}
We propose a lightweight approach to sample a good estimate based on the samples from the multimodal (\textit{diffusion model \(\{\s_t^{(0)}\}_k\)}) and unimodal (\textit{Zero-Shot FM} $\{\hat{s}^b_{t}\}_l$) distributions.
Let
\(
\mathcal{D}_\text{diffusion}
= \{\x_1, \dots, \x_k\},
\quad
\mathcal{D}_\text{timeseries}
= \{\mathbf{y}_1, \dots, \mathbf{y}_l\},
\)
where $\x_i, \mathbf{y}_j \in \mathbb{R}^n$. Then \candidsel constructs $\mathbf{z}\in \mathbb{R}^n$ by
\[
z_d \;=\; y_{j^*(d), d} \quad\text{where}\quad
j^*(d) \;=\;\arg\min_{j}
\Bigl(\;\min_{i}\bigl|\;x_{i,d} - y_{j,d}\bigr|\Bigr),
\]

where $d=1\dots n$.
In other words, for each dimension $d$, we choose the sample from $\mathcal{D}_\text{timeseries}$ that has the closest sample in $\mathcal{D}_\text{diffusion}$. The process is straightforward yet effective, and under ideal sampling conditions for a toy dataset in \cref{fig:dcm}, \candidsel approximately samples from the product distribution. \candidsel uses a non-parametric search to find the forecast sample with the highest score, which corresponds to the minimum dimension-wise distance. \candidsel's prediction error is governed by the accuracy of the forecast samples in the unimodal $\mathcal{D}_\text{timeseries}$ that achieves this best score. As we will demonstrate in the experiments, this approach empirically yields lower maximum errors and is more stable compared to other methods.

\paragraph{\forl Algorithm}
Algorithm~\ref{alg:porlcandselection} summarizes the entire inference process at test time. We begin the episode by relying on the forecasted states \(\tilde{s}_0\). As more transitions \((\Delta o_{t}, a_{t-1})\) become available, the \forl diffusion model proposes candidate states \(\{\s_t^{(0)}\}_k\) through \textit{retrospection}—reasoning over the past in-episode experience to adapt state estimation on the fly when they begin to diverge from predictions. We then invoke \candidsel to blend the diffusion model's candidates with the foundation model's unimodal forecasts and obtain the final state estimate \(\tilde{s}_t\). We use an off-the-shelf offline RL policy such as Diffusion-QL (\difql) \cite{wang2023diffusion} to select the agent's action \(a_t\).

\begin{wrapfigure}[12]{r}{0.5\linewidth}
    \centering
     \vspace{-2.1em}
    \includegraphics[width=1.0\linewidth]{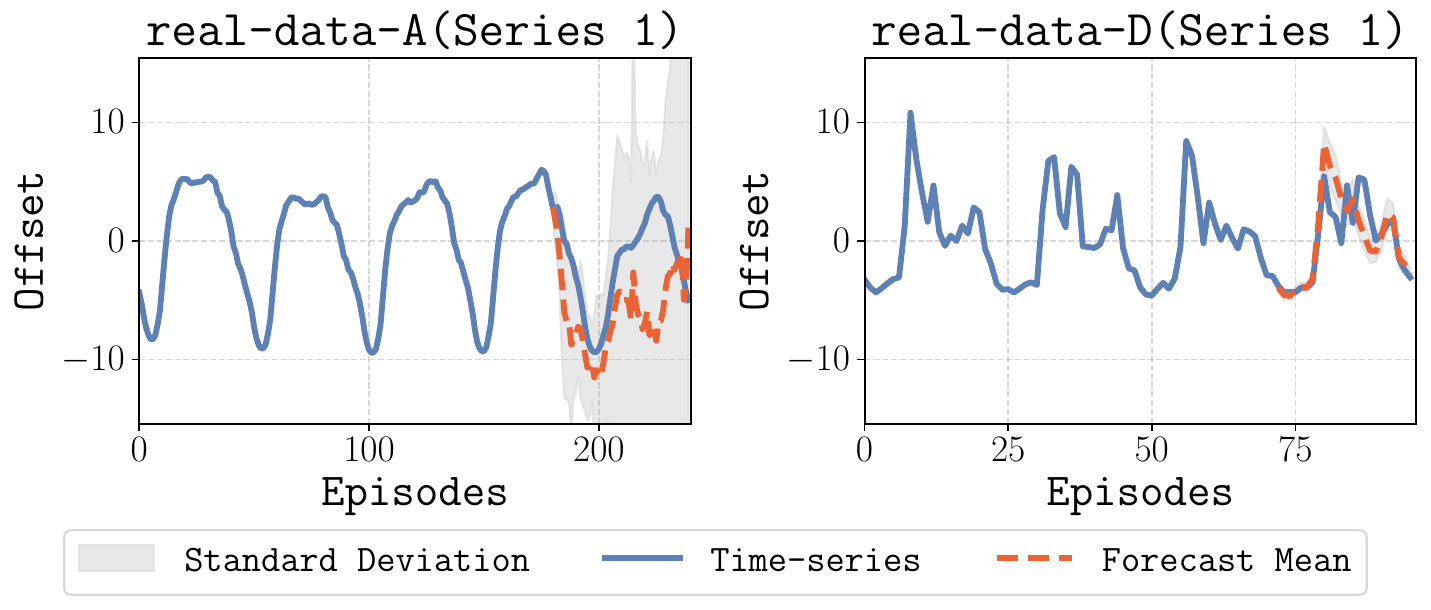}
 \caption{
 Zero-shot forecasting results of Lag-Llama \cite{rasul2023lag} for the first univariate series (plotted) from the \texttt{real-data-A,D} datasets; experiments use the \textbf{first two series from each dataset}.}
\label{fig:tsdatasetforecasts}
 \vskip1em
\end{wrapfigure}

\paragraph{Summary} By combining a powerful \emph{zero-shot} forecasting model with a \emph{conditional diffusion} mechanism, \forl addresses partial observability in continuous state and action space when ground-truth offsets are unavailable. This procedure is performed in the \textit{absence of ground-truth offsets for past, current, and future episodes over the interval $j:j+P$ at test time}. \candidsel provides a computationally inexpensive yet effective way of using the multimodal diffusion candidates and unimodal time-series forecasts. This robust adaptation approach yields a state estimate \(\tilde{s}_t\), aligned with the agent's retrospective experience in the stationary offline RL dataset, incorporating a prospective external offset forecast.

\input{tables/avgrewall}

\section{Experiments} 
We evaluate \forl across navigation and manipulation tasks in D4RL \citep{fu2020d4rl} and OGBench \cite{ogbench_park2025} offline RL environments, each augmented with five real-world non-stationarity domains sourced from \citep{gluonts_jmlr}. \cref{fig:tsdatasetforecasts} presents the ground truth, forecast mean, and standard deviation from Lag-Llama \cite{rasul2023lag} for the \textit{first series} of \aed and \electricitynips. Our experiments address the following questions: \textbf{(1)} Does FORL maintain state-of-the-art performance when confronted with unseen non-stationary offsets?
\textbf{(2)} How can we use \forl when we have no access to delayed past ground truth offsets?
\textbf{(3)} How does \candidsel compare to other fusion approaches?
\textbf{(4)} Can \forl handle intra-episode non-stationarity?
\textbf{(5)} How gracefully does performance degrade as offset magnitude $\alpha$ is scaled from $0 \text{ (no offset}) \rightarrow 1 \text{ (our evaluation setup})$? \textbf{(6)} Can \forl serve as a plug-and-play module for different offline RL algorithms without retraining?
Extended results, forecasts for the remaining series, and implementation details are provided in the Appendix. Results average 5 seeds, unless noted.

\paragraph{Baselines}
We compare our approach with the following baselines:
\difql \cite{wang2023diffusion}, Flow Q-learning (\fql) \cite{fql_park2025} are diffusion-based and flow-based offline RL policies, respectively, that do not incorporate forecast information.
\difqlmeanlag, \fqlmeanlag extend \difql and \fql by using the sample mean of the forecasted states $\{\hat{s}_t\}_l$ at each timestep (using the constant per-episode predicted $b^j$). \difqlmedlag similarly extends \difql using the median. \dmbplag is a variant of the Diffusion Model–Based Predictor (\dmbp)\cite{zhihedmbp} (a robust offline RL algorithm designed to mitigate state-observation perturbations at test time, detailed in Appendix~\ref{app:baselines}) that integrates forecasted states from \zeroshotfm \cite{rasul2023lag} into its state prediction module. By using the model learned from the offline data, \dmbplag aims to refine the forecasted states to make robust state estimations. The underlying policies throughout our experiments are identical policy checkpoints for both our method and the baselines. 

\begin{wrapfigure}[9]{r}{0.45\linewidth}
   \vspace{-2.25em}
    \includegraphics[width=\linewidth]{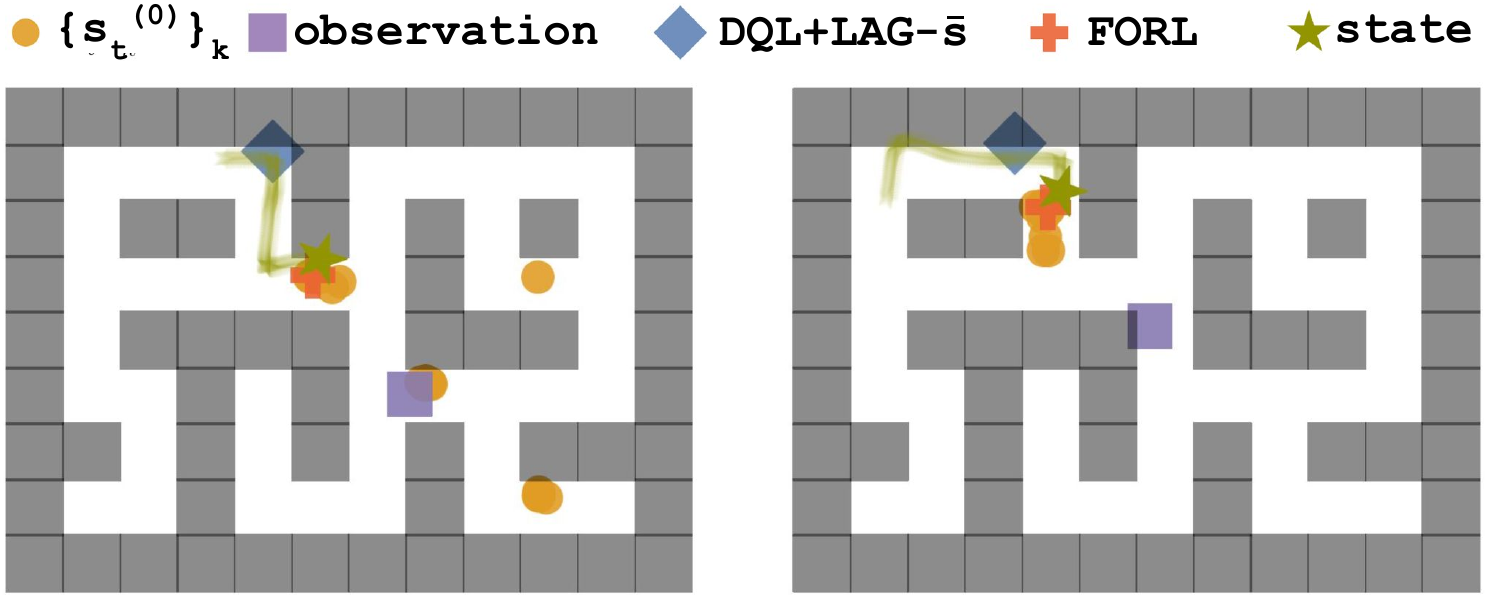}
    \vspace{-1.2em}
\caption{\label{fig:mazepredfigs}Visualization of states, predicted states as the agent navigates the environment.}
 \vskip1em
\end{wrapfigure}
\begin{wrapfigure}[7]{r}{0.28\linewidth}
\vskip-5.5em
\includegraphics[width=0.95\linewidth]{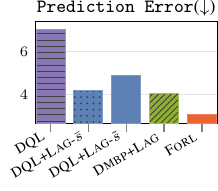}
\vskip-0.6em
\caption{Prediction Error in recovering true agent state.}\label{fig:predacc}
\end{wrapfigure}

\paragraph{Illustrative Example}
\cref{fig:mazepredfigs,fig:mazepredfigs2detailedallablations} illustrate an agent navigating the \mazel environment where the true position is labeled as ``state''. The agent receives an observation indicating where it \textit{believes} it is located due to unknown time-dependent factors. The candidate states predicted by the \forl diffusion model are shown as circles. Importantly, the agent's $(\Delta o, a)$-trajectory can reveal possible states for the agent. \forl's diffusion model (\dm) component predicts these candidate states by using observation changes $(\Delta o)$ and corresponding actions $(a)$. The possible candidate regions where the agent can be are limited, and our model successfully captures these locations. \forl's candidate selection module (\candidsel) uses the samples from the forecaster and the diffusion model to recover a close estimate for the state. In contrast, the baseline \difqlmeanlag relies on the forecaster \cite{rasul2023lag} for state predictions, which are significantly farther from the actual state. Consistent with the results in \cref{fig:predacc}, \forl reduces prediction errors at test-time, thereby improving performance. 

\subsection{Results}
FORL outperforms both pure forecasting (\difqlmeanlag) and the two-stage strategy that first predicts offsets with a time-series model and then applies a noise-robust offline RL algorithm (\dmbplag). Its advantage is consistent across previously unseen non-stationary perturbations from five domains, each introducing a distinct univariate series into a separate state dimension at test time.
We present the average normalized scores over the prediction length $P$ across multiple episodes run in the D4RL \cite{fu2020d4rl} and OGBench \cite{ogbench_park2025} for each time-series in Table \ref{tab:avgrewtable}. We conduct pairwise Welch's t-tests across all settings.
Figure~\ref{fig:predacc} plots the $\ell_2$ norm between the ground-truth states $s_t$ and those predicted by \forl and the baselines in the \antmaze and \mazed environments. Consistent with the average scores, \forl achieves the lowest prediction error on average.

\begin{wrapfigure}[8]{r}{0.30\linewidth}
\vskip-4.1em
\includegraphics[width=0.98\linewidth]{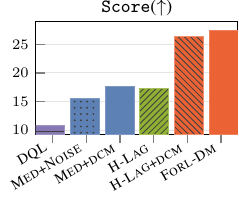}
\vskip-0.5em
\caption{DM Ablations}\label{fig:avgrew_dmablation}
\end{wrapfigure}

\subsubsection{No Access to Past Offsets}
\label{subsec:noaccesspastoffsets_main}
We evaluate different variants of using \dm and \zeroshotfm when we do not have any access to past offsets in \cref{fig:avgrew_dmablation}. \textbf{\forl-\dm (\dm):} Diffusion Model utilizes the candidate states generated by the \forl's diffusion model component (Section~\ref{subsec:porldm}), which can be a multimodal distribution (\cref{fig:hypomaze}). Compared to \dm, the full \forl framework yields a 97.8\% relative performance improvement. Notably, \dm performs on par with our extended baselines that incorporate historical offsets and forecasting—\dmbplag, \difqlmeanlag, and \difqlmedlag. Moreover, without access to historical offset information before evaluation, \dm achieves a 151.4\% improvement over \difql, demonstrating its efficacy as a standalone module trained solely on a standard, stationary offline RL dataset without offset labels. \textbf{\hlag:} We maintain a history of offsets generated by \dm over the most recent $C$ episodes (excluding the evaluation interval $P$, since offsets are not revealed after episode termination at test-time). We then feed this history into the \zeroshotfm to generate offset samples for the next $P$ evaluation episodes. These samples are applied directly at test time. \textbf{\hlagdcm:} We initially follow the same procedure in \hlag to obtain predictions from \zeroshotfm. Then, we apply \textbf{\candidsel} to these predicted offsets and the candidate states generated by \forl's diffusion model. We also compare against \meddcm and \mednoise, simpler median-based heuristics detailed in \cref{app:dmablations}. Empirically, \hlagdcm outperforms \hlag, demonstrating that \candidsel with \forl's diffusion model can improve robustness. Overall, scores and prediction errors indicate that just using the samples from \dm has better scores on average, while \hlagdcm is more stable in \cref{fig:avgrewcomparison_nopastoffsets}.

\subsubsection{ \candidselmethodname (\candidsel) Ablations}
\begin{wrapfigure}[7]{r}{0.30\linewidth}
\vskip-4.1em
\includegraphics[width=0.9\linewidth]{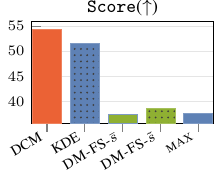}
\vskip-0.5em
\caption{Candidate Selection}\label{fig:avgrewtik_dcm}
\end{wrapfigure}
We compare \forl(\candidsel) against four alternative fusion strategies. \scott: For each dimension, we fit a kernel density estimator (KDE) on $\mathcal{D}_\text{diffusion}=\{\boldsymbol{s}_t^{(0)}\}_k$ and then we evaluate that probability density function for each point in $\mathcal{D}_\text{timeseries}$. Then, we take the product of these densities in each dimension to obtain the weight for each sample $\hat{s}^b_t$. 
We obtain a single representative sample by taking the weighted average of samples in $\mathcal{D}_\text{timeseries}$.
To ensure stability, when the sum of the weights is near zero, we use the mean of the $\mathcal{D}_\text{timeseries}$ as the states. We use Scott's rule \cite{scot1992multivariate} to compute the bandwidth. \closemean, \closemedian select the closest prediction from DM to the mean and median of the \zeroshotfm's predictions, respectively. \forl(\maxlikelihood) constructs a diagonal multivariate distribution from the dimension-wise mean and standard deviation of the forecasted states, then selects the sample predicted by our diffusion model with the highest likelihood under that distribution. Although all baselines fuse information using the same two sets generated by the diffusion model and \zeroshotfm, \candidsel has higher performance. In \cref{tab:dcm_max_minerrorcomparisonfig6} we compute the maximum, minimum, and mean prediction error values over the test episodes used in \cref{fig:mazepredfigs}. \forl(\candidsel) yields significantly stable prediction errors ($\texttt{Maximum Error}\downarrow$:\textbf{2.40}) for both maximum error and mean error compared to \forl(\maxlikelihood) ($\texttt{Maximum Error}\downarrow$:\textbf{9.33}) demonstrating its robustness.

\subsubsection{Intra-episode Non-stationarity}
\begin{wrapfigure}[9]{r}{0.28\linewidth}
\vskip-4em
\includegraphics[width=\linewidth]{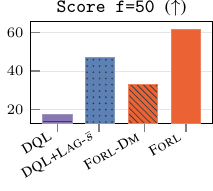}
\vskip-0.5em
\caption{Intra-Episode Performance}\label{fig:avgrew_intraepisode}
\end{wrapfigure}
Our framework can natively handle intra-episode offsets, where the offset changes every $f=50$ timesteps. In this setting, the offsets become available after the episode terminates, but the agent is subject to a time-dependent unknown offset within the episode. Zero-shot forecasting foundation module can generate samples before the episode begins. Our diffusion model (\forl-\dm) itself does not rely on the forecasts of the foundation module and only tracks observation changes and actions which are invariant to the offsets. The \candidsel can adaptively fuse information from both models at each timestep without requiring any hyperparameters. \cref{tab:avgrew_intraepisode,fig:avgrew_intraepisode} show the average scores for \difql vs. \forl-\dm and \difqlmeanlag vs. \forl. Among the algorithms that do not use any past ground truth offsets \difql and \forl-\dm, only using the diffusion model of \forl significantly increases performance. When we have access to past offsets, \forl obtains a superior performance. This shows that our method covers both cases, when information is available and not available, even when offsets are not constant throughout the episode.

\subsubsection{Offset-Scaling} 
\begin{wrapfigure}[8]{r}{0.4\linewidth}
    \centering
        \vspace{-4.1em}
    \includegraphics[width=\linewidth]{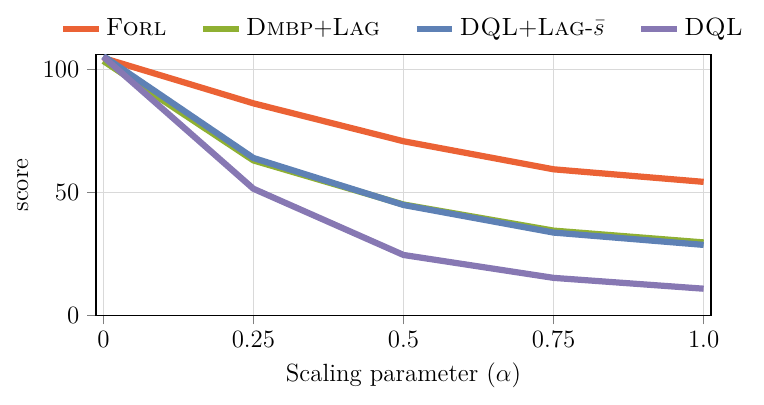}
    \vskip-0.5em
\caption{\label{fig:linetikavgrew} Impact of offset scaling $(\alpha)$ on average normalized scores.}
\end{wrapfigure}
We scale the offsets with $\alpha$ across all maze experiments. We conduct experiments in 5 environments (all antmaze and maze2d used in Table \ref{tab:avgrewtable}) across 5 time-series dataset setups with $\alpha \in $\{0, 0.25, 0.5, 0.75, 1.0\}, where $\alpha = 0$ is the standard offline RL environment used during training and $\alpha = 1.0$ is our evaluation setup. The results show that \forl outperforms the baselines, confirming its robustness. Even a small scaling of $0.25$ results in a sudden drop in performance, whereas \forl only experiences a gradual decrease in Figure~\ref{fig:linetikavgrew}. Detailed results for each environment and $\alpha$ pairs are provided in Appendix Figure~\ref{fig:alpha-maze-comparison}.

\input{tables/mazelargerewtable}

\subsubsection{Policy-Agnostic}
In the \mazel experiments (in Table \ref{tab:offline}, \mazem in Appendix \cref{table:rorltd3bc_all}), we use Robust Offline RL (\rorl) \citep{yang2022rorl},  and \tdthreebc \cite{fujimoto2021a} offline RL algorithms instead of \difql \citep{wang2023diffusion}, to analyze the effect of offline RL policy choice during evaluation. \rorlmeanlag and \tdthreebcmeanlag extend \rorl and \tdthreebc by using the sample mean of the forecasted states $\{\hat{s}_t\}_l$ at each timestep (using the constant per-episode predicted $\overline{\hat{b}^j}$). Results indicate that using a robust offline RL algorithm during training significantly increases performance (71.9) compared to \difql (38.8) and \tdthreebc (34.4) at test time when used with \forl, no increase when used alone, and a marginal increase with Lag-Llama and \dmbplag. We observe similar performance gains when applying \forl to other policies (\cref{app:iql,app:fql}), including Implicit Q-Learning (\iql) \citep{iql_Kostrikov_2021} and \fql \citep{fql_park2025}, as detailed in \cref{tab:iqlperformance} and \cref{tab:avgrew_og_dcmablation}.

\section{Related Work}
\paragraph{Reinforcement Learning in Non-Stationary Environments}
Existing works in non-stationary reinforcement learning (RL) predominantly focus on adapting to changing transition dynamics and reward functions. \citet{ackermann2024offline} propose an offline RL framework that incorporates structured non-stationarity in reward and transition functions by learning hidden task representations and predicting them at test time. Although our work also investigates the intersection of non-stationary environments and offline RL, we assume stationarity during training. To learn adaptive policies \textit{online}, meta-learning algorithms have been proposed as a promising approach \citep{finn2019online,al-shedivat2018continuous,ada_umcnp_2024}. \citet{al-shedivat2018continuous} explores a competitive multi-agent environment where transition dynamics change. While these approaches provide valuable insights, they often require samples from the current environment and struggle in non-trivial non-stationarity, highlighting the need for more future-oriented methods \cite{chandak2020optimizing,lee2023tempo}. Examples of such future-oriented approaches include Proactively Synchronizing Tempo (ProST) \cite{lee2023tempo} and Prognosticator \cite{chandak2020optimizing}, which address the evolution of transition and reward functions over time. ProST leverages a forecaster, namely, Auto-Regressive Integrated Moving Average (ARIMA), and a model predictor to optimize for future policies in environments to overcome the time-synchronization issue in time-elapsing MDPs. This approach aligns with our focus on time-varying environments and similarly utilizes real-world finance (e.g., stock price) time-series datasets to model non-stationarity. Both ProST and Prognosticator assume that states are fully observable during testing and that online interaction with the environment is possible during training—conditions that are not always feasible in the real world. Instead, our approach assumes that states are not fully observable and that direct interaction with the environment during training is not feasible, necessitating that the policy be learned exclusively from a pre-collected dataset.

\paragraph{Robust offline RL} Testing-time robust offline RL methods \dmbp \citep{zhihedmbp}, \rorl \citep{yang2022rorl} examine scenarios where a noise-free, stationary dataset is used for training, but corruption is introduced during testing. This is distinct from \citep{zhihedmbp}, training-time robust offline RL \cite{ye2023corruption,zhang2022corruption}, which assumes a corrupted training dataset. Both \rorl \citep{yang2022rorl}  and \dmbp \citep{zhihedmbp} assume access only to a clean, uncorrupted offline RL dataset, as \forl, and they are evaluated in a perturbed environment. To the best of our knowledge, \forl is the first work to extend this setting to a non-Markovian, time-evolving, non-stationary deployment environment. We focus on time-dependent exogenous factors from real-data that are aligned with the definition of a non-stationary environment \cite{khetarpal2022towards}.

\paragraph{Diffusion models in offline RL}
Diffusion models \cite{DBLP:journals/corr/Sohl-DicksteinW15} have seen widespread adoption in RL \cite{zhu2023diffusion,chen2024deep} due to their remarkable expressiveness, particularly in representing multimodal distributions, scalability, and stable training properties. In the context of offline RL, diffusion models have been used for representing policies \cite{wang2023diffusion,he2023diffcps,hansen2023idql,adaralsrdp2024}, planners \cite{janner2022diffuser,coleman2023discrete}, data synthesis \cite{mtdiff_he_2023,liang2023adaptdiffuser}, and removing noise \cite{zhihedmbp}. Notably, Diffusion Q-learning \cite{wang2023diffusion} leverages conditional diffusion model policies to learn from offline RL datasets, maintaining proximity to behavior policy while utilizing Q-value function guidance. In contrast, our method harnesses diffusion models to learn from a sequence of actions and effect tuples, leveraging the multimodal capabilities of diffusion models to identify diverse candidate locations of the hidden states.
\section{Conclusion}
We introduce \textbf{F}orecasting in Non-stationary \textbf{O}ffline \textbf{RL} (\forl), a novel framework designed to be robust to passive non-stationarities that arise at test time. This is crucial when an agent trained on an offline RL dataset is deployed in a non-stationary environment or when the environment begins to exhibit partial observability due to unknown, time-varying factors. \forl leverages diffusion probabilistic models and zero-shot time series foundation models to correct unknown offsets in observations, thereby enhancing the adaptability of learned policies. Our empirical results across diverse time-series datasets, OGBench \cite{ogbench_park2025} and D4RL \cite{fu2020d4rl} benchmarks, demonstrate that \forl not only bridges the gap between forecasting and non-stationary offline RL but also consistently outperforms the baselines. Our approach is currently limited by the assumption of additive perturbations. For future work, we plan to extend our work to more general observation transformations. 
\begin{ack}
Georg Martius is a member of the Machine Learning Cluster of Excellence, EXC number 2064/1 – Project number 390727645. Co-funded by the European Union (ERC, REAL-RL, 101045454). Views and opinions expressed are, however, those of the author(s) only and do not necessarily reflect those of the European Union or the European Research Council. Neither the European Union nor the granting authority can be held responsible for them. This work was supported by the German Federal Ministry of Education and Research (BMBF): Tübingen AI Center, FKZ: 01IS18039A. This work was in part supported by the INVERSE project (101136067) funded by the European Union and JSPS KAKENHI Grant Numbers JP23K24926, JP25H01236. The numerical calculations reported in this paper were partially performed at TUBITAK ULAKBIM, High Performance and Grid Computing Center (TRUBA resources). The authors would like to thank René Geist, Tomáš Daniš, Ji Shi, and Leonard Franz for their valuable comments on the manuscript.
\end{ack}
\bibliography{references}
\bibliographystyle{unsrtnat}
%\newpage
%\input{forlchecklist}
\newpage
\appendix
\onecolumn

\begin{center}
  \hrule height 4pt
  \vspace{0.25in}
  \vspace{-\parskip}
  
  {\LARGE \bfseries Forecasting in Offline Reinforcement Learning for Non-stationary Environments:\\ Supplementary Material \par}
  
  \vspace{0.29in}
  \vspace{-\parskip}
  \hrule height 1pt
  \vspace{0.09in}
  
  \vspace{0.3in} 
  
  \begin{tabular}[t]{c}
    \bfseries Suzan Ece Ada$^{1,2}$ \quad
    \bfseries Georg Martius$^{2}$ \quad
    \bfseries Emre Ugur$^{1}$ \quad
    \bfseries Erhan Oztop$^{3,4}$ \\[1ex]
    $^{1}$Bogazici University, Türkiye \quad
    $^{2}$University of Tübingen, Germany \\
    $^{3}$Ozyegin University, Türkiye \quad
    $^{4}$Osaka University, Japan \\
    \texttt{ece.ada@bogazici.edu.tr}
  \end{tabular}
\end{center}

\section{Related Work: Continued}\label{app:relatedworkextended}
\paragraph{Offline Reinforcement Learning}
A high-level overview of existing work in offline reinforcement learning (RL) identifies three predominant strategies: policy constraint methods \cite{Fujimoto2018OffPolicyDR,Siegel2020Keep,wang2023diffusion}, pessimistic value function methods which assign low values to OOD actions \cite{ma2021conservative}, and model-based offline RL methods. Policy constraint methods actively avoid querying OOD actions during training by leveraging probabilistic metrics which can be explicit \cite{jaques2019way,wu2019behavior}, implicit \cite{peng2019advantage,nair2020awac} $f$-divergence \cite{nowozin2016f}, or integral probability metrics \cite{kumar2019stabilizing}. These metrics ensure that the learned policy $\pi_\theta$ remains close to the behavior policy $\pi_\beta$ that generated the offline RL dataset\cite{levineofflinerltutorial_20202}. Similarly, pessimistic value function approaches regularize the value function or the Q-function to avoid overestimations in OOD regions. Model-based offline RL methods \cite{DBLP:journals/corr/abs-2102-08363}, on the other hand, focus on learning the environment's dynamics, benefiting from the strengths of supervised learning approaches. However, in the same vein, these methods are susceptible to the distribution shift problem in Offline RL \cite{levineofflinerltutorial_20202}. We detail the offline RL algorithms used in our experiments in \cref{app:offlinerl_policies_detailed}.

\citet{park2024value} emphasize the challenges of generalizing policies to test-time states which are not in the support of the offline RL dataset. While prior works have investigated the issue of generalization \citep{adaralsrdp2024,Mazoure2022Nips}, in testing-time robust RL methods \citep{zhihedmbp} this challenge is exacerbated through the introduction of noise into the states by an unknown adversary.

\section{Background}\label{app:background}

\subsection{Reinforcement Learning}
\label{app:background_reinforcementlearning}
Markov Decision Processes (MDPs) are often used to formalize Reinforcement Learning (RL). MDP is defined by the tuple $\mathcal{M}\doteq \bigl( \mathcal {S}, \mathcal{ A },\mathcal{T}, \mathcal{R}, \rho_0,\gamma \bigr)$ where $\mathcal{ S } $ is the state space, $\mathcal{ A } $ is the action space, $\mathcal{T}$ is the transition function (which may be deterministic or stochastic), $  \mathcal{R}: \mathcal { S } \times \mathcal{ A } \rightarrow \mathbb{R}$ is the reward function, $\rho_0$ is the initial state distribution and $ \gamma \in [0,1)$ is the discount factor \cite{puterman2014markov}. In online and off-policy RL algorithms  an agent can interact with the environment using a parameterized policy $\pi_\theta(\boldsymbol{a}|\s)$, to maximize the expected return $\mathbb{E}_{\pi}\left[\sum_{t} \gamma^t r\left(\s_t, \boldsymbol{a}_t\right)\right]$. In contrast, offline RL requires the agent to learn from a static dataset $\mathcal{D}=\{(\s_k,\boldsymbol{a}_k,\s'_k,r_k)\}^N_{k=1}$ generated by a generally unknown behavior policy $\pi_\beta(\boldsymbol{a}|\s)$ \cite{levineofflinerltutorial_20202}.

\subsection{Non-Stationary Environments} \label{app:background:nonstationary}
We next review tangential definitions and formalisms used for non-stationary environments.
\begin{definition}[Partially Observable Markov Decision Processes \cite{kaelbling1998planning,chandak2022reinforcement,khetarpal2022towards}]
A Partially Observable Markov Decision Process (POMDP) is given by the tuple
\[
\hat{\mathcal{M}} \,=\, \bigl( \mathcal{S}, \mathcal{A}, \mathcal{O}, \mathcal{T}, \mathcal{R}, \x, \rho_0, \gamma\bigr),
\]
where \(\mathcal{O}\) is the observation space and \(\x\) is the observation function. This function can be deterministic (\(\x: \mathcal{S} \rightarrow \mathcal{O}\)) \cite{bonet2012deterministic,khetarpal2022towards} or stochastic (\(\x(o \mid s)\)) for $o \in \mathcal{O}, s\in \mathcal{S}$ \cite{chandak2022reinforcement}.
\end{definition}

To capture a broader range of non-stationary RL scenarios, \citet{khetarpal2022towards} present a \emph{general non-stationary RL} formulation, which allows each component of the underlying MDP or POMDP to evolve over time. Concretely, this time-varying tuple is denoted as $\bigl(\mathcal{S}(t), \mathcal{A}(t), \mathcal{T}(t), \mathcal{R}(t), \x(t), \mathcal{O}(t) \bigr)$ where each element is represented by a function \(\varphi(i,t)\), indicating its variation with time \(t\) and input $i$ \cite{khetarpal2022towards}. \textit{Scope}, denoted by a set \(\kappa\), specifies which of these components vary.

\begin{definition}[Scope of Non-Stationarity \cite{khetarpal2022towards}]
\label{def:defgencrl}
Given the general non-stationary RL framework, the \emph{scope of non-stationarity} is the subset
\[
\kappa \,\subseteq\, \{\mathcal{S}, \mathcal{A}, \mathcal{R}, \mathcal{T}, \x, \mathcal{O}\},
\]
indicating which components of the environment evolve over time.
\end{definition}

Notably, \citet{chandak2022reinforcement} defines Non-Stationary Decision Process (NSDP) as a sequence of POMDPs, with non-stationarity confined to the subset \(\kappa \subseteq \{\mathcal{R}, \mathcal{T}, \x\}\), where the initial state distribution \(\rho_{0,j}\) varies between POMDPs. Existing research in non‑stationary RL largely focuses on the episodic evolution of transition dynamics and reward functions \cite{ackermann2024offline}. In contrast, the evolution of observation functions remains underexplored, despite its potential for real-world applicability and thus demands further investigation \cite{khetarpal2022towards}. Handling inaccurate state information is crucial in non-stationary RL since an agent may encounter non-stationarity due to its own imperfect perception of the state while the underlying physics of the environment remains unchanged \cite{khetarpal2022towards}. 

\subsection{Diffusion Models}\label{app:background:diffusion}
Diffusion models \cite{DBLP:journals/corr/Sohl-DicksteinW15} aim to model the data distribution with $p_\theta(\boldsymbol{x}_0):=\int p_\theta(\boldsymbol{x}_{0:T}) \, d\boldsymbol{x}_{1:T}$ from samples $x_0$ in the dataset. The joint distribution $p_\theta(\boldsymbol{x}_{0:T})$, is modeled as a Markov chain, where $\boldsymbol{x_1}, ... , \boldsymbol{x_T}$ are the latent variables with the same dimensionality as the data samples. The joint distribution is given by
\begin{equation}
p_\theta\left(\boldsymbol{x}_{0:T}\right):=\mathcal{N}(\boldsymbol{x}_T;\mathbf{0},\mathbf{I}) \prod_{t=1}^T p_\theta\left(\boldsymbol{x}_{t-1} \mid \boldsymbol{x}_t\right) \quad
\end{equation}
where $p_\theta\left(\boldsymbol{x}_{t-1} \mid \boldsymbol{x}_t\right):=\mathcal{N}\left(\boldsymbol{x}_{t-1} ; \boldsymbol{\mu}_\theta\left(\boldsymbol{x}_t, t\right), \mathbf{\Sigma}_\theta\left(\boldsymbol{x}_t, t\right)\right)$. The forward process $q(\boldsymbol{x}_{1:T} \mid \boldsymbol{x}_0):=\prod_{t=1}^{T} q(\boldsymbol{x}_t \mid \boldsymbol{x}_{t-1})$ involves adding a small amount of Gaussian noise to the data sample at each diffusion timestep to obtain the latent variables following a variance schedule $\{\beta_t=1-\alpha_t \in (0,1)\}^T_{t=1}$. Here, the encoder transitions are $q(\boldsymbol{x}_t \mid \boldsymbol{x}_{t-1}):=\mathcal{N}\left(\boldsymbol{x}_t; \sqrt{\alpha_t} \boldsymbol{x}_{t-1}, (1 - \alpha_t) \mathbf{I}\right)$ \cite{ho2020ddpm_neurips}.
Assuming we have access to the true data sample during training, using recursion and the reparameterization trick, we can obtain samples $\boldsymbol{x}_t$ at any timestep t in closed form with $q\left(\boldsymbol{x}_t \mid \boldsymbol{x}_0\right)=\mathcal{N}\left(\boldsymbol{x}_t ; \sqrt{\bar{\alpha}_t} \boldsymbol{x}_0,\left(1-\bar{\alpha}_t\right) \mathbf{I}\right)$ where $\bar{\alpha}_t=\prod_{i=1}^t \alpha_i$ \cite{ho2020ddpm_neurips}. During training, the evidence lower bound is oftentimes maximized by a simplified surrogate objective \cite{ho2020ddpm_neurips}. After learning the parameters of the reverse process, we can sample $\boldsymbol{x}_T$ from $\mathcal{N}(\boldsymbol{x}_T;\mathbf{0},\mathbf{I})$ to start generating samples through an iterative denoising procedure.

\section{Details on \forl}
\subsection{Conditional Diffusion Model Details}\label{supp:sec:diffmodel}
Our aim is to learn the reverse diffusion process, by modeling $p_\theta\left(\s_t^{(n-1)} \mid \s_t^{(n)}, \boldsymbol{\tau}_{(t,w)}\right)$ as a Gaussian distribution $\mathcal{N}(\s^{(n-1)}_t
;\mu_\theta(\s^{(n)}_t,\boldsymbol{\tau}_{(t,w)}, n),\Sigma_\theta(\s^{(n)}_t,\boldsymbol{\tau}_{(t,w)}, n))$ where $n$ is the diffusion timestep and $t$ is the RL timestep. We  approximate this mean $\mu_\theta(\s^{(n)}_t,\boldsymbol{\tau}_{(t,w)}, n)$ using a conditional noise model $\eps_\theta$ with

$$
\frac{\s^{(n)}_t}{\sqrt{\alpha_{(n)}}} - \frac{1 - \alpha_{(n)}}{\sqrt{1 - \bar{\alpha}_{(n)}} \sqrt{\alpha_{(n)}}} \eps_\theta(\s^{(n)}_t,\boldsymbol{\tau}_{(t,w)}, n)
$$

and fix the covariance $\Sigma_\theta(\s^{(n)}_t,\boldsymbol{\tau}_{(t,w)}, n)$ with $(1-\alpha_{(n)})\mathbf{I}$ \cite{ho2020ddpm_neurips}.
Selecting a large number of diffusion timesteps can significantly increase the computational complexity of our algorithm. Hence, we use variance preserving stochastic differential equations (SDE) \cite{song2020score} following the formulation in \cite{xiao2021tackling} $\alpha_{(n)} = e^{-\left(\beta_{\text{min}}\left(\frac{1}{N}\right) + (\beta_{\text{max}} - \beta_{\text{min}}) \frac{2n-1}{2N^2}\right)}$ where $\beta_{\text{max}}=10$ and $\beta_{\text{min}}=0.1$. 
We use the noise prediction model \cite{ho2020ddpm_neurips}
with reverse diffusion chain
\begin{equation}
\s_t^{(n-1)} \mid \s_t^{(n)} =\frac{\s_t^{(n)}}{\sqrt{\alpha_{(n)}}} - \frac{1-\alpha_{(n)}}{\sqrt{\alpha_{(n)} (1 - \bar{\alpha}_{(n)})}} \eps_\theta(\s_t^{(n)}, \boldsymbol{\tau}_{(t,w)}, n) + \sqrt{1-\alpha_{(n)}} \eps
\end{equation}
where $\quad \eps \sim \mathcal{N}(0, \mathbf{I})$ for $n = N, \dots, 1$, and $\eps=0$ for $n=1$ \cite{ho2020ddpm_neurips}.

By using the conditional version of the simplified surrogate objective from \cite{ho2020ddpm_neurips} we minimize the \forl model loss $\mathcal{L}_p(\theta)$, with \cref{eq:porlloss} in the main paper.
We can train the noise prediction model by sampling $\s_t^{(n)}$ for any diffusion timestep in the forward diffusion process, utilizing reparametrization and recursion \cite{ho2020ddpm_neurips}. We use the true data sample $\boldsymbol{s_t}$ from the offline RL dataset to obtain the noisy state $\s_t^{(n)}=\sqrt{\bar{\alpha}_{(n)}} \boldsymbol{s_t} + \sqrt{1 - \bar{\alpha}_{(n)}} \eps$. Leveraging our model's capacity to learn multimodal distributions, we generate a set of $k$ samples (predicted state candidates) $\{\s_t^{(0)}\}$ in parallel from the reverse diffusion chain using \cref{eq:revchainsample}.

\begin{algorithm}[tb]
\caption{Training}
\label{alg:training_algorithm}
\textbf{Require}: Offline RL dataset $\mathcal{D}$\\
\textbf{Initialize}: $\eps_\theta$
\begin{algorithmic}[1]
\STATE \textbf{for} each iteration \textbf{do}
\STATE \hspace{1em} $\{\boldsymbol{(\tau}_{(t,w)},\s_t)\} \sim \mathcal{D}$
\STATE \hspace{1em} $n \sim \mathcal{U}(\{1,2, \ldots, N\})$
\STATE \hspace{1em} $\eps \sim \mathcal{N}(\mathbf{0}, \boldsymbol{I})$
\STATE \hspace{1em} Take gradient descent step on \\$\nabla_\theta\left\| \eps - \eps_\theta \left(\sqrt{\bar{\alpha}_{(n)}} \boldsymbol{s_t} + \sqrt{1 - \bar{\alpha}_{(n)}} \eps, \boldsymbol{\tau}_{(t,w)} , n \right) \right\|^2$
\STATE \textbf{end for}
\STATE \textbf{return} $\epsilon_\theta$
\end{algorithmic}
\end{algorithm}

\begin{figure*}[htbp]
	\centering
		\includegraphics[width=0.9\textwidth]{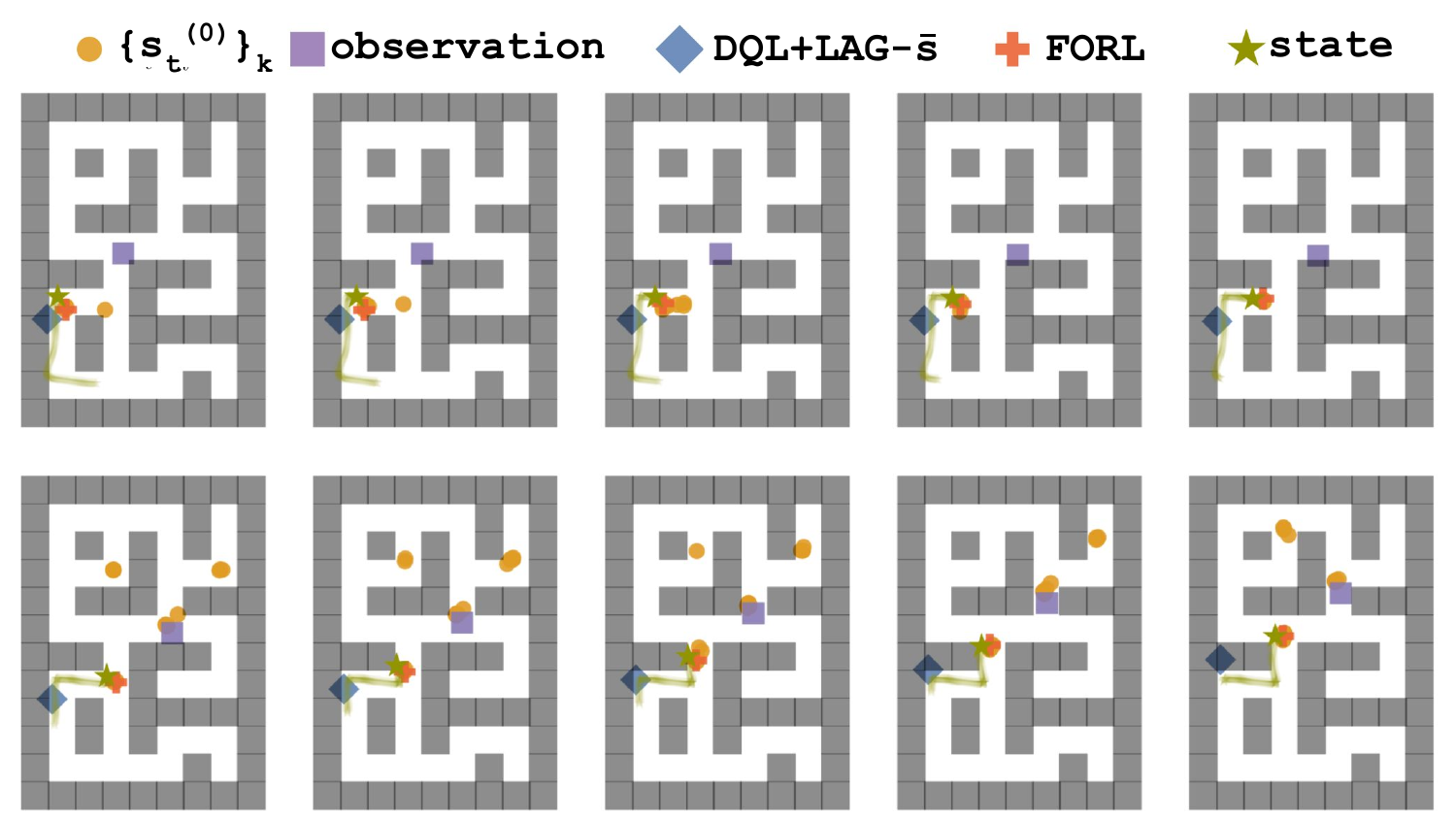}
        \caption{As the agent navigates in the \mazel environment \cite{fu2020d4rl}, illustrations of states; predicted states from \forl (ours); \forl diffusion model predictions $\{\s_{t}^{(0)}\}_k$; observations; and predicted states from \difqlmeanlag are shown. Environment timesteps progress from left to right.}
	\label{fig:mazepredfigs2}
\end{figure*}

\section{Baselines}
\label{app:baselines}
In this section we explain the baselines used in Table~\ref{tab:avgrewtable} and  Figure~\ref{fig:avgrewcomparison}. Diffusion Q-Learning (\difql) \cite{wang2023diffusion} serves as our base policy across all baselines and for our method in D4RL experiments \cite{fu2020d4rl}. Details on \difql are provided in Section~\ref{app:difql}

\subsection{\difql Extensions with \zeroshotfm}

For our baselines, we use a diffusion-based offline RL policy, \difql \cite{wang2023diffusion}, with three variations: 
\begin{itemize}
\item \textbf{\difql:} \difql policy directly generates actions conditioned on the observations. We include it to demonstrate how large episodic offsets degrade policy performance, and its consistently poor performance underscores the difficulty introduced by our offset settings.
\item \textbf{\difqlmeanlag:} We use \zeroshotfm (Lag-Llama \citep{rasul2023lag}) to forecast episodic offsets over a horizon of $P$ episodes using pre-deployment offset history, and then subtract the mean predicted offset corresponding to that episode from the observations. Thus, \difql generates actions conditioned on the sample mean of the forecasted states $\{\hat{s}^b_t\}_l$. Although \difqlmeanlag outperforms \difql, a straightforward application of forecasting and decision-making reveals performance degradation when offsets lack adaptive correction.
\item \textbf{\difqlmedlag:} We use the sample median instead of the sample mean in \difqlmeanlag, to mitigate outlier effects. As shown in Figure~\ref{fig:avgrewcomparison}, median-based bias removal yields no significant improvement over the mean-based approach.
\end{itemize}

\subsection{\dmbplag}
We extend the Diffusion Model–Based Predictor (\dmbp) \cite{zhihedmbp}, a model-based, robust offline RL method trained on unperturbed D4RL benchmarks, to correct test-time state observation perturbations. \dmbp has proven robust under Gaussian noise (varying $\sigma$), uniform noise, and adversarial perturbations--including maximum action‐difference (MAD) and minimum Q‐value (Min‐Q) attacks. Similar to our framework, perturbed states are passed to \dmbp, which removes the perturbations; the generated state is then fed to the policy. For a fair comparison, we adopt the policies also used in their work for our experiments, \rorl \citep{yang2022rorl} and \tdthreebc \citep{fujimoto2021a} alongside \difql.

\dmbplag extends \dmbp with the forecasting module in our experiments. At each timestep, we remove offset biases using the sample mean of forecasted states $\{\hat{s}^b_t\}_l$ (as in \difqlmeanlag) before \dmbp correction. \dmbplag thus sequentially applies forecast‐based offset compensation, followed by model based perturbation removal, and then queries the policy. \difql \citep{wang2023diffusion} remains the shared policy. A naive integration of \dmbp relies on initial ground‐truth states and underperforms.

For the evaluation of \dmbp we use the open source code and the suggested hyperparameters available in \cite{zhyang2024dmbpcode}. To improve the performance of \dmbp, we conducted evaluations at test-time and reported the best-performing results for a range of different diffusion timesteps. \dmbp requires the diffusion timesteps to be manually defined based on different noise scales and types. Hence, we use the diffusion timesteps across a range of noise scales $\{0.15, 0.25, 0.5\}$ for maze and $\{0.05, 0.1, 0.15, 0.25\}$ for kitchen environments and identify that the best performance requires different noise scales across environment datasets. In particular, we report the best performing noise scale of $0.15$ for the \mazem environment, $0.25$ for the \mazel environment, $0.5$ for the \antu environment, $0.15$ for the \antm environment, $0.25$ for the \antl environment, $0.05$ for \kitchc. 

\section{Offline Reinforcement Algorithms}
\label{app:offlinerl_policies_detailed}
\paragraph{\difql \cite{wang2023diffusion}}
\label{app:difql}
\difql \cite{wang2023diffusion} is an offline RL algorithm that uses policy regularization via a conditional diffusion model \cite{ho2020ddpm_neurips}. \citet{wang2023diffusion} shows that Gaussian policies lack the expressiveness needed to capture the possibly multimodal and skewed behavior policy in offline datasets, which, in turn, limits performance. To remedy this, \difql uses a conditional diffusion model for the behavior‑cloning term, shown as the first part of Equation \ref{eq:difql_policy}, based on a state‑conditioned version of the simplified Denoising Diffusion Probabilistic Models (DDPM) objective\cite{ho2020ddpm_neurips}. To steer action generation toward high‐reward regions, the policy improvement loss also includes Q‑value guidance, shown as the second term below \cite{wang2023diffusion}

\begin{equation}
\mathbb{E}_{\substack{n\sim\mathcal{U},\,(s,a)\sim\mathcal{D},\,\\\epsilon\sim\mathcal{N}(0,I)}}
\Bigl[\bigl\|\epsilon - \epsilon_{\phi}\!\bigl(\sqrt{\bar{\alpha}_{(n)}}\,a + \sqrt{1-\bar{\alpha}_{(n)}}\,\epsilon,\;s,\;n\bigr)\bigr\|^{2}\Bigr]
\;-\;\alpha\,\mathbb{E}_{s\sim\mathcal{D},\,a^{0}\sim\pi_{\phi}}\bigl[Q_{\psi}(s,a^{0})\bigr].
\label{eq:difql_policy}
\end{equation}

Here, a reverse diffusion process conditioned on state $s$, denoted by $\pi_\phi\left(\boldsymbol{a}\mid \boldsymbol{s}\right)$, represents the policy. The Q‑networks are trained using the double Q‑learning trick \cite{van2016deep} and Bellman operator minimization \cite{Fujimoto2018OffPolicyDR,DBLP:journals/corr/LillicrapHPHETS15}, with \cite{wang2023diffusion}

\begin{equation}
  \mathbb{E}_{(\boldsymbol{s}_t, \boldsymbol{a}_t, \boldsymbol{s}_{t+1}) \sim \mathcal{D}, \boldsymbol{a}^0_{t+1} \sim \pi_{\phi^{\prime}} }\left[\left\|\left(r(\boldsymbol{s}, \boldsymbol{a})+\gamma \min _{i=1,2} Q_{\psi_i^{\prime}}(\boldsymbol{s}_{t+1}, \boldsymbol{a}^0_{t+1})\right)-Q_{\psi_i}(\boldsymbol{s}_t, \boldsymbol{a}_t)\right\|^2\right] 
  \label{eq:difql_criticloss}
\end{equation}

where $a^0_{t+1}$ is sampled from the diffusion policy conditioned on $s_{t+1}$, and $Q_{\psi_i}$, $Q_{\psi'_i}$,$\phi^{\prime}$ denote the critic and target‐critic networks, target policy network, respectively.

\paragraph{\tdthreebc \cite{fujimoto2021a}}
\label{app:tdthreebc}
TD3BC \cite{fujimoto2021a} extends the Twin Delayed Deep Deterministic policy gradient algorithm (TD3) \cite{fujimoto2018td3} to the offline RL setup. \tdthreebc incorporates a behavior cloning regularization term, normalizes state features within the offline RL dataset, and scales the Q-function using a hyperparameter with an added normalization term. TD3BC is a straightforward yet effective method that is also computationally efficient. The results in \cref{tab:offline} and \cref{table:rorltd3bc_all} show that although extending \tdthreebc with the forecasting module (\tdthreebcmeanlag) and a combination of the forecasting module and a diffusion model-based predictor (\dmbplag-T) improve performance, \forl-T achieves better performance across a diverse range of non-stationarities.

\paragraph{\rorl \cite{yang2022rorl}}
\label{app:rorl}
RORL~\cite{yang2022rorl} addresses adversarial perturbations of the observation function by learning a conservative policy that aims to be robust to out‑of‑distribution (OOD) state and action pairs. To achieve this, it introduces a conservative smoothing mechanism that balances mitigating abrupt changes in the value function for proximate states and avoiding value overestimation in risky regions that are absent from the dataset. Concretely, RORL regularizes both the policy and the value function, leveraging bootstrapping Q-functions and conservative smoothing of the perturbed states. This formulation yields robust training under adversarial perturbations in the observation function while preserving strong performance even in unperturbed environments. However, although using RORL as our base policy improved performance in \mazed environments, \forl significantly outperforms both its naive usage, the extension with our forecasting module \rorlmeanlag, and \dmbplag-R in \cref{tab:offline} and \cref{table:rorltd3bc_all}. These results demonstrate that policies designed to be robust to sensor noise or adversarial attacks fail to cope with evolving observation functions that introduce non‑stationarity into the environment.

\paragraph{\iql \citep{iql_Kostrikov_2021}}
\label{app:iql} 
Implicit Q-Learning \cite{iql_Kostrikov_2021} first learns a value function by expectile loss and a Q-function by Mean Squared Error (MSE) Loss without using the policy and instead using actions from the dataset. In doing so, they avoid approximating the values of unseen actions. Then, it learns the policy using advantage weighted regression \cite{peters2007reinforcement,wang2018exponentially,peng2019advantage,nair2020awac} using the learned Q-function and value function. We use the open-source implementation of \iql from \cite{tarasov2022corl} which references the source \cite{gwthomas_iql_pytorch}. 
Results in \cref{tab:iqlperformance} show \forl-I outperforms the baselines \iql, \iqlmeanlag, and \dmbplag-I across all environments. 

\paragraph{\fql \citep{fql_park2025}}
\label{app:fql} 
Flow Q-learning (\fql) \cite{fql_park2025} is a recent offline RL policy that shows strong performance on the OGBench \cite{ogbench_park2025}. We use \fql for the offline RL environments in OGBench and adopt the hyperparameters from the open-source implementation \footnote{https://github.com/seohongpark/fql/}. Similar to \difql, which uses diffusion models, \fql can learn an expressive policy. \fql trains two policies: (i) a flow policy trained with flow matching on the offline RL dataset for behavior cloning conditioned on the state, and (ii) a one-step policy trained with a distillation loss using the flow policy \cite{liu2022flow,liu2023instaflow,frans2024one,ding2024dollar,li2024reward} and a critic loss. This approach avoids expensive backpropagation through time; thus, it is fast during inference and training. Results in \cref{tab:avgrewtable,tab:avgrew_og_dcmablation} show \forl (also referred to as \forl(\candidsel)) outperforms the baselines when all baselines use \fql policy.

\input{tables/otherpoliciestable}
\input{tables/iql_table}

\section{Scaling Offsets}
\begin{figure}[!t]
\centering
  \begin{adjustbox}{max width=\columnwidth,clip}
\includegraphics[width=\linewidth]{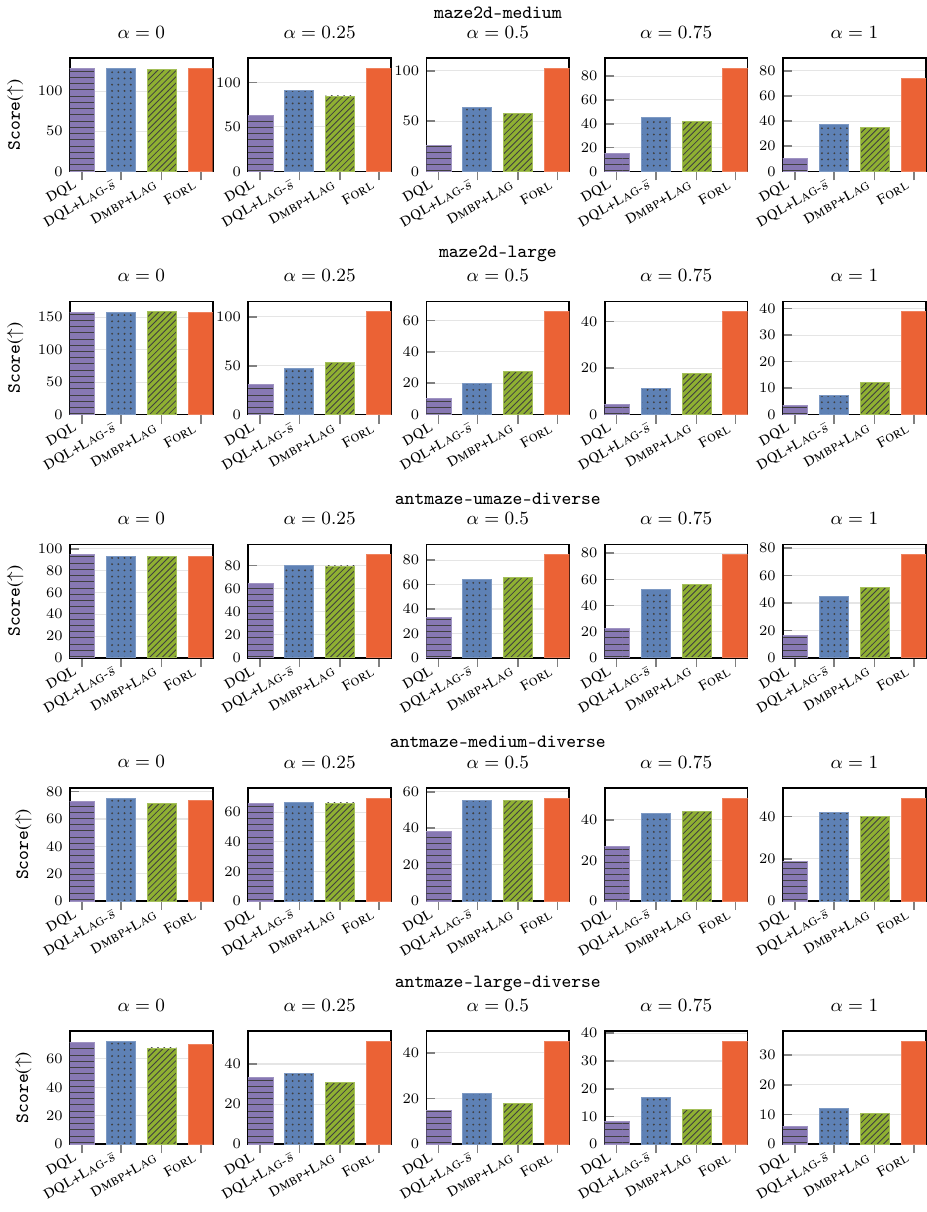}
\end{adjustbox}
\caption{\textbf{Average normalized scores of \forl (ours) and baselines across offset scaling factors $\alpha \in\{0, 0.25, 0.5, 0.75, 1\}$ in the \mazem, \mazel, \antu, \antm, \antl environments.} Scaling factor $\alpha=0$ corresponds to a stationary D4RL \cite{fu2020d4rl} test environment; $\alpha=1$ matches the original experimental configuration in Fig.~\ref{fig:avgrewcomparison}. Results are averaged over 5 non-stationarities (\aed, \electricity, \electricityhourly, \electricitynips, \exchangerate) and 5 random seeds.}

\label{fig:alpha-maze-comparison}
\end{figure}
We conduct an analysis to quantify \forl's sensitivity to different levels of non-stationarity. We scale the offset magnitude from 0 (standard D4RL offline RL environment \cite{fu2020d4rl} used in training) to 1 (our experiments) using scaling factors $\{0,0.25,0.5,0.75,1.0\}$, and report performance over five random seeds across five environments in \mazed and \antmaze in Figure~\ref{fig:alpha-maze-comparison}. This analysis delivers two key insights: first, \forl matches baseline performance on the stationary offline RL dataset on which it was trained; second, it maintains superior results throughout the full range of non-stationarity magnitudes. Crucially, \forl's performance degrades gracefully as the magnitude of offsets increases, whereas \difql (without forecasting) suffers steep drops. Furthermore, both \dmbplag and \difqlmeanlag decline in a similar manner. \dmbplag degrades slightly more gracefully than \difqlmeanlag in \mazel and \antu.

\section{What if we do not have access to past offsets?}
\label{app:dmablations}
\begin{figure}[!t]
    \centering
\includegraphics[width=\linewidth]{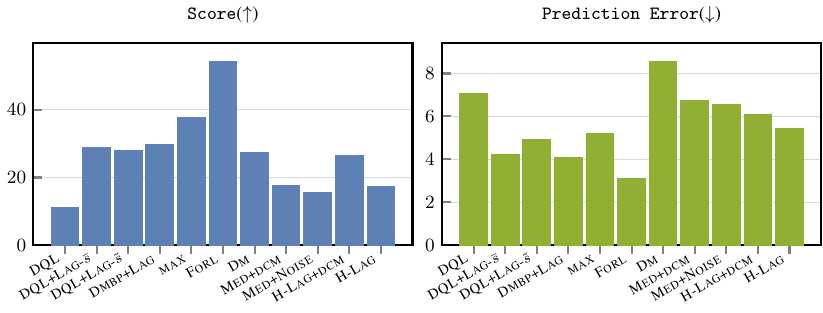}
\caption{
\footnotesize
\textbf{Comparison of average normalized scores and prediction errors across all 25 experiments in D4RL\cite{fu2020d4rl}} between \textbf{\forl~(ours)} and baseline methods, each evaluated over \textit{5 random seeds}. \textnormal{LAG} denotes the integration of a zero-shot time-series foundation model~\cite{rasul2023lag}. While \textbf{\forl} and \textbf{\maxlikelihood} also utilize this model, the subscripts are omitted for brevity.
}
\label{fig:avgrewcomparison}
\end{figure}

\begin{figure}[!t]
    \centering
\includegraphics[width=0.6\linewidth]{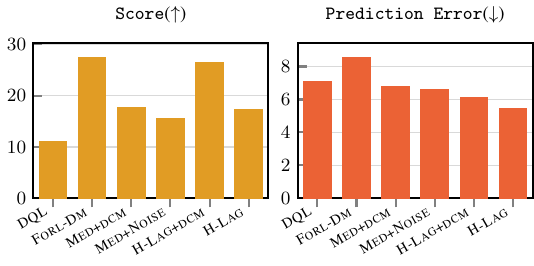}
\caption{
\textbf{Performance comparison without access to past offsets.} Average normalized scores and prediction errors for \textbf{\forl~(ours)} versus baselines, aggregated over 25 experiments (5 random seeds each)  in D4RL\cite{fu2020d4rl}. \textnormal{LAG} denotes the integration of a zero-shot time-series foundation model~\cite{rasul2023lag}. However, in this setting \zeroshotfm uses the samples from \dm instead of past offsets.}
\label{fig:avgrewcomparison_nopastoffsets}
\end{figure}

To analyze the challenging identifiability issue arising from (i) non‑smoothly varying offsets and (ii) the unobservability of ground truth offsets throughout the evaluation interval, we implement a set of methods for the setting where we never have delayed access to past ground truth offsets. \cref{fig:avgrewcomparison} shows the average normalized scores and prediction accuracies over 25 environment–non-stationarity pairs across five random seeds in navigation control tasks with continuous state and action spaces. Overall, these methods underperform compared to \forl. Among the cases with no access to past offsets (see \cref{fig:avgrewcomparison_nopastoffsets}), the best-performing methods are our proposed candidate state generation module (\forl-\dm) and \hlagdcm, a version of \forl that utilizes \zeroshotfm in addition to \dm which we detailed in \cref{subsec:noaccesspastoffsets_main}.

\paragraph{\forl-\dm (\dm)}  
\forl-\dm, which we also refer to as \dm for brevity, directly uses the state predicted by the diffusion model component. Given the multimodal nature of these candidate states, this selection does not fully leverage our framework. Yet, \dm outperforms the baselines when no past offsets are used. Additional comparisons in \cref{tab:forldmablation_maze2dmed} with other standard statistical methods like \dmbmod, \dmbransac, \dmbwelford, \dmbwelfordreset indicate that only using the sample predicted by the diffusion model yields higher scores on average. 

\paragraph{\meddcm} 
We compute the median of the predicted offsets from the previous episode, beginning with the first episode predicted by \forl's diffusion model. We then fit a Gaussian distribution centered at this median and sample $l$ offsets from it, matching the sample count of \textit{Zero-Shot FM}. Next, we apply \textbf{\candidsel} to these samples with the candidates generated by the diffusion model. 

\begin{figure}[H]
	\centering
		\includegraphics[width=0.6\textwidth]{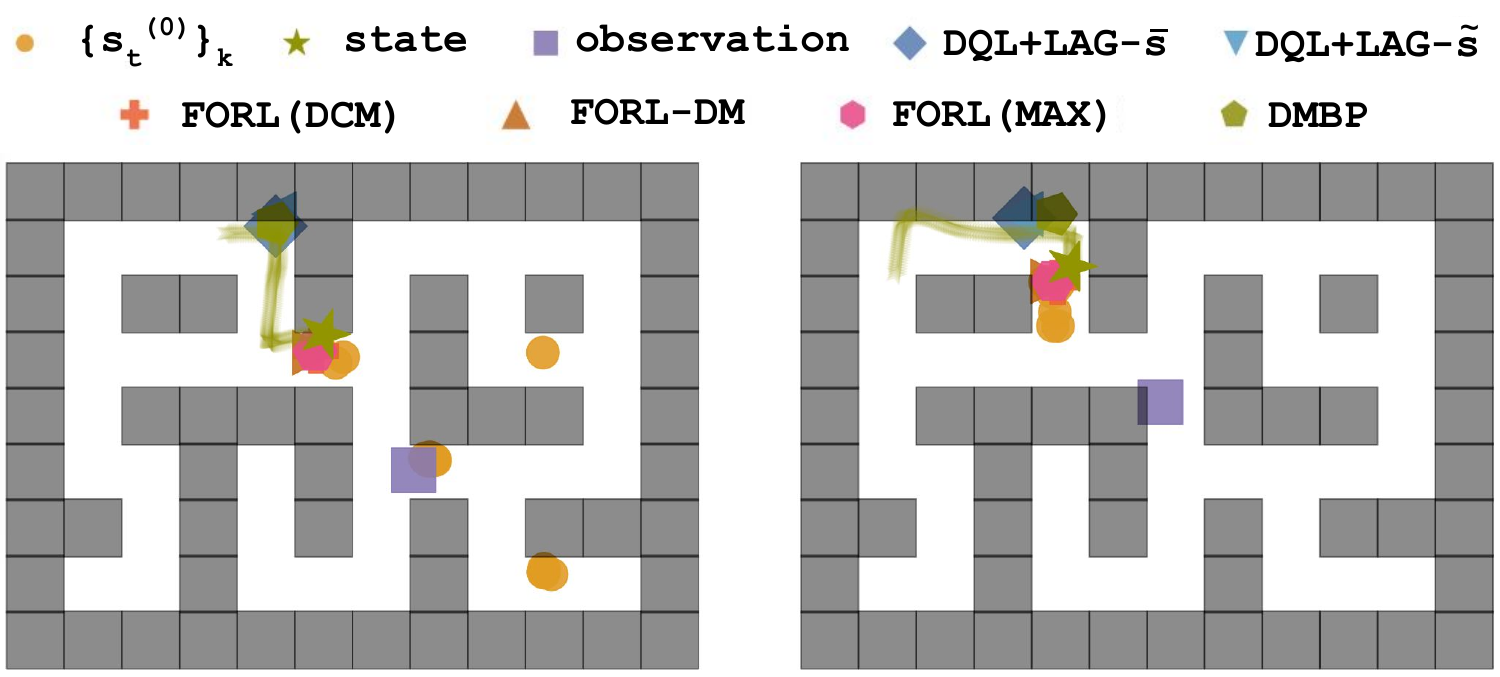}
 \caption{\textbf{Illustrations of states, observations, diffusion model predictions $\{\s_{t}^{(0)}\}_k$, and predicted states from  \forl(\candidsel), \forl-\dm, \difqlmeanlag, \difqlmedlag, \forl(\maxlikelihood), \dmbplag are shown.} These visualizations are from the same setting presented in \cref{fig:mazepredfigs2}. Maximum, minimum, and mean prediction errors across episodes for this task are provided in \cref{tab:dcm_max_minerrorcomparisonfig6}.}
	\label{fig:mazepredfigs2detailedallablations}
\end{figure}

\paragraph{\mednoise} 
We begin by computing the median offset produced by the diffusion model during the first evaluation episode similar to \meddcm. Thereafter, we treat these offsets as evolving according to a random walk, where each offset is predicted as the previous value with white noise increments. 

\paragraph{\dmbmod}
\dmbmod follows a state estimation based on robust statistics \cite{Robuststatistics_huber}. \dmbmod computes the coordinate-wise median of the differences between the observation and each denoiser prediction and discards any sample whose absolute deviation from this median exceeds $\epsilon$ x Median Absolute Deviation (MAD). Then, it takes the median of the remaining inliers to obtain a robust offset estimate and subtracts that offset from the observation to obtain the state $\tilde{s_t}$.

\paragraph{\dmbransac}
\dmbransac is a RANdom SAmple Consensus (RANSAC) \cite{ransac_Fischler_1981} based state estimation using the samples from our \dm. \dmbransac calculates the offsets using the states generated by our \dm and the observation, setting an adaptive per-dimension threshold as $\epsilon$ x MAD. Then, it repeatedly samples a random offset candidate and chooses the candidate whose inlier set (differences within that threshold) is the largest. Then, it takes the average of those inliers to estimate the offset.

\paragraph{\dmbwelford}
\dmbwelford computes a numerically stable, global running mean \cite{Welford1962NoteOA,knuth1997art} of offsets from \dm aggregated across all timesteps and episodes.

\paragraph{\dmbwelfordreset}
\dmbwelfordreset computes an online average of offsets \cite{Welford1962NoteOA,knuth1997art} from \dm (Running-$\mu$) per episode $p$. Unlike \dmbwelford, in \dmbwelfordreset the statistics are reset to zero at the beginning of each episode.

\input{tables/avgrew_dm_ablation}

\input{tables/avgrewtable_og_significant}

\section{Candidate Selection} 

The results in \cref{tab:dcm_ablation_table}, \cref{tab:dcm_max_minerrorcomparisonfig6}, and \cref{tab:avgrew_og_dcmablation} show that \forl(\candidsel) more consistently performs better than other methods. Although the KDE-based \scott method performs well in \antm, it significantly underperforms in \ogcube (\cref{tab:avgrew_og_dcmablation}). Moreover, it requires bandwidth selection and a fallback mechanism to handle numerical instability, which highlights the practical advantage of \candidsel.

\forl(\maxlikelihood), also referred to as \maxlikelihood for brevity, can fail when the forecast mean of $D_{timeseries}$ is biased, misleading it to select a candidate from a geometrically distant mode that appears more likely under an inaccurate forecast. In contrast, \candidsel succeeds because its state estimation is not dependent on the forecast's mean, but on a non-parametric search for the forecast sample with the highest score (minimum dimension-wise distance). Hence, \candidsel's prediction error is governed by the accuracy of the forecast sample in  $D_{timeseries}$ with the best score. Empirically, this yields lower maximum and mean errors compared to \maxlikelihood. The timeseries forecaster can generate a large set of samples that can be systematically biased, which is why we observe that \difqlmeanlag and \difqlmedlag also have high maximum prediction errors in \cref{tab:dcm_max_minerrorcomparisonfig6}. While for this specific setting in \cref{fig:mazepredfigs2detailedallablations}, \forl(\maxlikelihood) and \forl-\dm are close to \forl(\candidsel), although worse, we observe that across the test episodes, \forl(\candidsel) outperforms \forl(\maxlikelihood) and \forl-\dm in terms of maximum and mean error, demonstrating stability.

\input{tables/dcmablation_maxminerrorfig6}

\input{tables/avgrew_dcm_ablation_table}

\input{tables/avgrew_intraepisode}

\subsection{Sensitivity to Diffusion-generated Sample Size}

\begin{figure}[!t]
    \centering
\includegraphics[width=0.25\linewidth]{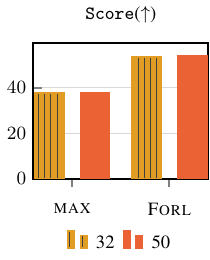}
\caption{\label{fig:numdentik}\textbf{Average normalized scores over 25 experiments with diffusion-generated sample sizes of 32 and 50, each conducted with 5 random seeds.} The results indicate that the diffusion model's performance remains consistent across varying sample sizes, demonstrating robustness to the number of candidate states generated.}
\end{figure}

\maxlikelihood uses candidate states predicted by the diffusion model in the \forl framework and Lag-Llama but uses maximum likelihood instead of \candidsel. Given the multimodal nature of the candidate state distributions, we conduct a sensitivity analysis on the number of denoiser samples, a shared hyperparameter for both $\maxlikelihood$ and $\forl$. We report results averaged over 5 random seeds across 25 tasks (\antmaze and \mazed with \texttt{real-data-A,B,C,D,E}). The results in \cref{fig:numdentik} indicate that our diffusion model's performance remains consistent across varying sample sizes, demonstrating robustness to the number of candidate states generated. Notably, the \dmbp algorithm uses 50 denoiser samples.

\section{Zero-shot Foundation Model and Time-Series Datasets}

\begin{figure*}
	\centering
		\includegraphics[width=\textwidth]{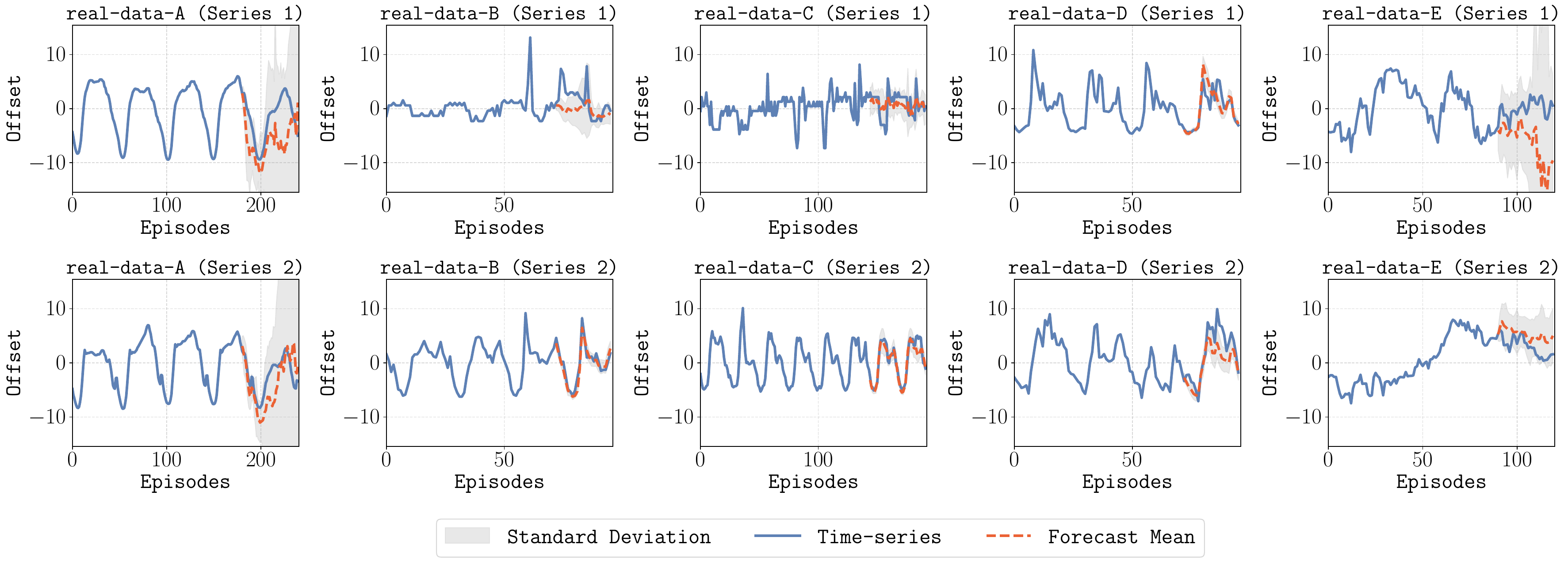}
        \caption{Zero-shot forecasting results with Lag-Llama \cite{rasul2023lag} for first 2 time-series in univariate time-series datasets: \aed (Australian-electricity-demand), \exchangerate (Exchange Rate), \electricity (Electricity), \electricityhourly (Electricity Hourly),\electricitynips (Electricity Nips).}
	\label{fig:tsdatasetforecasts-all}
\end{figure*}

We extract the first two univariate series from five time‐series datasets: \aed (\texttt{australian-electricity-demand}) \cite{godahewa2021monash}, \electricity (\texttt{electricity})\cite{Dheeru2017}, \electricityhourly (\texttt{electricity-hourly}) \cite{godahewa2021monash,Dheeru2017}, \electricitynips (\texttt{electricity-nips}) \cite{Dheeru2017,salinas2019high} and \exchangerate (\texttt{exchange-rate}\footnote{\url{https://github.com/laiguokun/multivariate-time-series-data}}) \cite{lai2018modeling}, \cite{godahewa2021monash} all accessed via GluonTS \cite{gluonts_jmlr,gluonts_arxiv}. Figure~\ref{fig:tsdatasetforecasts} presents the ground truth, forecast mean, and standard deviation from Lag-Llama \cite{rasul2023lag} for the first series of \aed and \electricitynips; forecasts for the remaining series and domains are provided in Figure~\ref{fig:tsdatasetforecasts-all}. 

We aim to capture a broad spectrum of scenarios for a comprehensive evaluation. For instance, the \texttt{electricity-hourly} dataset consists of hourly electricity usage data from various consumers, while the \texttt{australian-electricity-demand} dataset has 30-minute interval records of electricity demand across different Australian states. The \texttt{exchange-rate} dataset, on the other hand, includes daily exchange rates of multiple currencies, including those of Australia, the United Kingdom, Canada, Switzerland, China, Japan, New Zealand, and Singapore.

To effectively represent diverse offset patterns in multiple directions, we apply feature scaling to the time-series data using a normalization $x_c' = \frac{x - \bar{x}}{\text{max}(x) - \text{min}(x)}$ where sample mean $\bar{x}$, $\text{min}(x)$ and $\text{max}(x)$ are computed from the available data up to the context length. Furthermore, we scale these values using the minimum and maximum state values observed in the offline RL dataset \cite{fu2020d4rl} for navigation and minimum and maximum state values of the initial state distribution at test time for manipulation \cite{fu2020d4rl}, ensuring that the experiments cover diverse observation spaces that can accurately represent a wide range of scenarios. We group our results in terms of time-series datasets in \cref{tab:timeseriescategorized_avgrewtable}.

\subsection{Sensitivity of \forl to Forecasting Errors}
\label{sec:sensitivity_to_forecasting_errors}
To analyze the sensitivity of \forl to forecasting errors, we compare its performance against \difqlmeanlag, which only uses the forecaster's predictions. \cref{tab:sensitivity-error-reduction} presents the average prediction error ($\downarrow$) across all datasets in \antmaze and \mazed environments with 5 seeds.

\input{tables/error_reductiontable}
In all datasets, \forl outperforms the \difqlmeanlag baseline. \forl achieves its greatest impact on moderately challenging forecasts (a 56.5\% error reduction on \electricitynips). Its behavior at the extremes further demonstrates its robustness:

\begin{itemize}
    \item \forl still refines the best forecast by 10.2\% (\electricityhourly)
    \item \forl improves the worst forecast by 4.3\% (\exchangerate).
\end{itemize}

\subsection{State Prediction Accuracy}\label{app:pred-accur}
To evaluate the state‐prediction accuracy of our \forl framework, we compare it against \difqlmeanlag. For each method, we report the mean $\ell_2$ error between the true state $s_t$ and the predicted state $\tilde{s}_t$ obtained during evaluation for a diffusion-generated sample size of 32. 

In the resulting average prediction error table in \cref{app:tab:allcompareepredacctable}:
\begin{itemize}
\item Each \textbf{row} corresponds to the state estimation algorithm used at test time to generate states $ \tilde{s}_t $, which are then provided to the policy to select actions.
\item Each \textbf{column} corresponds to a method whose state estimates are evaluated on that same rollout.
\end{itemize}
The entry at row $i$, column $j$ is the mean $\ell_2$ error when method $m_i$ is used in the environment, but predictions are produced by method $m_j$. When $i = j$, this entry measures the self‐prediction error of each method; when $i \neq j$, it measures the error under an alternate method. 

Across all method pairs, \forl achieves lower mean $\ell_2$ errors, even in off‐diagonal evaluations, demonstrating its superior state‐prediction performance compared to \difqlmeanlag. These findings are consistent with the normalized environment scores in \cref{tab:avgrewtable}.
\input{tables/allcompareepredacctable}
\input{tables/tsdatasetavgrew}

\section{Preliminary Results for Affine Transformation with Uniform Scaling and Bias}
\label{sec:supplementary_affine}

We use the fourth series in each time-series domain to perform isotropic scaling for the dimensions affected by non-stationarity using a scaling factor of $\beta = 0.5$, with bias coming from the first two series, respectively. We apply feature scaling to time-series data with $x_c' = 1 - \beta + \beta \cdot \exp\left(\frac{x - \bar{x}}{2 \cdot (\text{max}(x) - \text{min}(x))}\right)$. The offset scaling for the bias uses $\alpha = 1$, which is the standard value in our experiments. As in the other ablations with scaling offsets, we use the \difql policy. \cref{tab:avgrewtable_affine} shows that \forl outperforms the baselines under this transformation. A large-scale analysis of more general transformations is left for future work.

\input{tables/avgrewtable_affine}

\section{Offline Reinforcement Learning Environments}
\subsection{D4RL}
\label{subsec:d4rl}
\begin{figure}[H]
 \vspace*{1mm}
	\centering
		\includegraphics[width=0.194\columnwidth]{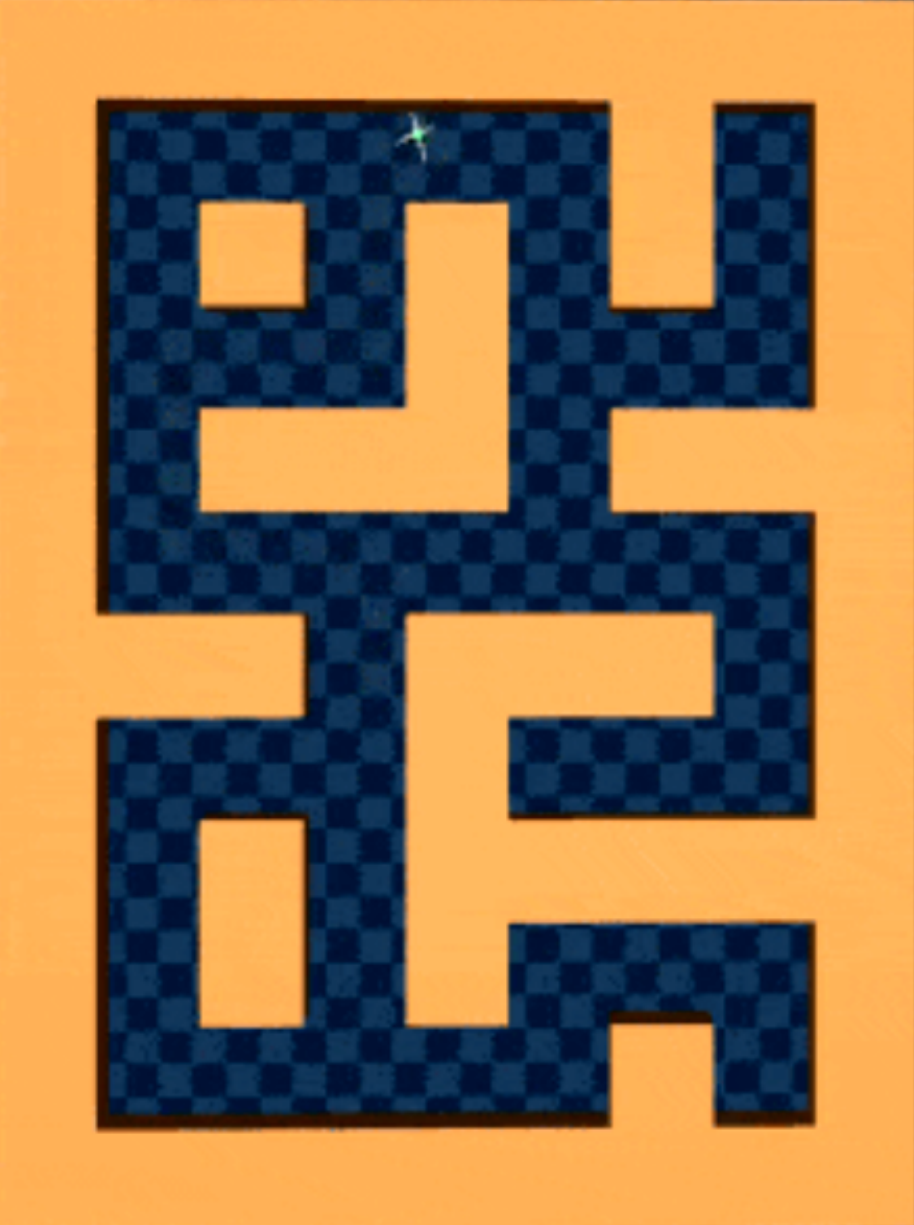}
	       \includegraphics[width=0.26\columnwidth]{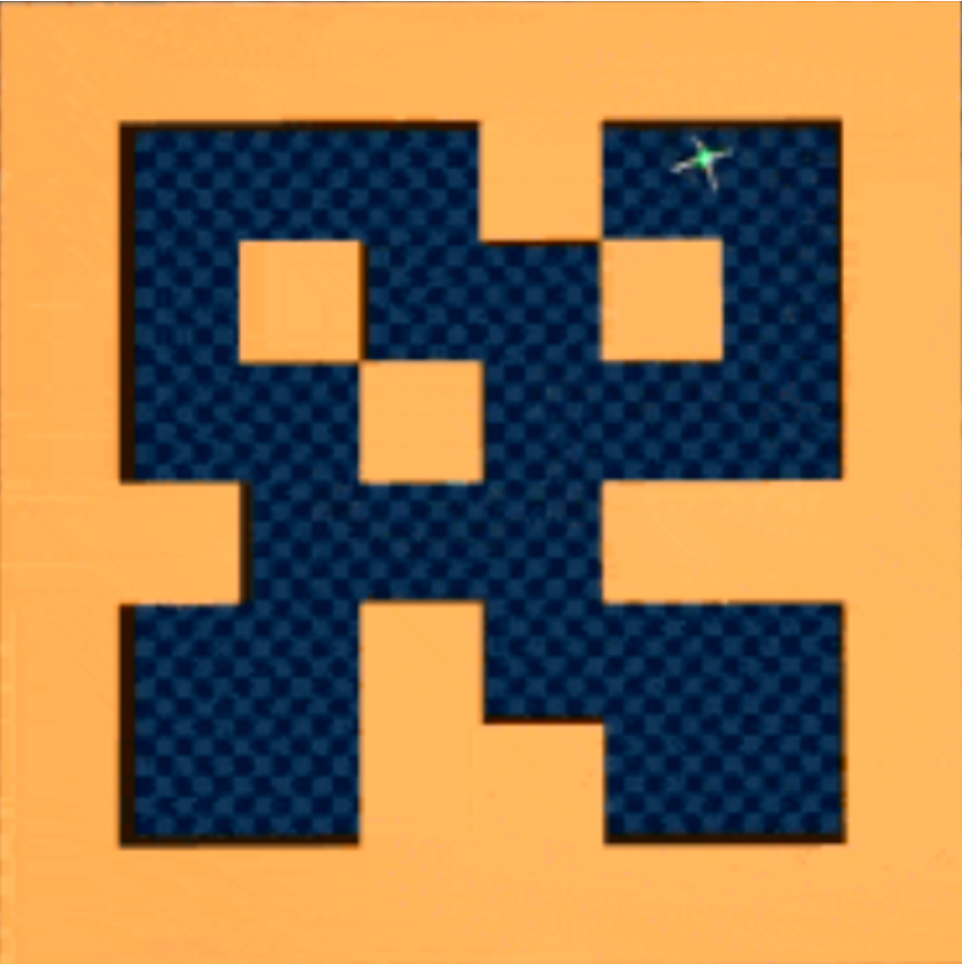}
		\includegraphics[width=0.26\columnwidth]{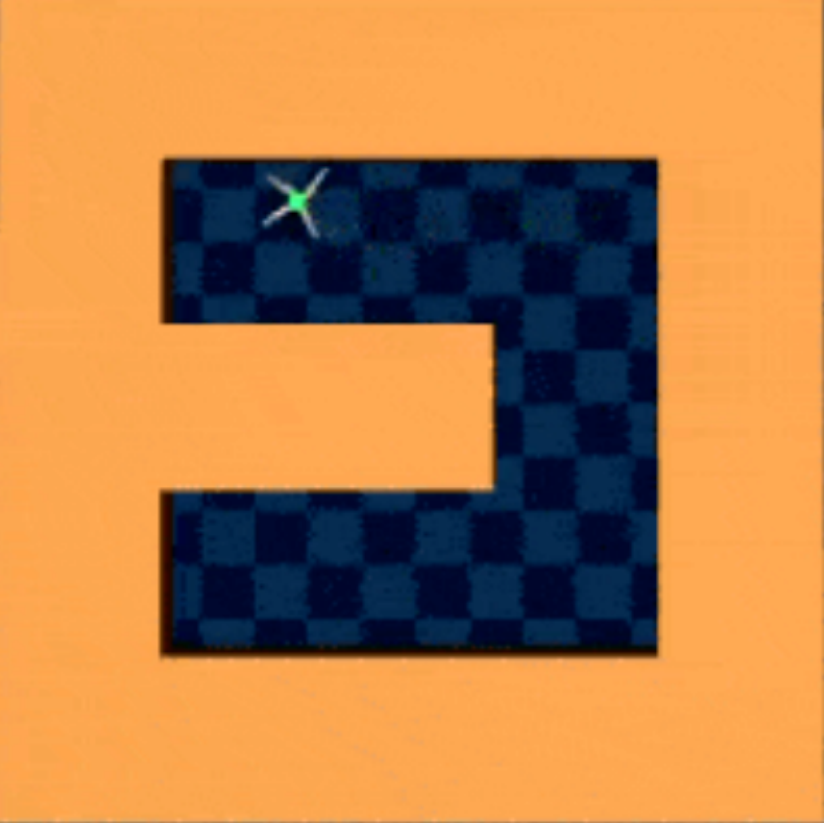}
        \includegraphics[width=0.2625\columnwidth]{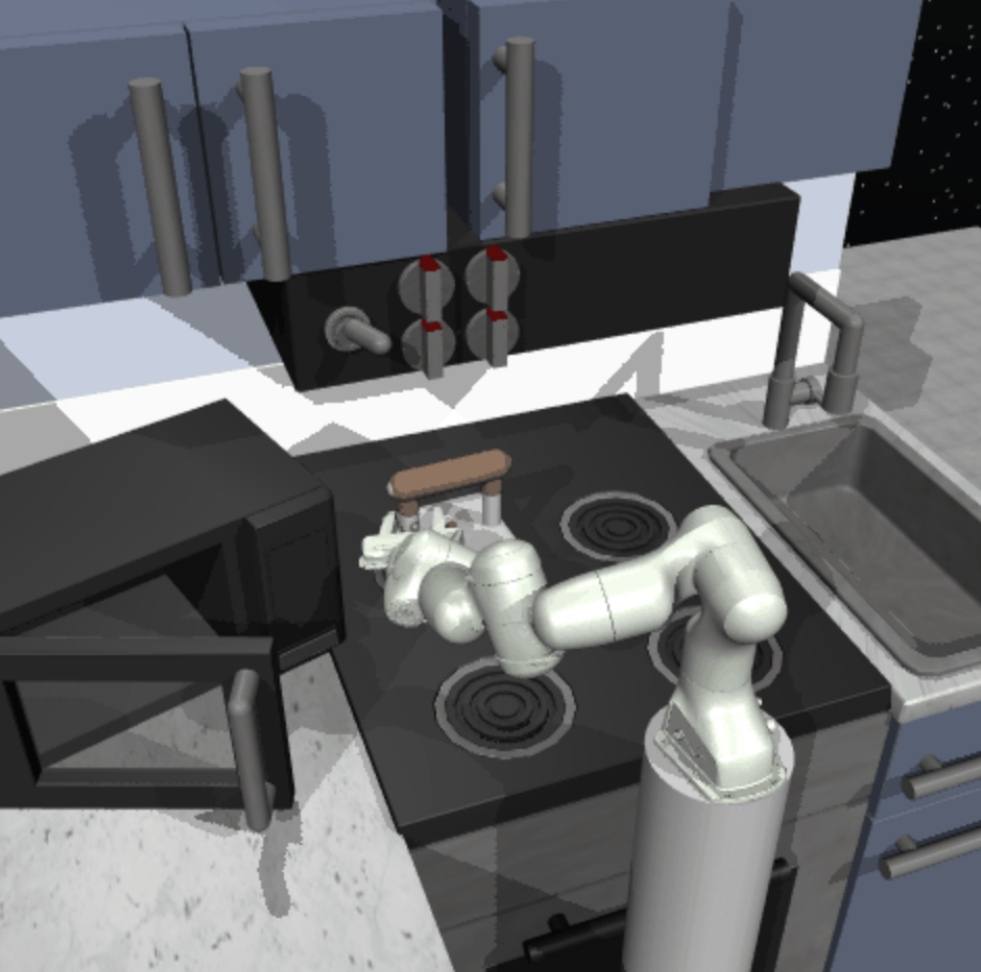}
        \caption{$\texttt{antmaze-large}$, $\texttt{antmaze-medium}$, $\texttt{antmaze-umaze}$ (-v1) and \kitchc environments in D4RL benchmark \cite{fu2020d4rl}.}
	\label{fig:antmazed4rl}
\end{figure}

We use the standard D4RL \cite{fu2020d4rl} offline RL environments \cite{fu2020d4rl} with no modifications during training, namely \antm, \mazem, \antl, \mazel, and \antu, where initial states are randomized both in the evaluation environment and in the offline dataset. \cref{fig:antmazed4rl} illustrates the environments used from the D4RL benchmark, \kitchc, \antl, \antm, \antu. The \mazel and \mazem environments share the same maze configurations as \antl and \antm, respectively. 

For manipulation tasks, we train on the standard \kitchc environment. We sample the base joint angles from $U([-1.5,0.17])$ and the shoulder‐joint angles from $U([-1.78,-1.16])$ which are set based on the minimum and maximum state space dimension intervals in the offline RL dataset to evaluate partial identifiability at test-time. The offsets affect the state dimensions associated with the base and shoulder joint angles.

\subsection{OGBench}
\label{subsec:ogbench}
\begin{figure}[htbp]
 \vspace*{1mm}
	\centering
        \includegraphics[width=0.26\columnwidth]{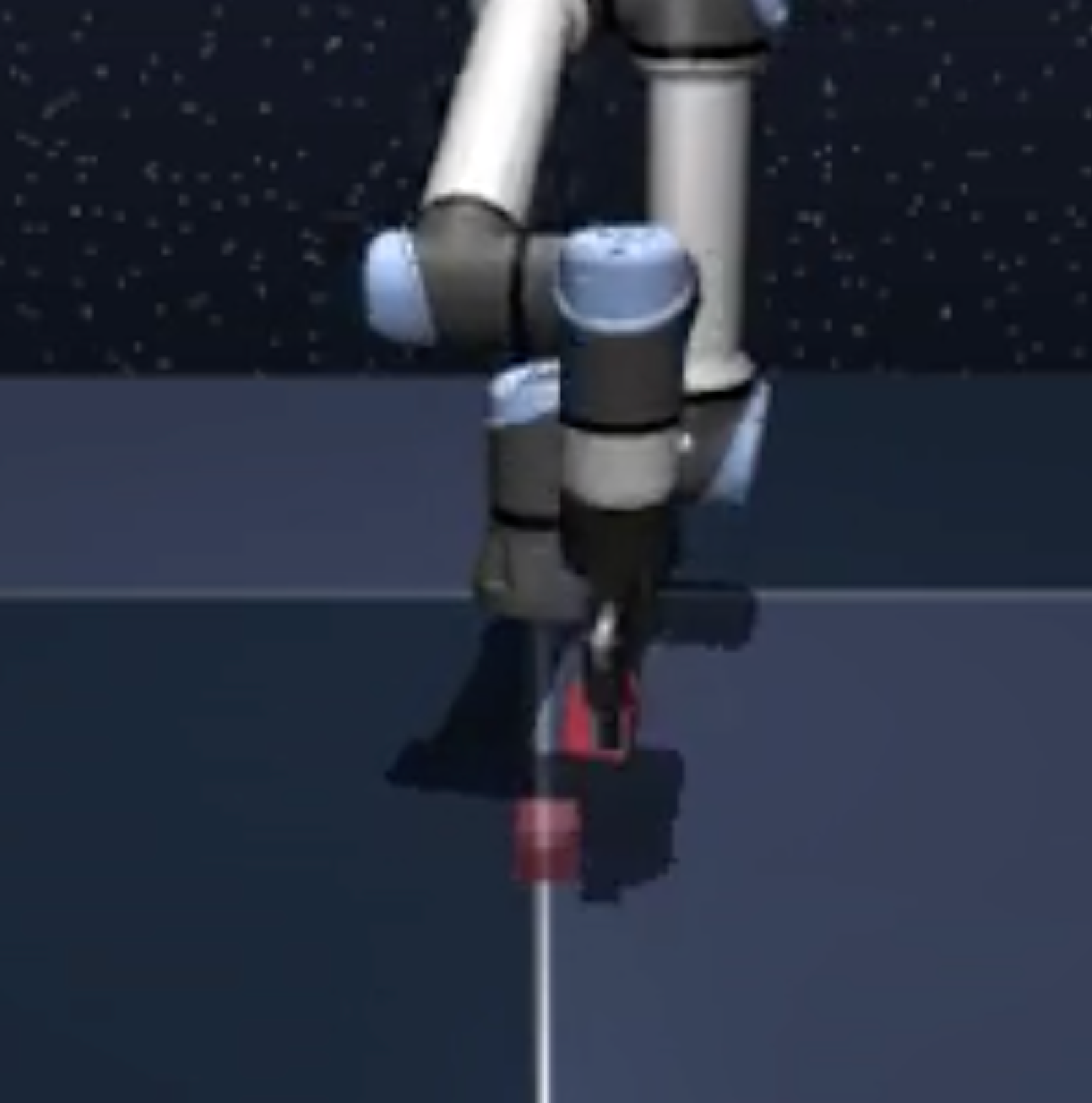}
        \includegraphics[width=0.4\columnwidth]{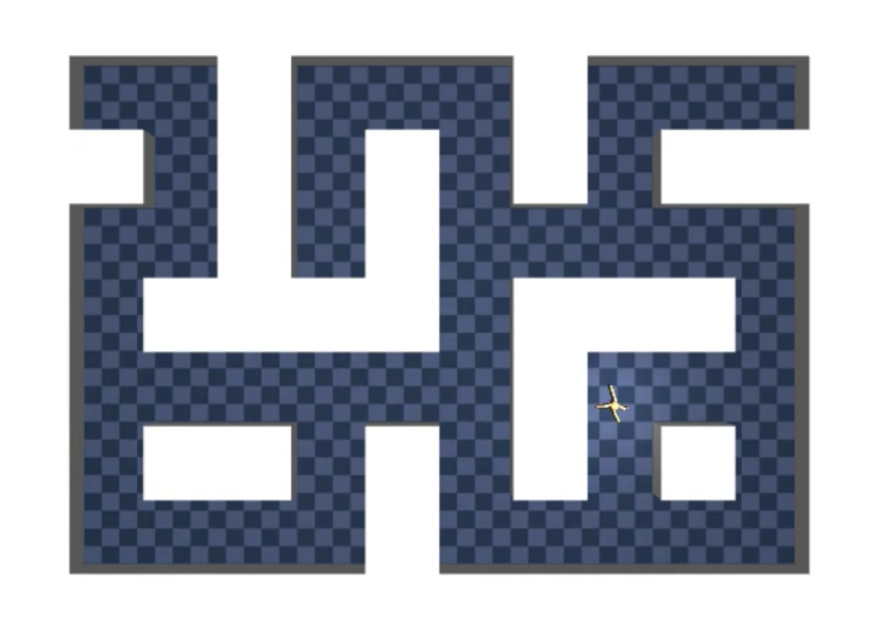}
        \caption{\ogcube and \ogantl environments in OGBench benchmark \cite{ogbench_park2025}.}
	\label{fig:ogbench_env_imgs}
\end{figure}

OGBench benchmark \cite{ogbench_park2025} contains both standard and goal-conditioned offline reinforcement learning tasks. To induce non-stationarity at test time, we follow the procedure from our D4RL experiments and use time-series data from GluonTS \cite{gluonts_arxiv,gluonts_jmlr}. For all tasks in \cref{fig:ogbench_env_imgs}, we use the default \texttt{singletask-v0} variant. We report results using the \fql algorithm \cite{fql_park2025} with its officially recommended hyperparameters. For the \ogantl environment, we use the first two time series from the GluonTS \texttt{real-data-A,B,C,D,E} datasets and apply an offset scaling factor of $\alpha = 0.5$. For \ogcube, we apply offsets to the first 17 observation dimensions, which include all joint positions, joint velocities, and end effector variables (position and yaw), using the first 17 time series from each of the GluonTS \texttt{real-data-A,B,C,D,E} datasets. Because the \aed dataset only has five time series, we cycle through them repeatedly until all 17 dimensions are covered. Across all \ogcube experiments, we use an offset scaling factor of $\alpha = 0.25$.
\begin{figure}[!t]
\centering
  \begin{adjustbox}{max width=\columnwidth,clip}
\includegraphics[width=0.7\linewidth]{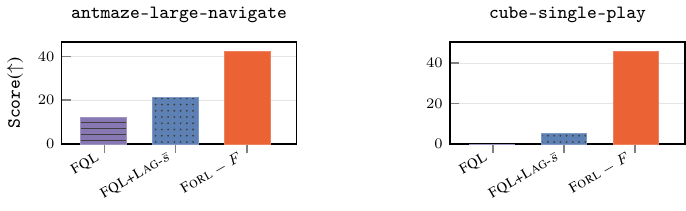}
\end{adjustbox}
\caption{\textbf{Average normalized scores of \forl (ours) and baselines for OGBench}}
\label{fig:ogbenchavgrew}
\end{figure}

\section{Implementation Details}
\label{sec:supplementary_implementation}

During training, we use the original offline RL dataset \textbf{without offsets}. At test-time, the offsets affect the first two state dimensions, where each offset sequence is drawn from the first two univariate time-series from diverse datasets. The agent's policy receives only the offset‐corrupted observations, with no direct access to the true underlying states throughout $P$ episodes. The time-series forecasting model, given the past $C$ ground-truth offsets $(b^{j-C},\cdots,b^{j-1})$, predicts the future offsets $(b^{j},\cdots,b^{j+P})$ during testing. \forl leverages these predictions and in-episode experience to dynamically adapt to unknown external perturbations. All experiments use 5 random seeds, except for (i) the preliminary affine-transformation results (\cref{sec:supplementary_affine}) and (ii) the focused error analysis across all evaluation episodes for the task shown in \cref{fig:mazepredfigs2detailedallablations} with results reported in \cref{tab:dcm_max_minerrorcomparisonfig6}.

We select the hyperparameters in Table \ref{tab:combined_hyperparameters} based on the validation loss of the \forl diffusion model in D4RL and OGbench standard offline RL datasets. The validation loss is computed using the \dm loss function in Eq. \ref{eq:porlloss}. Hyperparameter optimization was conducted using a grid search, with the following ranges for \mazed and \antmaze: diffusion timesteps $N=\{10,20\}$, number of hidden layers following the Temporal Unet model $\#layers=\{1,2,3\}$, window size $w=\{128,256\}$, and learning rate $lr=\{0.0004,0.0006,0.0009\}$. For the \kitchc and \ogcube $\textit{embedding dimension}=\{64,128\}$, learning rates $lr=\{0.0004,0.0009\}$ and $w=\{32,64\}$ are used for the grid search.

The architecture of our \forl Model is a noise prediction conditional TemporalUnet diffusion model \cite{ronneberger2015u,janner2022diffuser,zhihedmbp}. Different from the architecture used in \cite{zhihedmbp}, we concatenate each element in $\boldsymbol{\tau}_{(t,w)}$ with $\s_t^{(n)}$ and feed it to our model without additional encoders, using the diffusion timestep embedding in the Residual Temporal Blocks. For the TemporalUnet architecture, we concatenate the Unet model output with the time-embedding before feeding it to fully connected layers, particularly in the antmaze environments due to the large input size. The set of hyperparameters is provided in Table \ref{tab:combined_hyperparameters}. Although the \forl conditional diffusion model is specifically utilized for time-dependent offsets in the first two dimensions of the state vector, it is trained for general-purpose state prediction, enabling it to predict all dimensions of the state in \mazed, \antmaze, and OGBench environments. This approach is taken because we do not assume prior knowledge of the evaluation environment. 

The method for setting seeds involves a function that initializes the seed across all relevant libraries (PyTorch, CUDA, NumPy, Gym Environment, and Python's random module) to ensure the replicability of results. We use the open source implementation of \dmbp\footnote{\url{https://github.com/zhyang2226/DMBP/tree/main}} with the suggested hyperparameters \cite{zhyang2024dmbpcode}, and the pretrained Lag-Llama\footnote{\url{https://github.com/time-series-foundation-models/Lag-Llama}} model\cite{rasul2023lag}.

\section{Experiments compute resources}

Experiments were primarily conducted on an HPC cluster with NVIDIA A100 GPUs (40GB HBM2, PCIe 4.0/NVLink interconnect) and AMD EPYC 7302 CPUs (32 cores, 1TB RAM, 3TB local SSD), as well as on a workstation with an NVIDIA GeForce RTX 4090 (24GB GDDR6X), 128GB RAM, and a 2TB PCIe 4.0 NVMe SSD. A small portion of the experiments also ran on a cluster equipped with 4x NVIDIA V100 GPUs (16GB NVLink), 2x Intel Xeon Gold 6248R CPUs, and 384GB RAM. The total compute for published results is approximately 7,300 GPU-hours; additional failed and preliminary runs total approximately 1,500 GPU-hours. 

\input{tables/policy_hyperparams}

\section{Licenses for Existing Assets and Libraries}

\subsection{Existing Assets}

\begin{itemize}
    \item The \textbf{D4RL}\cite{fu2020d4rl}, including the Franka~Kitchen
          tasks, are distributed under the Creative Commons
          Attribution~4.0 (data) and Apache~2.0 (code) licenses as in
          \url{https://github.com/Farama-Foundation/D4RL}.%
    \item \textbf{MuJoCo}\cite{todorov2012mujoco} is released under the Apache~2.0
          license as indicated in
          \url{https://github.com/google-deepmind/mujoco/blob/main/LICENSE}.%
    \item \textbf{Gymnasium} (formerly OpenAI~Gym) is distributed under the MIT
          license as indicated in
          \url{https://github.com/Farama-Foundation/Gymnasium/blob/main/LICENSE}.%
    \item The \textbf{\electricity} dataset (UCI ``Electricity Load Diagrams
          2011-2014'')\cite{electricityloaddiagrams20112014_321} is distributed under the Creative Commons Attribution 4.0 International
          license as indicated in
          \url{https://archive.ics.uci.edu/ml/datasets/electricityloaddiagrams20112014}.%
    \item The \textbf{\electricitynips} and
          \textbf{\electricityhourly} variants are derived from the same UCI
          data and inherit the CC-BY-4.0 license.%
    \item The \textbf{\aed} dataset\footnote{Half-hourly
          demand for five Australian states} \cite{godahewa2021monash} is distributed under the
          Creative Commons Attribution 4.0 International license as indicated in
          \url{https://doi.org/10.5281/zenodo.4659727}.%
    \item The \textbf{\exchangerate} dataset introduced by
          \citet{lai2018modeling} with publicly available
          financial data; no explicit license is provided in the original
          repository (\url{https://github.com/laiguokun/multivariate-time-series-data}), and it is used in \cite{godahewa2021monash}, which distributes its datasets under the Creative Commons Attribution 4.0 International license.%
\end{itemize}

\subsection{Libraries}

The libraries used in our experiments are:

\begin{enumerate}
    \item \texttt{diffuser} uses the MIT License.\footnote{\url{https://github.com/jannerm/diffuser/blob/master/LICENSE}}
    \item \texttt{einops} uses the MIT License.\footnote{\url{https://github.com/arogozhnikov/einops/blob/main/LICENSE}}
    \item \texttt{imageio} uses the BSD 2-Clause License.\footnote{\url{https://github.com/imageio/imageio/blob/master/LICENSE}}
    \item \texttt{loguru} uses the MIT License.\footnote{\url{https://github.com/Delgan/loguru/blob/master/LICENSE}}
    \item \texttt{matplotlib} \cite{hunter2007matplotlib} uses a PSF-based license.\footnote{\url{https://github.com/matplotlib/matplotlib/blob/master/LICENSE/LICENSE}}
    \item \texttt{mujoco\_py} uses the MIT License.\footnote{\url{https://github.com/openai/mujoco-py/blob/master/LICENSE}}
    \item \texttt{numpy} uses the BSD 3-Clause License.\footnote{\url{https://numpy.org/doc/stable/license.html}}
    \item \texttt{pandas} uses the BSD 3-Clause License.\footnote{\url{https://github.com/pandas-dev/pandas/blob/main/LICENSE}}
    \item \texttt{scikit-video} uses the BSD 3-Clause License.\footnote{\url{https://github.com/scikit-video/scikit-video/blob/master/LICENSE.txt}}
    \item \texttt{torch} (PyTorch)) is distributed under a permissive, BSD-style license that includes an express patent grant. \footnote{\url{https://github.com/pytorch/pytorch/blob/main/LICENSE}}
    \item \texttt{tqdm} is licensed under MIT and MPL-2.0.\footnote{\url{https://github.com/tqdm/tqdm/blob/master/LICENCE}}
    \item \texttt{ogbench} uses the MIT License.\footnote{\url{https://github.com/seohongpark/ogbench/blob/master/LICENSE}}
\end{enumerate}

\end{document}

%% file: header.tex
\usepackage{xspace}

\makeatletter
\DeclareRobustCommand\onedot{\futurelet\@let@token\@onedot}
\def\@onedot{\ifx\@let@token.\else.\null\fi\xspace}
\makeatother

\newcommand{\ie}{i.e\onedot}

\usepackage{xcolor}
\definecolor{ourblue}{rgb}{0.368,0.507,0.71}
\definecolor{ourorange}{rgb}{0.881,0.611,0.142}
\definecolor{ourgreen}{rgb}{0.56,0.692,0.195}
\definecolor{ourred}{rgb}{0.923,0.386,0.209}
\definecolor{ourviolet}{rgb}{0.528,0.471,0.701}
\definecolor{ourbrown}{rgb}{0.772,0.432,0.102}
\definecolor{ourlightblue}{rgb}{0.364,0.619,0.782}
\definecolor{ourdarkgreen}{rgb}{0.572,0.586,0.0}
\definecolor{ourdarkred}{HTML}{94300f}         
\definecolor{ourlightred}{HTML}{f7c5b2}  

\newcommand{\rorl}{\textsc{Rorl}\xspace}
\newcommand{\iql}{\textsc{Iql}\xspace}

\newcommand{\dm}{\textsc{Dm}\xspace}
\newcommand{\meddcm}{\mbox{\textsc{Med}+\textsc{dcm}}\xspace}
\newcommand{\mednoise}{\mbox{\textsc{Med}+\textsc{Noise}}\xspace}
\newcommand{\hlagdcm}{\mbox{\textsc{H-Lag}+\textsc{dcm}}\xspace}
\newcommand{\hlag}{\mbox{\textsc{H-Lag}}\xspace}
\newcommand{\tdthreebc}{\textsc{Td}\textnormal{3}\textsc{Bc}\xspace}
\newcommand{\forl}{\textsc{Forl}\xspace}
\newcommand{\rorlmeanlag}{\mbox{\textsc{Rorl}+\textsc{Lag}-$\bar{s}$}\xspace}
\newcommand{\iqlmeanlag}{\mbox{\textsc{Iql}+\textsc{Lag}-$\bar{s}$}\xspace}
\newcommand{\tdthreebcmeanlag}{\mbox{\textsc{Td}\textnormal{3}\textsc{Bc}+\textsc{Lag}-$\bar{s}$}\xspace}
\newcommand{\dmbp}{\textsc{Dmbp}\xspace}
\newcommand{\difql}{\textsc{DQL}\xspace}

\newcommand{\zeroshotfm}{\textit{Zero-Shot FM}\xspace}

\newcommand{\maxlikelihood}{\textsc{max}\xspace}

\newcommand{\difqlmeanlag}{\mbox{\textsc{DQL}+\textsc{Lag}-$\bar{s}$}\xspace}
\newcommand{\difqlmedlag}{\mbox{\textsc{DQL}+\textsc{Lag}-$\tilde{s}$}\xspace}
\newcommand{\dmbplag}{\textsc{Dmbp}+\textsc{Lag}\xspace}
\newcommand{\candidselmethodname}{Dimension-wise Closest Match\xspace}
\newcommand{\candidsel}{DCM\xspace}
\newcommand{\fql}{\textsc{FQL}\xspace}
\newcommand{\fqlmeanlag}{\mbox{\textsc{FQL}+\textsc{Lag}-$\bar{s}$}\xspace}

\newcommand{\forldcm}{\textsc{Forl(DCM)}\xspace}
\newcommand{\scott}{\textsc{Forl(KDE)}\xspace}
\newcommand{\closemean}{\mbox{\textsc{DM-FS}-$\bar{s}$}\xspace}
\newcommand{\closemedian}{\mbox{\textsc{DM-FS}-$\tilde{s}$}\xspace}

\newcommand{\dmbmod}{\textsc{Dm-Mad}\xspace}
\newcommand{\dmbransac}{\textsc{Dm-Ransac}\xspace}
\newcommand{\dmbwelford}{\mbox{\textsc{Dm-Running}-$\mu$}\xspace}
\newcommand{\dmbwelfordreset}{\mbox{\textsc{Dm-Running}-$\mu$-p}\xspace}

\newcommand{\x}{\mathbf{x}}
\newcommand{\s}{\boldsymbol{s}}
\newcommand{\eps}{\boldsymbol{\epsilon}}
\newcommand{\antmaze}{\texttt{antmaze}\xspace}
\newcommand{\mazed}{\texttt{maze2d}\xspace}
\newcommand{\kitchc}{\texttt{kitchen-complete}\xspace}
\newcommand{\antu}{\texttt{antmaze-umaze-diverse}\xspace}
\newcommand{\antl}{\texttt{antmaze-large-diverse}\xspace}
\newcommand{\antm}{\texttt{antmaze-medium-diverse}\xspace}
\newcommand{\mazel}{\texttt{maze2d-large}\xspace}
\newcommand{\mazem}{\texttt{maze2d-medium}\xspace}
\newcommand{\aed}{\texttt{real-data-A}\xspace}
\newcommand{\electricity}{\texttt{real-data-B}\xspace}
\newcommand{\electricityhourly}{\texttt{real-data-C}\xspace}
\newcommand{\electricitynips}{\texttt{real-data-D}\xspace}
\newcommand{\exchangerate}{\texttt{real-data-E}\xspace}
\newcommand{\ogcube}{\texttt{cube-single-play}\xspace}
\newcommand{\ogantl}{\texttt{antmaze-large-navigate}\xspace}

%% file: tables/avgrewall.tex
\begin{wraptable}[45]{R}{0.5\linewidth}
\vspace{-2\baselineskip}
\vspace{-12pt}
\caption{\textbf{Normalized scores (mean ± std.) for \forl framework and the baselines.} Bold are the best values, and those not significantly different ($p > 0.05$, Welch's t-test).}
\label{tab:avgrewtable}
\begin{adjustbox}{max width=.5\textwidth}
\begin{tabular}{@{}l@{\hspace{-10ex}}rccc@{}}
\toprule
\textbf{\mazem}    & \multicolumn{1}{c}{\difql}& \difqlmeanlag   & \dmbplag        & \textcolor{ourdarkred}{\forl(ours)}        \\
\midrule
\aed               & \fmt{30.2}{6.5}         & \fmt{30.2}{8.6}               & \fmt{25.1}{9.8}           & \textbf{\fmt{63.3}{6.7}}    \\
\electricity       & \fmt{14.1}{12.1}        & \textbf{\fmt{53.4}{14.6}}     & \textbf{\fmt{41.2}{21.1}} & \textbf{\fmt{66.5}{18.2}}            \\
\electricityhourly & \fmt{-2.3}{3.3}         & \fmt{56.7}{18.5}              & \fmt{56.9}{18.4}          & \textbf{\fmt{86.3}{15.7}}   \\
\electricitynips   & \fmt{4.7}{5.0}          & \fmt{36.9}{16.3}              & \fmt{38.5}{14.2}          & \textbf{\fmt{103.4}{11.9}}  \\
\exchangerate      & \fmt{3.5}{8.8}          & \fmt{8.7}{6.0}                & \fmt{11.4}{2.8}           & \textbf{\fmt{51.2}{13.7}}   \\
\textbf{Average}   & \fmt{10.0}{}               & \fmt{37.2}{}                     & \fmt{34.6}{}                 & \textbf{\fmt{74.1}{}}                   \\ \midrule
\textbf{\mazel}   \\
\midrule
\aed               & \fmt{16.2}{5.5}         & \fmt{2.4}{1.1}                & \fmt{4.2}{5.8}            & \textbf{\fmt{42.9}{4.1}}    \\
\electricity       & \fmt{-0.5}{2.9}         & \fmt{5.5}{9.0}                & \fmt{15.0}{14.6}          & \textbf{\fmt{34.9}{9.2}}    \\
\electricityhourly & \fmt{0.9}{1.7}          & \fmt{16.6}{7.5}               & \fmt{26.8}{8.4}           & \textbf{\fmt{45.6}{4.1}}    \\
\electricitynips   & \fmt{3.0}{6.6}          & \fmt{8.6}{3.2}                & \fmt{13.4}{4.1}           & \textbf{\fmt{58.4}{6.5}}    \\
\exchangerate      & \fmt{-2.1}{0.4}         & \textbf{\fmt{2.6}{3.4}}       & \textbf{\fmt{0.9}{3.7}}   & \textbf{\fmt{12.0}{9.9}}    \\
\textbf{Average}   & \fmt{3.5}{}                & \fmt{7.1}{}                      & \fmt{12.1}{}                 & \textbf{\fmt{38.8}{}}          \\
\midrule
\textbf{\antu}     \\
\midrule
\aed               & \fmt{22.7}{3.0}         & \fmt{41.0}{5.2}               & \fmt{45.7}{4.8}           & \textbf{\fmt{65.3}{8.7}}             \\
\electricity       & \fmt{24.2}{3.5}         & \fmt{48.3}{7.0}               &  \textbf{\fmt{62.5}{13.2}}          & \textbf{\fmt{74.2}{10.8}}   \\
\electricityhourly & \fmt{21.7}{3.5}         & \fmt{50.4}{8.3}               & \fmt{60.4}{3.9}           & \textbf{\fmt{78.8}{8.5}}    \\
\electricitynips   & \fmt{5.8}{2.3}          & \fmt{26.7}{6.3}               & \fmt{29.2}{5.9}           & \textbf{\fmt{75.8}{8.0}}    \\
\exchangerate      & \fmt{6.0}{6.8}          & \fmt{58.0}{16.6}              & \fmt{59.3}{7.6}           & \textbf{\fmt{81.3}{6.9}}    \\
\textbf{Average}   & \fmt{16.1}{}               & \fmt{44.9}{}                     & \fmt{51.4}{}                 & \textbf{\fmt{75.1}{}}                   \\
\midrule
\textbf{\antm}     \\ 
\midrule
\aed               & \fmt{31.0}{6.5}         & \textbf{\fmt{40.0}{5.7}}    & \textbf{\fmt{39.7}{4.0}}    & \textbf{\fmt{44.0}{7.9}}             \\
\electricity       & \fmt{23.3}{4.8}         & \textbf{\fmt{48.3}{4.8}}      & \textbf{\fmt{43.3}{16.0}} & \textbf{\fmt{55.8}{7.0}}    \\
\electricityhourly & \fmt{10.0}{2.3}         & \textbf{\fmt{48.3}{3.4}}      & \textbf{\fmt{49.6}{3.7}}  & \textbf{\fmt{52.9}{9.5}}    \\
\electricitynips   & \fmt{11.7}{5.4}         & \fmt{46.7}{7.5}               & \fmt{41.7}{6.6}           & \textbf{\fmt{64.2}{8.6}}             \\
\exchangerate      & \textbf{\fmt{18.7}{4.5}} & \textbf{\fmt{27.3}{8.6}}      & \textbf{\fmt{26.0}{5.5}}  & \textbf{\fmt{26.7}{4.7}}             \\
\textbf{Average}   & \fmt{18.9}{}               & \fmt{42.1}{}                     & \fmt{40.1}{}                 & \textbf{\fmt{48.7}{}}                   \\
\midrule
\textbf{\antl}     \\ 
\midrule
\aed               & \fmt{11.0}{1.9}         & \fmt{11.3}{4.9}               & \fmt{9.0}{4.5}            & \textbf{\fmt{34.3}{5.7}}    \\
\electricity       & \fmt{5.8}{4.8}          & \fmt{9.2}{4.6}                & \fmt{8.3}{2.9}            & \textbf{\fmt{46.7}{11.9}}   \\
\electricityhourly & \fmt{5.4}{2.4}          & \fmt{22.1}{5.6}               & \fmt{17.9}{3.8}           & \textbf{\fmt{33.8}{6.8}}    \\
\electricitynips   & \fmt{2.5}{2.3}          & \fmt{14.2}{3.7}               & \fmt{14.2}{6.3}           & \textbf{\fmt{46.7}{12.6}}   \\
\exchangerate      & \textbf{\fmt{5.3}{3.8}} &  \textbf{\fmt{3.3}{2.4}}      &  \textbf{\fmt{3.3}{0.0}}   & \textbf{\fmt{11.3}{7.3}}    \\
\textbf{Average}   & \fmt{6.0}{}                & \fmt{12.0}{}                     & \fmt{10.5}{}                 & \textbf{\fmt{34.6}{}}          \\
\midrule
\textbf{\kitchc}     \\ 
\midrule
\aed    & \textbf{\fmt{16.6}{1.4}} & \fmt{7.2}{1.9} & \fmt{8.7}{1.3} & \textbf{\fmt{12.0}{3.9}} \\
\electricity & \fmt{12.9}{4.1} & \textbf{\fmt{32.7}{6.5}} & \fmt{20.0}{3.1} & \textbf{\fmt{33.1}{5.6}} \\
\electricityhourly & \fmt{13.4}{1.7} & \textbf{\fmt{23.9}{6.6}} & \textbf{\fmt{20.5}{3.3}} & \textbf{\fmt{23.9}{6.0}} \\
\electricitynips  & \fmt{7.5}{2.5} & \textbf{\fmt{24.0}{9.2}} & \textbf{\fmt{28.1}{8.1}} & \textbf{\fmt{27.1}{10.1}} \\
\exchangerate     & \textbf{\fmt{18.5}{6.0}} & \fmt{2.8}{2.1} & \fmt{6.2}{1.7} & \fmt{10.3}{3.0} \\
\textbf{Average} & \fmt{13.8}{}  & \fmt{18.1}{}  & \fmt{16.7}{} & \textbf{\fmt{21.3}{}}\\
\end{tabular}
\end{adjustbox}
\begin{adjustbox}{max width=.5\textwidth}
\begin{tabular}{@{}l@{\hspace{-1ex}}rccc@{}}
\toprule
\textbf{\ogcube}    & \multicolumn{1}{c}{\fql}& \fqlmeanlag   & \textcolor{ourdarkred}{\forl-F (ours)} \\    
\midrule
\aed & \fmt{0.0}{0.0} & \fmt{0.0}{0.0} & \textbf{\fmt{23.7}{3.6}} \\
\electricity & \fmt{0.0}{0.0} & \fmt{15.0}{7.0} & \textbf{\fmt{60.0}{7.0}} \\
\electricityhourly & \fmt{0.4}{0.9} & \fmt{10.0}{1.7} & \textbf{\fmt{42.1}{5.6}} \\
\electricitynips & \fmt{0.0}{0.0} & \fmt{0.8}{1.9} & \textbf{\fmt{70.0}{13.0}} \\
\exchangerate & \fmt{0.0}{0.0} & \fmt{0.0}{0.0} & \textbf{\fmt{32.7}{9.5}} \\
\textbf{Average} & \fmt{0.1}{} & \fmt{5.2}{} & \textbf{\fmt{45.7}{}} \\
\midrule
\textbf{\ogantl}    & \multicolumn{1}{c}{}& & \\
\midrule
\aed & \textbf{\fmt{22.7}{2.2}} & \fmt{1.3}{0.7} & \textbf{\fmt{24.3}{4.3}} \\
\electricity & \fmt{21.7}{5.4} & \textbf{\fmt{29.2}{8.8}} & \textbf{\fmt{40.0}{7.6}} \\
\electricityhourly & \fmt{5.0}{1.1} & \fmt{34.6}{6.7} & \textbf{\fmt{55.8}{3.7}} \\
\electricitynips & \fmt{0.8}{1.9} & \fmt{37.5}{5.1} & \textbf{\fmt{75.8}{5.4}} \\
\exchangerate & \textbf{\fmt{10.0}{4.1}} & \fmt{3.3}{0.0} & \textbf{\fmt{15.3}{8.7}} \\
\textbf{Average} & \fmt{12.0}{} & \fmt{21.2}{} & \textbf{\fmt{42.2}{}} \\
\bottomrule
\end{tabular}
\end{adjustbox}
\end{wraptable}

%% file: tables/mazelargerewtable.tex
\begin{table*}[t!]
\caption{\textbf{Normalized scores (mean ± std.) for FORL and baselines on \mazel.} Bolds denote the best scores and those not significantly different (Welch’s t-test, p > 0.05). Suffixes -T and -R denote the use of TD3+BC \cite{fujimoto2021a} and RORL \cite{yang2022rorl}, respectively.
}
\label{tab:offline}
\centering
\adjustbox{max width=\linewidth}{
\begin{tabular}{lcccccccccc}
\toprule
\multicolumn{1}{c}{} & \multicolumn{4}{c}{\texttt{\tdthreebc Policy}}& \multicolumn{4}{c}{\texttt{\rorl Policy}} \\
\cmidrule(lr){2-5} \cmidrule(lr){6-9}
\textbf{\mazel} & \tdthreebc             & \tdthreebcmeanlag   & \dmbplag-T        & \textcolor{ourdarkred}{\forl(ours)-T}  &   \rorl             & \rorlmeanlag   & \dmbplag-R        & \textcolor{ourdarkred}{\forl(ours)-R} \\
\midrule
\aed &  \textbf{\fmt{14.7}{5.7}} & \fmt{2.5}{2.7} & \fmt{4.8}{3.9} & \textbf{\fmt{20.7}{3.5}} & \fmt{12.2}{2.3} & \fmt{13.0}{2.0} & \fmt{4.3}{5.0} &  \textbf{\fmt{56.9}{3.0}}  \\
\electricity  & \fmt{-0.9}{2.0} & \fmt{4.6}{8.9} & \fmt{11.7}{12.6} & \textbf{\fmt{56.8}{14.4}}  & \fmt{1.2}{5.5} & \fmt{13.1}{14.7} & \fmt{28.5}{11.7} &  \textbf{\fmt{98.5}{19.0}}  \\
\electricityhourly  & \fmt{0.8}{1.9} & \fmt{21.6}{8.4} & \fmt{29.5}{13.7} & \textbf{\fmt{56.9}{14.6}} & \fmt{3.1}{0.9} & \fmt{60.6}{8.5} & \fmt{39.4}{6.1} &  \textbf{\fmt{139.0}{15.1}}  \\
\electricitynips & \fmt{2.5}{4.4} &  \fmt{14.9}{4.3} &  \fmt{14.4}{6.8} & \textbf{\fmt{29.5}{10.3}}  & \fmt{-1.6}{0.7} & \fmt{17.9}{6.8} & \fmt{17.5}{6.0} & \textbf{\fmt{33.1}{2.3}} \\
\exchangerate & \fmt{-2.3}{0.2} & \fmt{1.0}{4.2} & \fmt{2.0}{3.9}  & \textbf{\fmt{8.0}{4.2}} & \fmt{-0.9}{2.0} & \fmt{3.3}{4.4} & \fmt{2.2}{4.5} & \textbf{\fmt{32.2}{15.3}} \\
\textbf{Average} &  \fmt{3.0}{}   &  \fmt{8.9}{}   &  \fmt{12.5}{}   & \textbf{\fmt{34.4}{} } & \fmt{2.8}{}  &  \fmt{21.6}{}   &  \fmt{18.4}{}   &  \textbf{\fmt{71.9}{}}  \\
\bottomrule
\end{tabular}
}
\vskip-1.6em
\end{table*}

%% file: tables/otherpoliciestable.tex
\begin{table*}[t]
\vspace{-10pt}
\caption{
\textbf{Normalized scores (mean ± std.) for FORL and baselines with \tdthreebc and \rorl on \mazem.} Algorithms are grouped by their underlying policies—TD3 with Behavior Cloning (TD3+BC) \cite{fujimoto2021a} and Robust Offline Reinforcement Learning (RORL) \cite{yang2022rorl} to highlight that performance variations stem from the algorithms themselves rather than the policies employed. Suffixes -T and -R denote the use of TD3+BC and RORL policies, respectively.
}
\label{table:rorltd3bc_all}
\centering
\vspace{5pt}
\adjustbox{max width=\linewidth}
{
\begin{tabular}{lcccccccccc}
\toprule
\multicolumn{1}{c}{} & \multicolumn{4}{c}{\texttt{\tdthreebc Policy}}& \multicolumn{4}{c}{\texttt{\rorl Policy}} \\
\cmidrule(lr){2-5} \cmidrule(lr){6-9}
\textbf{\mazem} & \tdthreebc             & \tdthreebcmeanlag   & \dmbplag-T        & \textcolor{ourdarkred}{\forl(ours)-T}  &   \rorl             & \rorlmeanlag   & \dmbplag-R        & \textcolor{ourdarkred}{\forl(ours)-R} \\
\midrule
\aed & \textbf{\fmt{37.4}{9.5}} & \fmt{16.2}{5.1} & \fmt{16.2}{7.3} &  \fmt{22.1}{6.6}  & \textbf{\fmt{80.7}{14.2}} & \fmt{57.9}{8.6} & \fmt{47.8}{10.2} & \fmt{52.7}{5.0} \\
\electricity  &  \fmt{3.6}{2.3}  &  \textbf{\fmt{6.1}{4.4}} &  \textbf{\fmt{14.4}{8.1}} & \textbf{\fmt{28.6}{19.0}} & \fmt{37.8}{7.5} & \textbf{\fmt{85.9}{19.8}} & \textbf{\fmt{91.0}{21.6}} & \textbf{\fmt{109.6}{19.5}} \\
\electricityhourly & \fmt{-2.3}{1.3} & \textbf{\fmt{30.0}{10.0}} &  \textbf{\fmt{19.3}{6.2}} &  \textbf{\fmt{24.5}{10.2}} & \fmt{33.7}{4.6} & \fmt{89.2}{15.8} & \fmt{93.4}{16.3} & \textbf{\fmt{125.4}{14.5}} \\
\electricitynips & \fmt{6.3}{3.4} & \fmt{15.5}{3.8} & \fmt{12.2}{2.6} &   \textbf{\fmt{38.7}{13.4}} & \fmt{37.0}{17.0} & \fmt{71.2}{27.2} & \fmt{77.7}{26.2} & \textbf{\fmt{136.1}{11.9}} \\
\exchangerate & \fmt{-3.7}{1.5} &  \textbf{\fmt{9.7}{5.2}} &  \textbf{\fmt{11.5}{6.6}}  & \textbf{\fmt{15.1}{8.8}} & \textbf{\fmt{60.9}{13.5}} & \fmt{10.0}{7.3} & \fmt{14.2}{8.6} & \textbf{\fmt{61.2}{14.6}} \\
\textbf{Average} &  \fmt{8.3}{}  &  \fmt{15.5}{}  & \fmt{14.7}{}  & \textbf{\fmt{25.8}{} } & \fmt{50.0}{}   & \fmt{62.8}{}  & \fmt{64.8}{} & \textbf{\fmt{97.0}{} } \\
\bottomrule
\end{tabular}
}
\end{table*}

%% file: tables/iql_table.tex
\begin{table*}[t!]
\caption{
\textbf{Normalized scores (mean ± std.) for FORL and baselines with \iql.} Suffix -I denote the use of \iql algorithm \cite{iql_Kostrikov_2021}.}
\label{tab:iqlperformance}
\centering
\begin{adjustbox}{max width=\textwidth}
\begin{tabular}{@{}l@{\hspace{-2ex}}lcccc@{}}
\toprule
\textbf{\mazem}    & \multicolumn{1}{c}{\iql}& \iqlmeanlag   & \dmbplag-I        & \textcolor{ourdarkred}{\forl(ours)-I}        \\
\midrule
\aed & \textbf{\fmt{39.5}{7.8}} & \fmt{16.0}{4.9} & \fmt{12.2}{7.7} & \fmt{19.5}{2.9} \\
\electricity & \fmt{3.7}{7.3} & \fmt{13.1}{8.7} & \fmt{14.0}{9.5} & \textbf{\fmt{32.3}{13.1}} \\
\electricityhourly & \fmt{-1.1}{2.5} & \textbf{\fmt{33.1}{11.9}} & \textbf{\fmt{28.0}{10.9}} & \textbf{\fmt{31.8}{9.9}} \\
\electricitynips & \fmt{10.1}{2.6} & \fmt{12.4}{5.6} & \fmt{12.7}{9.0} & \textbf{\fmt{40.0}{10.1}} \\
\exchangerate & \fmt{-4.5}{0.2} & \fmt{12.4}{5.2} & \fmt{9.9}{2.9} & \textbf{\fmt{24.0}{8.4}} \\
\textbf{Average} & \fmt{9.6}{} & \fmt{17.4}{} & \fmt{15.4}{} & \textbf{\fmt{29.5}{}} \\\midrule
\textbf{\mazel}   \\
\midrule
\aed & \textbf{\fmt{16.2}{6.2}} & \fmt{12.5}{5.5} & \fmt{6.0}{4.7} & \textbf{\fmt{24.3}{9.4}} \\
\electricity & \fmt{-0.6}{2.6} & \fmt{3.3}{7.7} & \fmt{13.5}{10.5} & \textbf{\fmt{42.8}{9.8}} \\
\electricityhourly & \fmt{0.1}{1.5} & \fmt{27.8}{13.1} & \fmt{27.9}{5.5} & \textbf{\fmt{46.7}{9.2}} \\
\electricitynips & \fmt{1.2}{3.7} & \fmt{11.2}{7.4} & \fmt{8.0}{7.5} & \textbf{\fmt{23.9}{8.4}} \\
\exchangerate & \fmt{-2.3}{0.2} & \fmt{-1.5}{1.1} & \textbf{\fmt{1.5}{4.1}} & \textbf{\fmt{7.9}{7.2}} \\
\textbf{Average} & \fmt{2.9}{} & \fmt{10.7}{} & \fmt{11.4}{} & \textbf{\fmt{29.1}{}} \\
\midrule
\textbf{\antu}     \\
\midrule
\aed & \fmt{23.0}{1.4} & \textbf{\fmt{43.7}{6.1}} & \textbf{\fmt{44.0}{9.6}} & \textbf{\fmt{50.3}{11.0}} \\
\electricity & \fmt{21.7}{5.4} & \textbf{\fmt{55.0}{6.2}} & \textbf{\fmt{61.7}{9.5}} & \textbf{\fmt{70.8}{13.2}} \\
\electricityhourly & \fmt{20.0}{3.2} & \fmt{45.4}{4.0} & \fmt{58.3}{5.1} & \textbf{\fmt{73.8}{5.4}} \\
\electricitynips & \fmt{6.7}{5.6} & \fmt{26.7}{8.6} & \fmt{29.2}{6.6} & \textbf{\fmt{77.5}{4.8}} \\
\exchangerate & \fmt{8.0}{8.7} & \textbf{\fmt{60.0}{11.8}} & \textbf{\fmt{56.0}{9.8}} & \textbf{\fmt{68.0}{13.0}} \\
\textbf{Average} & \fmt{15.9}{} & \fmt{46.1}{} & \fmt{49.8}{} & \textbf{\fmt{68.1}{}} \\
\midrule
\textbf{\antm}     \\ 
\midrule
\aed & \textbf{\fmt{18.3}{5.5}} & \textbf{\fmt{21.0}{5.8}} & \textbf{\fmt{24.7}{4.0}} & \fmt{17.7}{3.5} \\
\electricity & \textbf{\fmt{7.5}{3.5}} & \textbf{\fmt{7.5}{5.4}} & \textbf{\fmt{8.3}{7.8}} & \textbf{\fmt{11.7}{7.5}} \\
\electricityhourly & \fmt{2.5}{3.7} & \textbf{\fmt{18.8}{6.1}} & \textbf{\fmt{15.8}{4.8}} & \textbf{\fmt{16.2}{7.4}} \\
\electricitynips & \fmt{8.3}{4.2} & \textbf{\fmt{12.5}{5.9}} & \textbf{\fmt{12.5}{7.8}} & \textbf{\fmt{16.7}{2.9}} \\
\exchangerate & \fmt{3.3}{4.1} & \textbf{\fmt{22.7}{8.0}} & \textbf{\fmt{22.7}{12.3}} & \textbf{\fmt{22.7}{6.4}} \\
\textbf{Average} & \fmt{8.0}{} & \fmt{16.5}{} & \fmt{16.8}{} & \textbf{\fmt{17.0}{}} \\
\bottomrule
\end{tabular}
\end{adjustbox}
\end{table*}

%% file: tables/avgrew_dm_ablation.tex
\begin{table*}[t!]
\caption{\textbf{Normalized scores (mean ± std.) for no-access to past offsets setting.} This table shows the performance comparison of other heuristics variants using \forl-\dm or leveraging a \zeroshotfm in combination with \forl-\dm when we do not have access to past offsets.}
\label{tab:forldmablation}
\centering
\begin{adjustbox}{max width=\textwidth}
\begin{tabular}{lcccccc}
\toprule
\textbf{\mazem} & \difql & \mednoise & \meddcm & \hlag & \hlagdcm & \forl-\dm \\
\midrule
\aed & \fmt{30.2}{6.5} & \fmt{27.4}{14.5} & \fmt{27.4}{12.2} & \fmt{29.7}{11.3} & \textbf{\fmt{48.8}{10.0}} & \textbf{\fmt{55.2}{10.6}} \\
\electricity & \fmt{14.1}{12.1} & \fmt{23.9}{14.3} & \textbf{\fmt{33.1}{19.5}} & \fmt{4.5}{12.3} & \textbf{\fmt{50.3}{12.9}} & \textbf{\fmt{56.8}{24.7}} \\
\electricityhourly & \fmt{-2.3}{3.3} & \fmt{17.6}{8.1} & \fmt{-2.6}{2.5} & \fmt{26.5}{5.4} & \fmt{30.3}{5.6} & \textbf{\fmt{52.8}{10.3}} \\
\electricitynips & \fmt{4.7}{5.0} & \fmt{18.9}{9.1} & \textbf{\fmt{64.1}{12.7}} & \fmt{18.2}{13.7} & \fmt{1.8}{3.8} & \textbf{\fmt{60.1}{20.2}} \\
\exchangerate & \fmt{3.5}{8.8} & \fmt{21.8}{15.6} & \fmt{-2.1}{1.6} & \textbf{\fmt{56.7}{11.6}} & \textbf{\fmt{45.7}{12.4}} & \textbf{\fmt{60.5}{18.2}} \\
\textbf{Average} & \fmt{10.0}{} & \fmt{21.9}{} & \fmt{24.0}{} & \fmt{27.1}{} & \fmt{35.4}{} & \textbf{\fmt{57.1}{}} \\
\midrule
\multicolumn{7}{l}{\textbf{\mazel}} \\
\midrule
\aed & \textbf{\fmt{16.2}{5.5}} & \fmt{6.4}{2.9} & \fmt{-1.8}{0.5} & \fmt{7.1}{3.9} & \fmt{7.3}{2.0} & \textbf{\fmt{11.1}{4.3}} \\
\electricity & \fmt{-0.5}{2.9} & \fmt{-0.1}{1.7} & \fmt{-1.4}{1.7} & \fmt{-1.4}{2.3} & \textbf{\fmt{0.9}{4.5}} & \textbf{\fmt{13.4}{10.3}} \\
\electricityhourly & \fmt{0.9}{1.7} & \fmt{3.2}{4.7} & \fmt{-2.0}{0.6} & \fmt{3.4}{2.0} & \fmt{0.5}{1.7} & \textbf{\fmt{9.4}{2.5}} \\
\electricitynips & \fmt{3.0}{6.6} & \fmt{3.7}{6.8} & \textbf{\fmt{47.6}{16.4}} & \fmt{1.0}{4.5} & \fmt{2.9}{4.0} & \fmt{7.4}{7.2} \\
\exchangerate & \fmt{-2.1}{0.4} & \textbf{\fmt{3.8}{4.7}} & \textbf{\fmt{-0.5}{3.7}} & \textbf{\fmt{1.8}{2.7}} & \textbf{\fmt{7.2}{2.3}} & \textbf{\fmt{7.7}{7.6}} \\
\textbf{Average} & \fmt{3.5}{} & \fmt{3.4}{} & \fmt{8.4}{} & \fmt{2.4}{} & \fmt{3.8}{} & \textbf{\fmt{9.8}{}} \\
\midrule
\multicolumn{7}{l}{\textbf{\antu}} \\
\midrule
\aed & \fmt{22.7}{3.0} & \fmt{8.3}{2.4} & \fmt{8.3}{4.7} & \fmt{32.3}{6.3} & \textbf{\fmt{62.7}{11.2}} & \textbf{\fmt{48.7}{9.1}} \\
\electricity & \fmt{24.2}{3.5} & \fmt{9.2}{5.4} & \fmt{25.0}{13.2} & \fmt{41.7}{5.1} & \fmt{33.3}{2.9} & \textbf{\fmt{56.7}{8.1}} \\
\electricityhourly & \fmt{21.7}{3.5} & \fmt{10.8}{7.7} & \textbf{\fmt{75.8}{6.4}} & \fmt{36.7}{3.2} & \fmt{59.2}{11.5} & \fmt{55.0}{8.7} \\
\electricitynips & \fmt{5.8}{2.3} & \fmt{17.5}{11.6} & \fmt{6.7}{6.3} & \fmt{10.8}{2.3} & \fmt{42.5}{7.5} & \textbf{\fmt{54.2}{7.2}} \\
\exchangerate & \fmt{6.0}{6.8} & \textbf{\fmt{50.0}{13.5}} & \fmt{2.0}{3.0} & \fmt{14.0}{8.3} & \textbf{\fmt{42.7}{11.9}} & \textbf{\fmt{53.3}{7.8}} \\
\textbf{Average} & \fmt{16.1}{} & \fmt{19.2}{} & \fmt{23.6}{} & \fmt{27.1}{} & \fmt{48.1}{} & \textbf{\fmt{53.6}{}} \\
\midrule
\multicolumn{7}{l}{\textbf{\antm}} \\
\midrule
\aed & \fmt{31.0}{6.5} & \fmt{34.3}{10.4} & \textbf{\fmt{55.0}{6.8}} & \fmt{15.0}{4.9} & \fmt{17.3}{6.3} & \fmt{4.3}{0.9} \\
\electricity & \fmt{23.3}{4.8} & \fmt{15.8}{9.9} & \textbf{\fmt{42.5}{4.6}} & \fmt{33.3}{4.2} & \fmt{28.3}{3.5} & \fmt{6.7}{4.8} \\
\electricityhourly & \fmt{10.0}{2.3} & \fmt{23.3}{5.4} & \fmt{10.0}{4.3} & \fmt{31.2}{4.9} & \textbf{\fmt{40.4}{6.7}} & \fmt{4.6}{1.7} \\
\electricitynips & \fmt{11.7}{5.4} & \fmt{26.7}{6.3} & \textbf{\fmt{33.3}{15.9}} & \fmt{9.2}{3.5} & \textbf{\fmt{36.7}{6.8}} & \fmt{3.3}{3.5} \\
\exchangerate & \fmt{18.7}{4.5} & \textbf{\fmt{26.7}{18.1}} & \fmt{0.0}{0.0} & \fmt{14.7}{5.6} & \textbf{\fmt{46.0}{16.2}} & \fmt{6.0}{4.3} \\
\textbf{Average} & \fmt{18.9}{} & \fmt{25.4}{} & \fmt{28.2}{} & \fmt{20.7}{} & \textbf{\fmt{33.7}{}} & \fmt{5.0}{} \\
\midrule
\multicolumn{7}{l}{\textbf{\antl}} \\
\midrule
\aed & \textbf{\fmt{11.0}{1.9}} & \fmt{5.0}{3.9} & \fmt{1.3}{1.4} & \textbf{\fmt{9.0}{1.9}} & \textbf{\fmt{11.3}{4.1}} & \textbf{\fmt{11.0}{3.0}} \\
\electricity & \textbf{\fmt{5.8}{4.8}} & \textbf{\fmt{7.5}{3.5}} & \textbf{\fmt{5.0}{5.4}} & \textbf{\fmt{9.2}{1.9}} & \textbf{\fmt{7.5}{1.9}} & \textbf{\fmt{11.7}{4.6}} \\
\electricityhourly & \fmt{5.4}{2.4} & \fmt{5.4}{2.4} & \fmt{1.7}{0.9} & \textbf{\fmt{10.8}{2.7}} & \textbf{\fmt{12.9}{5.8}} & \textbf{\fmt{12.1}{2.7}} \\
\electricitynips & \fmt{2.5}{2.3} & \textbf{\fmt{15.8}{6.2}} & \textbf{\fmt{11.7}{4.6}} & \textbf{\fmt{8.3}{6.6}} & \textbf{\fmt{14.2}{6.3}} & \textbf{\fmt{15.0}{9.6}} \\
\exchangerate & \textbf{\fmt{5.3}{3.8}} & \textbf{\fmt{5.3}{3.0}} & \fmt{1.3}{1.8} & \textbf{\fmt{8.7}{3.0}} & \textbf{\fmt{6.0}{2.8}} & \textbf{\fmt{8.7}{5.1}} \\
\textbf{Average} & \fmt{6.0}{} & \fmt{7.8}{} & \fmt{4.2}{} & \fmt{9.2}{} & \fmt{10.4}{} & \textbf{\fmt{11.7}{}} \\
\bottomrule
\end{tabular}
\end{adjustbox}
\end{table*}

\begin{table*}[t!]
\footnotesize
\caption{\textbf{Additional results for no-access to past offsets setting.} We present alternative ways of using \forl's diffusion model (\dm) component when we do not have access to past offsets.}
\label{tab:forldmablation_maze2dmed}
\centering
\begin{adjustbox}{max width=\textwidth}
\begin{tabular}{lccccc}
\toprule
\textbf{\mazem} & \dmbmod & \dmbransac & \dmbwelford & \dmbwelfordreset & \forl-\dm \\
\midrule
\aed & \textbf{\fmt{44.3}{10.6}} & \textbf{\fmt{46.8}{12.4}} & \textbf{\fmt{37.8}{13.6}} & \fmt{37.4}{8.7} & \textbf{\fmt{55.2}{10.6}} \\
\electricity & \textbf{\fmt{46.8}{19.0}} & \textbf{\fmt{44.6}{17.1}} & \textbf{\fmt{42.4}{24.5}} & \textbf{\fmt{51.6}{11.7}} & \textbf{\fmt{56.8}{24.7}} \\
\electricityhourly & \textbf{\fmt{46.4}{10.9}} & \textbf{\fmt{44.5}{10.1}} & \textbf{\fmt{41.8}{16.0}} & \textbf{\fmt{61.7}{14.1}} & \textbf{\fmt{52.8}{10.3}} \\
\electricitynips & \textbf{\fmt{47.7}{22.9}} & \textbf{\fmt{49.9}{23.2}} & \textbf{\fmt{44.4}{24.1}} & \textbf{\fmt{52.5}{13.3}} & \textbf{\fmt{60.1}{20.2}} \\
\exchangerate & \textbf{\fmt{42.9}{10.6}} & \textbf{\fmt{52.4}{15.7}} & \textbf{\fmt{48.3}{18.8}} & \fmt{24.2}{12.2} & \textbf{\fmt{60.5}{18.2}} \\
\textbf{Average} & \fmt{45.6}{} & \fmt{47.6}{} & \fmt{42.9}{} & \fmt{45.5}{} & \textbf{\fmt{57.1}{}} \\
\bottomrule
\end{tabular}
\end{adjustbox}
\end{table*}

%% file: tables/avgrewtable_og_significant.tex
\begin{table*}[t!]
\caption{\textbf{Normalized scores (mean ± std.) for \forl framework and the baselines.}}
\label{tab:avgrew_og_dcmablation}
\centering
\begin{adjustbox}{max width=\textwidth}
\begin{tabular}{lcccccc}
\toprule
\textbf{\ogcube}    & \multicolumn{1}{c}{\fql}& \forl-\dm-F & \fqlmeanlag & \forl(\maxlikelihood)-F & \scott-F & \textcolor{ourdarkred}{\forl-F (ours)}   \\
\midrule
\aed & \fmt{0.0}{0.0} & \fmt{3.0}{1.4} & \fmt{0.0}{0.0} & \fmt{0.0}{0.0} & \fmt{0.0}{0.0} & \textbf{\fmt{23.7}{3.6}} \\
\electricity & \fmt{0.0}{0.0} & \fmt{6.7}{5.6} & \fmt{15.0}{7.0} & \fmt{43.3}{4.8} & \fmt{2.5}{2.3} & \textbf{\fmt{60.0}{7.0}} \\
\electricityhourly & \fmt{0.4}{0.9} & \fmt{4.6}{1.7} & \fmt{10.0}{1.7} & \textbf{\fmt{39.6}{4.4}} & \textbf{\fmt{38.3}{3.8}} & \textbf{\fmt{42.1}{5.6}} \\
\electricitynips & \fmt{0.0}{0.0} & \fmt{2.5}{2.3} & \fmt{0.8}{1.9} & \fmt{7.5}{1.9} & \fmt{0.0}{0.0} & \textbf{\fmt{70.0}{13.0}} \\
\exchangerate & \fmt{0.0}{0.0} & \fmt{8.0}{3.0} & \fmt{0.0}{0.0} & \textbf{\fmt{21.3}{8.0}} & \fmt{0.0}{0.0} & \textbf{\fmt{32.7}{9.5}} \\
\textbf{Average} & \fmt{0.1}{} & \fmt{5.0}{} & \fmt{5.2}{} & \fmt{22.3}{} & \fmt{8.2}{} & \textbf{\fmt{45.7}{}} \\
\bottomrule
\end{tabular}
\end{adjustbox}
\end{table*}

%% file: tables/dcmablation_maxminerrorfig6.tex
\begin{table*}[t!]
\caption{\textbf{Comparison of algorithm performance on error metrics.}}
\label{tab:dcm_max_minerrorcomparisonfig6}
\centering
\begin{tabular}{@{}lrrr@{}}
\toprule
\textbf{Algorithm} & \textbf{Minimum Error $\downarrow$} & \textbf{Maximum Error $\downarrow$} & \textbf{Mean Error $\downarrow$} \\
\midrule
\forl-\candidsel & 0.02 & \textbf{2.40} & \textbf{0.87 $\pm$ 0.60} \\
\forl-\maxlikelihood & \textbf{0.01} & 9.33 & 2.05 $\pm$ 1.74 \\
\difql & 2.26 & 11.30 & 5.51 $\pm$ 2.13 \\
\forl-\dm \textit{(no past offsets)} & \textbf{0.01} & 9.94 & 3.68 $\pm$ 2.25 \\
\difqlmeanlag & 0.04 & 6.28 & 1.76 $\pm$ 1.13 \\
\difqlmedlag & 0.05 & 6.56 & 1.87 $\pm$ 1.19 \\
\dmbplag & 0.03 & 6.28 & 1.69 $\pm$ 1.11 \\
\bottomrule
\end{tabular}
\end{table*}

%% file: tables/avgrew_dcm_ablation_table.tex
\begin{table*}[t!]
\footnotesize
\caption{\textbf{Normalized scores (mean ± std.) for \candidsel candidate selection and the baselines.} Bold are the best values, and those not significantly different ($p > 0.05$, Welch's t-test).}
\label{tab:dcm_ablation_table}
\centering
\begin{adjustbox}{max width=\textwidth}
\begin{tabular}{@{}l@{\hspace{-2ex}}rcccc@{}}
\toprule
\textbf{\mazem} & \closemean & \closemedian & \forl(\maxlikelihood) & \scott & \textcolor{ourdarkred}{\forldcm} \\
\midrule
\aed & \fmt{40.6}{9.6} & \textbf{\fmt{73.7}{8.4}} & \fmt{41.2}{8.2} & \fmt{49.7}{5.8} & \textbf{\fmt{63.3}{6.7}} \\
\electricity & \fmt{59.8}{16.9} & \fmt{59.3}{20.0} & \fmt{58.9}{14.1} & \textbf{\fmt{96.2}{14.0}} & \fmt{66.5}{18.2} \\
\electricityhourly & \fmt{64.9}{17.0} & \fmt{67.8}{17.7} & \fmt{66.1}{16.4} & \textbf{\fmt{94.9}{13.8}} & \textbf{\fmt{86.3}{15.7}} \\
\electricitynips & \fmt{45.1}{21.6} & \fmt{45.0}{19.5} & \fmt{44.4}{21.6} & \textbf{\fmt{86.9}{14.5}} & \textbf{\fmt{103.4}{11.9}} \\
\exchangerate & \fmt{12.5}{6.3} & \fmt{20.8}{7.2} & \fmt{11.8}{5.5} & \textbf{\fmt{49.3}{16.5}} & \textbf{\fmt{51.2}{13.7}} \\
\textbf{Average} & \fmt{44.6}{} & \fmt{53.3}{} & \fmt{44.5}{} & \textbf{\fmt{75.4}{}} & \fmt{74.1}{} \\
\midrule
\textbf{\mazel} & &  & & & \\
\midrule
\aed & \fmt{11.9}{5.5} & \fmt{20.8}{6.2} & \fmt{11.1}{2.3} & \fmt{6.6}{2.2} & \textbf{\fmt{42.9}{4.1}} \\
\electricity & \textbf{\fmt{27.9}{14.7}} & \textbf{\fmt{25.1}{11.7}} & \textbf{\fmt{28.3}{7.1}} & \fmt{20.9}{7.2} & \textbf{\fmt{34.9}{9.2}} \\
\electricityhourly & \fmt{34.6}{6.8} & \fmt{32.4}{5.8} & \textbf{\fmt{34.6}{13.6}} & \fmt{36.0}{7.3} & \textbf{\fmt{45.6}{4.1}} \\
\electricitynips & \fmt{16.4}{12.0} & \fmt{15.5}{9.0} & \fmt{18.4}{9.9} & \fmt{11.8}{3.2} & \textbf{\fmt{58.4}{6.5}} \\
\exchangerate & \textbf{\fmt{8.4}{5.0}} & \textbf{\fmt{7.9}{3.9}} & \textbf{\fmt{9.2}{6.1}} & \textbf{\fmt{5.7}{5.1}} & \textbf{\fmt{12.0}{9.9}} \\
\textbf{Average} & \fmt{19.8}{} & \fmt{20.3}{} & \fmt{20.3}{} & \fmt{16.2}{} & \textbf{\fmt{38.8}{}} \\
\midrule
\textbf{\antu}  & &  & & & \\
\midrule
\aed & \fmt{56.0}{13.3} & \fmt{58.7}{7.4} & \fmt{59.7}{9.7} & \textbf{\fmt{80.3}{4.1}} & \fmt{65.3}{8.7} \\
\electricity & \textbf{\fmt{69.2}{11.3}} & \textbf{\fmt{76.7}{11.3}} & \fmt{65.0}{10.5} & \textbf{\fmt{82.5}{8.0}} & \textbf{\fmt{74.2}{10.8}} \\
\electricityhourly & \textbf{\fmt{72.1}{5.4}} & \textbf{\fmt{75.8}{4.8}} & \textbf{\fmt{76.2}{5.4}} & \textbf{\fmt{67.9}{7.9}} & \textbf{\fmt{78.8}{8.5}} \\
\electricitynips & \fmt{65.0}{11.3} & \fmt{61.7}{15.1} & \fmt{63.3}{6.8} & \textbf{\fmt{85.0}{4.8}} & \textbf{\fmt{75.8}{8.0}} \\
\exchangerate & \textbf{\fmt{76.7}{16.2}} & \textbf{\fmt{74.7}{18.3}} & \textbf{\fmt{72.0}{15.0}} & \textbf{\fmt{75.3}{8.0}} & \textbf{\fmt{81.3}{6.9}} \\
\textbf{Average} & \fmt{67.8}{} & \fmt{69.5}{} & \fmt{67.2}{} & \textbf{\fmt{78.2}{}} & \fmt{75.1}{} \\
\midrule
\textbf{\antm}   & &  & & & \\
\midrule
\aed & \fmt{39.0}{9.5} & \fmt{27.3}{7.5} & \fmt{36.0}{4.8} & \textbf{\fmt{62.7}{6.3}} & \fmt{44.0}{7.9} \\
\electricity & \fmt{34.2}{9.0} & \fmt{35.8}{8.1} & \fmt{36.7}{7.5} & \textbf{\fmt{59.2}{8.0}} & \textbf{\fmt{55.8}{7.0}} \\
\electricityhourly & \fmt{36.2}{2.8} & \fmt{34.6}{2.4} & \fmt{37.1}{3.7} & \textbf{\fmt{49.2}{10.1}} & \textbf{\fmt{52.9}{9.5}} \\
\electricitynips & \fmt{25.0}{6.6} & \fmt{17.5}{3.5} & \fmt{37.5}{6.6} & \textbf{\fmt{77.5}{8.6}} & \fmt{64.2}{8.6} \\
\exchangerate & \fmt{28.7}{3.0} & \fmt{25.3}{5.1} & \fmt{26.7}{7.1} & \textbf{\fmt{36.7}{5.8}} & \fmt{26.7}{4.7} \\
\textbf{Average} & \fmt{32.6}{} & \fmt{28.1}{} & \fmt{34.8}{} & \textbf{\fmt{57.1}{}} & \fmt{48.7}{} \\
\midrule
\textbf{\antl}   & &  & & & \\
\midrule
\aed & \textbf{\fmt{25.0}{7.9}} & \fmt{21.0}{3.5} & \fmt{21.7}{7.9} & \fmt{21.7}{8.3} & \textbf{\fmt{34.3}{5.7}} \\
\electricity & \fmt{25.0}{5.1} & \textbf{\fmt{30.0}{7.5}} & \fmt{20.0}{6.8} & \textbf{\fmt{47.5}{15.2}} & \textbf{\fmt{46.7}{11.9}} \\
\electricityhourly & \fmt{25.8}{4.6} & \fmt{21.7}{2.4} & \fmt{23.8}{6.4} & \textbf{\fmt{40.0}{7.6}} & \textbf{\fmt{33.8}{6.8}} \\
\electricitynips & \fmt{14.2}{7.0} & \fmt{15.8}{5.4} & \fmt{21.7}{7.5} & \textbf{\fmt{35.0}{14.3}} & \textbf{\fmt{46.7}{12.6}} \\
\exchangerate & \textbf{\fmt{21.3}{8.4}} & \textbf{\fmt{22.0}{7.7}} & \textbf{\fmt{20.7}{4.9}} & \fmt{8.0}{3.0} & \textbf{\fmt{11.3}{7.3}} \\
\textbf{Average} & \fmt{22.3}{} & \fmt{22.1}{} & \fmt{21.6}{} & \fmt{30.4}{} & \textbf{\fmt{34.6}{}} \\
\bottomrule
\end{tabular}
\end{adjustbox}
\end{table*}

%% file: tables/avgrew_intraepisode.tex
\begin{table*}[t!]
\caption{\textbf{Intra-episode non-stationarity results with $f=50$.} We compare methods with access to past offsets (\difqlmeanlag vs. \forl) and without (\difql vs. \forl-\dm).}
\label{tab:avgrew_intraepisode}
\centering
\begin{adjustbox}{max width=\textwidth}
\begin{tabular}{lcccc}
\toprule
\textbf{\mazem} & \difql & \forl-\dm & \difqlmeanlag & \textcolor{ourdarkred}{\forl} \\
\midrule
\aed& \fmt{31.6}{1.6} & \textbf{\fmt{55.7}{8.7}} & \fmt{29.7}{9.2} & \fmt{17.8}{4.4} \\
\electricity & \fmt{-4.7}{0.2} & \fmt{42.2}{9.3} & \fmt{63.0}{9.2} & \textbf{\fmt{100.1}{8.5}} \\
\electricityhourly & \fmt{-4.7}{0.2} & \fmt{39.2}{7.2} & \textbf{\fmt{88.4}{9.4}} & \textbf{\fmt{81.0}{8.3}} \\
\electricitynips & \fmt{25.9}{5.3} & \fmt{37.0}{8.4} & \fmt{43.1}{8.1} & \textbf{\fmt{101.0}{7.1}} \\
\exchangerate & \fmt{-4.7}{0.2} & \textbf{\fmt{57.0}{10.2}} & \fmt{5.8}{2.2} & \textbf{\fmt{66.7}{8.3}} \\
\textbf{Average} & \fmt{8.7}{} & \fmt{46.2}{} & \fmt{46.0}{} & \textbf{\fmt{73.3}{}} \\
\midrule
\multicolumn{5}{l}{\textbf{\antu}} \\
\midrule
\aed& \fmt{51.0}{7.4} & \fmt{47.2}{10.6} & \textbf{\fmt{70.6}{14.5}} & \textbf{\fmt{78.2}{8.0}} \\
\electricity & \textbf{\fmt{63.8}{10.2}} & \textbf{\fmt{51.2}{10.3}} & \fmt{26.0}{16.8} & \textbf{\fmt{61.4}{9.7}} \\
\electricityhourly & \fmt{61.0}{9.3} & \fmt{53.0}{6.6} & \textbf{\fmt{91.2}{2.5}} & \textbf{\fmt{89.0}{5.1}} \\
\electricitynips & \fmt{17.6}{5.9} & \fmt{53.8}{5.7} & \fmt{78.0}{4.4} & \textbf{\fmt{90.0}{2.9}} \\
\exchangerate & \fmt{0.2}{0.4} & \fmt{51.0}{11.6} & \textbf{\fmt{80.6}{9.7}} & \textbf{\fmt{85.6}{5.5}} \\
\textbf{Average} & \fmt{38.7}{} & \fmt{51.2}{} & \fmt{69.3}{} & \textbf{\fmt{80.8}{}} \\
\midrule
\multicolumn{5}{l}{\textbf{\antm}} \\
\midrule
\aed& \fmt{42.8}{6.9} & \fmt{7.2}{1.8} & \textbf{\fmt{50.4}{2.9}} & \textbf{\fmt{54.8}{5.2}} \\
\electricity & \fmt{5.0}{3.1} & \fmt{37.4}{3.8} & \textbf{\fmt{41.2}{4.8}} & \textbf{\fmt{47.2}{4.6}} \\
\electricityhourly & \fmt{12.0}{5.1} & \fmt{26.8}{2.4} & \textbf{\fmt{71.2}{2.9}} & \fmt{61.4}{4.5} \\
\electricitynips & \fmt{24.0}{4.7} & \fmt{30.4}{2.9} & \textbf{\fmt{61.6}{5.0}} & \textbf{\fmt{70.2}{6.6}} \\
\exchangerate & \fmt{2.2}{2.3} & \fmt{21.0}{3.3} & \textbf{\fmt{33.0}{5.0}} & \textbf{\fmt{37.0}{3.4}} \\
\textbf{Average} & \fmt{17.2}{} & \fmt{24.6}{} & \fmt{51.5}{} & \textbf{\fmt{54.1}{}} \\
\midrule
\multicolumn{5}{l}{\textbf{\antl}} \\
\midrule
\aed& \fmt{13.0}{5.5} & \fmt{5.8}{2.6} & \fmt{16.0}{3.8} & \textbf{\fmt{27.4}{3.6}} \\
\electricity & \fmt{1.2}{0.8} & \fmt{11.4}{4.9} & \fmt{19.4}{4.6} & \textbf{\fmt{50.2}{9.7}} \\
\electricityhourly & \fmt{4.8}{2.9} & \fmt{7.2}{0.8} & \fmt{31.8}{4.4} & \textbf{\fmt{48.6}{4.3}} \\
\electricitynips & \fmt{4.6}{1.3} & \fmt{11.0}{2.5} & \fmt{25.2}{2.9} & \textbf{\fmt{48.2}{6.2}} \\
\exchangerate & \fmt{2.4}{1.5} & \textbf{\fmt{13.6}{4.8}} & \textbf{\fmt{12.8}{3.4}} & \textbf{\fmt{13.4}{2.2}} \\
\textbf{Average} & \fmt{5.2}{} & \fmt{9.8}{} & \fmt{21.0}{} & \textbf{\fmt{37.6}{}} \\
\bottomrule
\end{tabular}
\end{adjustbox}
\end{table*}

%% file: tables/error_reductiontable.tex
\begin{table*}[t]
\centering
\caption{\textbf{Sensitivity analysis of \forl to forecasting errors.} We compare the average prediction error ($\downarrow$) of our method against the \difqlmeanlag baseline, which uses only the forecaster's predictions. The analysis is presented across five time-series datasets. Error reduction percentages are calculated from full-precision values before rounding.}
\vskip 0.15in
\setlength{\tabcolsep}{3pt}
\small
\begin{tabular}{l c c c}
\toprule
Dataset & \difqlmeanlag($\downarrow$) &\textcolor{ourdarkred}{\forl($\downarrow$)} & \textit{Error Reduction}($\uparrow$) \\
\midrule
\aed & 4.56 & \textbf{3.32} & 27.0\% \\
\electricity & 3.66 & \textbf{2.29} & 37.4\% \\
\electricityhourly & 3.0 & \textbf{2.69} & 10.2\% \\
\electricitynips & 4.29 & \textbf{1.87} & \textbf{56.5\%} \\
\exchangerate & 5.45 & \textbf{5.21} &  4.3\% \\
\bottomrule
\end{tabular}
\label{tab:sensitivity-error-reduction}
\end{table*}

%% file: tables/allcompareepredacctable.tex
\begin{table*}[htbp]
\centering
\caption{\textbf{Comparison of prediction errors ($\downarrow$).} We present state prediction accuracy for the proposed \forl framework with the baselines across 5 random seeds.}
\begin{adjustbox}{max width=\textwidth}
\small
\begin{tabular}{lcc}
\toprule $\textbf{\mazem}$ &\difqlmeanlag & $\textbf{\forl}$\\
\midrule \difqlmeanlag & $1.68\pm0.46$ & $\bf 1.35\pm0.48$ \\
 $\textbf{\forl}$ & $1.39\pm0.49$  & $\bf 1.25\pm0.43$ \\
\midrule  $\textbf{\mazel}$ &\difqlmeanlag & $\textbf{\forl}$\\
\midrule\difqlmeanlag & $2.37\pm0.5$ & $\bf 1.63\pm0.57$ \\
 $\textbf{\forl}$ & $1.91\pm0.44$ & $\bf 1.39\pm0.62$ \\
\midrule $\textbf{\antl}$&\difqlmeanlag & $\textbf{\forl}$\\
\midrule \difqlmeanlag & $8.07\pm1.83$ & $\bf 5.33\pm2.17$ \\
 $\textbf{\forl}$ & $7.16\pm1.71$ & $\bf 5.89\pm2.96$ \\
\midrule $\textbf{\antm}$ &\difqlmeanlag & $\textbf{\forl}$\\
\midrule\difqlmeanlag & $5.33\pm1.15$ & $\bf 4.75\pm1.98$ \\
$\textbf{\forl}$ & $4.94\pm1.17$ & $\bf 4.74\pm1.63$ \\
\midrule $\textbf{\antu}$ & \difqlmeanlag & $\textbf{\forl}$\\
\midrule\difqlmeanlag & $3.51\pm0.53$ & $\bf 2.04\pm0.47$ \\
 $\textbf{\forl}$ & $3.27\pm0.61$ & $\bf 2.31\pm0.51$ \\
\bottomrule
\end{tabular}
\end{adjustbox}
\label{app:tab:allcompareepredacctable}
\end{table*}

%% file: tables/tsdatasetavgrew.tex
\begin{table*}[t!]
\caption{\textbf{Normalized scores (mean ± std.) for \forl framework and the baselines grouped by time-series.} Bold are the best values, and those not significantly different ($p > 0.05$, Welch's t-test).}
\label{tab:timeseriescategorized_avgrewtable}
\centering
\begin{adjustbox}{max width=\textwidth}
\begin{tabular}{@{}l@{\hspace{+1ex}}rccc@{}}
\toprule
\textbf{\aed}    & \multicolumn{1}{c}{\difql}& \difqlmeanlag   & \dmbplag        & \textcolor{ourdarkred}{\forl(ours)}        \\
\midrule
\textbf{\mazem} & \fmt{30.2}{6.5}         & \fmt{30.2}{8.6}               & \fmt{25.1}{9.8}           & \textbf{\fmt{63.3}{6.7}}    \\
\textbf{\mazel} & \fmt{16.2}{5.5}         & \fmt{2.4}{1.1}                & \fmt{4.2}{5.8}            & \textbf{\fmt{42.9}{4.1}}    \\
\textbf{\antu}   & \fmt{22.7}{3.0}         & \fmt{41.0}{5.2}               & \fmt{45.7}{4.8}           & \textbf{\fmt{65.3}{8.7}}             \\
\textbf{\antm}   & \fmt{31.0}{6.5}         & \textbf{\fmt{40.0}{5.7}}    & \textbf{\fmt{39.7}{4.0}}    & \textbf{\fmt{44.0}{7.9}}             \\
\textbf{\antl} & \fmt{11.0}{1.9}         & \fmt{11.3}{4.9}               & \fmt{9.0}{4.5}            & \textbf{\fmt{34.3}{5.7}}    \\
\textbf{\kitchc}   & \textbf{\fmt{16.6}{1.4}} & \fmt{7.2}{1.9} & \fmt{8.7}{1.3} & \textbf{\fmt{12.0}{3.9}} \\
\midrule
\textbf{\electricity}   \\ 
\midrule
\textbf{\mazem}  & \fmt{14.1}{12.1}        & \textbf{\fmt{53.4}{14.6}}     & \textbf{\fmt{41.2}{21.1}} & \textbf{\fmt{66.5}{18.2}}            \\
\textbf{\mazel}  & \fmt{-0.5}{2.9}         & \fmt{5.5}{9.0}                & \fmt{15.0}{14.6}          & \textbf{\fmt{34.9}{9.2}}    \\
\textbf{\antu}   & \fmt{24.2}{3.5}         & \fmt{48.3}{7.0}               &  \textbf{\fmt{62.5}{13.2}}          & \textbf{\fmt{74.2}{10.8}}   \\
\textbf{\antm}   & \fmt{23.3}{4.8}         & \textbf{\fmt{48.3}{4.8}}      & \textbf{\fmt{43.3}{16.0}} & \textbf{\fmt{55.8}{7.0}}    \\
\textbf{\antl}  & \fmt{5.8}{4.8}          & \fmt{9.2}{4.6}                & \fmt{8.3}{2.9}            & \textbf{\fmt{46.7}{11.9}}   \\
\textbf{\kitchc}   & \fmt{12.9}{4.1} & \textbf{\fmt{32.7}{6.5}} & \fmt{20.0}{3.1} & \textbf{\fmt{33.1}{5.6}} \\
\midrule
\textbf{\electricityhourly}  \\ 
\midrule
\textbf{\mazem} & \fmt{-2.3}{3.3}         & \fmt{56.7}{18.5}              & \fmt{56.9}{18.4}          & \textbf{\fmt{86.3}{15.7}}   \\
\textbf{\mazel}  & \fmt{0.9}{1.7}          & \fmt{16.6}{7.5}               & \fmt{26.8}{8.4}           & \textbf{\fmt{45.6}{4.1}}    \\
\textbf{\antu}   & \fmt{21.7}{3.5}         & \fmt{50.4}{8.3}               & \fmt{60.4}{3.9}           & \textbf{\fmt{78.8}{8.5}}    \\
\textbf{\antm}   & \fmt{10.0}{2.3}         & \textbf{\fmt{48.3}{3.4}}      & \textbf{\fmt{49.6}{3.7}}  & \textbf{\fmt{52.9}{9.5}}    \\
\textbf{\antl}  & \fmt{5.4}{2.4}          & \fmt{22.1}{5.6}               & \fmt{17.9}{3.8}           & \textbf{\fmt{33.8}{6.8}}    \\
\textbf{\kitchc}   & \fmt{13.4}{1.7} & \textbf{\fmt{23.9}{6.6}} & \textbf{\fmt{20.5}{3.3}} & \textbf{\fmt{23.9}{6.0}} \\
\midrule
\textbf{\electricitynips}   \\ 
\midrule
\textbf{\mazem}  & \fmt{4.7}{5.0}          & \fmt{36.9}{16.3}              & \fmt{38.5}{14.2}          & \textbf{\fmt{103.4}{11.9}}  \\
\textbf{\mazel}  & \fmt{3.0}{6.6}          & \fmt{8.6}{3.2}                & \fmt{13.4}{4.1}           & \textbf{\fmt{58.4}{6.5}}    \\
\textbf{\antu}   & \fmt{5.8}{2.3}          & \fmt{26.7}{6.3}               & \fmt{29.2}{5.9}           & \textbf{\fmt{75.8}{8.0}}    \\
\textbf{\antm}   & \fmt{11.7}{5.4}         & \fmt{46.7}{7.5}               & \fmt{41.7}{6.6}           & \textbf{\fmt{64.2}{8.6}}             \\
\textbf{\antl}  & \fmt{2.5}{2.3}          & \fmt{14.2}{3.7}               & \fmt{14.2}{6.3}           & \textbf{\fmt{46.7}{12.6}}   \\
\textbf{\kitchc}   & \fmt{7.5}{2.5} & \textbf{\fmt{24.0}{9.2}} & \textbf{\fmt{28.1}{8.1}} & \textbf{\fmt{27.1}{10.1}} \\
\midrule
\textbf{\exchangerate} \\ 
\midrule
\textbf{\mazem}  & \fmt{3.5}{8.8}          & \fmt{8.7}{6.0}                & \fmt{11.4}{2.8}           & \textbf{\fmt{51.2}{13.7}}   \\
\textbf{\mazel}  & \fmt{-2.1}{0.4}         & \textbf{\fmt{2.6}{3.4}}       & \textbf{\fmt{0.9}{3.7}}   & \textbf{\fmt{12.0}{9.9}}    \\
\textbf{\antu}   & \fmt{6.0}{6.8}          & \fmt{58.0}{16.6}              & \fmt{59.3}{7.6}           & \textbf{\fmt{81.3}{6.9}}    \\
\textbf{\antm}    & \textbf{\fmt{18.7}{4.5}} & \textbf{\fmt{27.3}{8.6}}      & \textbf{\fmt{26.0}{5.5}}  & \textbf{\fmt{26.7}{4.7}}             \\
\textbf{\antl}   & \textbf{\fmt{5.3}{3.8}} &  \textbf{\fmt{3.3}{2.4}}      &  \textbf{\fmt{3.3}{0.0}}   & \textbf{\fmt{11.3}{7.3}}    \\
\textbf{\kitchc}  & \textbf{\fmt{18.5}{6.0}} & \fmt{2.8}{2.1} & \fmt{6.2}{1.7} & \fmt{10.3}{3.0} \\
\bottomrule
\end{tabular}
\end{adjustbox}
\end{table*}

%% file: tables/avgrewtable_affine.tex
\begin{table*}[t!]
\caption{\textbf{Performance under affine observation shifts.} Normalized scores in \mazel with time-varying uniform scaling and bias.}
\label{tab:avgrewtable_affine}
\centering
\begin{tabular}{@{}l@{\hspace{10ex}}rcc@{}}
\toprule
\textbf{\mazel}    & \multicolumn{1}{c}{\difql}& \difqlmeanlag   & \textcolor{ourdarkred}{\forl(ours)}        \\
\midrule
\aed & 5.9 & 6.1 & \textbf{39.7} \\
\electricity       & 2.4 & 1.6 & \textbf{13.8} \\
\electricityhourly     & 2.2 & 22.2 & \textbf{32.9} \\
\electricitynips    & 0.7 & 10.7 & \textbf{56.1} \\
\exchangerate     & -2.0 & -2.3 & \textbf{27.5} \\
\textbf{Average}  & 1.8 & 7.7 & \textbf{34.0} \\
\bottomrule
\end{tabular}
\end{table*}

%% file: tables/policy_hyperparams.tex
{\scriptsize
\begin{longtable}{l l l l}
  \caption[Hyperparameters for \difql across AntMaze/Maze/Kitchen]{%
    Hyperparameters for \difql \cite{wang2023diffusion,diffpol_repo,zhyang2024dmbpcode}
    across \kitchc, \mazed, and \antmaze environments.%
  }\label{table:diffql_full}\\
\toprule
\textbf{Hyperparameters} &  &  &  \\
\midrule
\endfirsthead
Maximum Timesteps             & 1\,000\,000         &          &         \\
$\gamma$                   & 0.99                &          &         \\
$\tau$                     & 0.005               &          &         \\
Learning rate decay        & true                &          &         \\
T                        & 10                  &          &         \\
$\beta$ Schedule           & \texttt{vp}         &          &         \\
Learning rate                       & $3\times10^{-4}$    &          &         \\
$\alpha$                 & 0.2                 &          &         \\
Batch Size             & 256                 &          &         \\
Hidden Size              & 256                 &          &         \\
Reward tune             & \texttt{no}         &          &         \\
Normalize                & false               &          &         \\
Optimizer                & Adam\cite{kingma2015adam}               &          &         \\
\midrule
                          & \textbf{\kitchc}    & \textbf{\mazed}    & \textbf{\antmaze} \\
gn                        & 10.0                & \parbox[t]{3.5cm}{\texttt{umaze‐diverse: 3.0}\\\texttt{medium‐diverse: 1.0}\\\texttt{large‐diverse: 7.0}} 
                                             & \parbox[t]{3.5cm}{\texttt{umaze‐diverse: 3.0}\\\texttt{medium‐diverse: 1.0}\\\texttt{large‐diverse: 7.0}} \\
$\eta$                       & 0.005               & \parbox[t]{3.5cm}{\texttt{umaze‐diverse: 2.0}\\\texttt{medium‐diverse: 3.0}\\\texttt{large‐diverse: 3.5}} 
                                             & \parbox[t]{3.5cm}{\texttt{umaze‐diverse: 2.0}\\\texttt{medium‐diverse: 3.0}\\\texttt{large‐diverse: 3.5}} \\
MaxQ Backup           & false               & true                & true    \\
\bottomrule
\end{longtable}
}

{\scriptsize
\begin{longtable}{l l l l}
  \caption[Hyperparameters for \iql]{%
    Hyperparameters for Implicit Q-Learning (\iql) \cite{iql_Kostrikov_2021,ikostrikov_iql_official,gwthomas_iql_pytorch,tarasov2022corl} across \mazed, and \antmaze environments.
  }\label{table:iql_full}\\
\toprule
\textbf{Hyperparameters} & \textbf{Value} \\
\midrule
\endfirsthead
Batch Size                & 256 \\
Discount ($\gamma$)       & 0.99 \\
Target Network Update ($\tau$) & 0.005 \\
\midrule
\mazed & $\beta = 3.0$ \\
         & $\tau_{\text{IQL}} = 0.7$ \\
         & Normalize Rewards = false \\
\antmaze & $\beta = 10.0$ \\
         & $\tau_{\text{IQL}} = 0.9$ \\
         & Normalize Rewards = true \\
\bottomrule
\end{longtable}
}
\newpage
{\scriptsize
\begin{longtable}{l l}
  \caption[Hyperparameters for \fql ]{%
    Hyperparameters for Flow Q-Learning (\fql) \cite{fql_park2025,fql_repo_official} for \ogcube and \ogantl.
  }\label{table:fql_full}\\
\toprule
\textbf{Hyperparameters} & \textbf{Value} \\
\midrule
\endfirsthead
Batch Size                & 256 \\
Learning Rate             & $0.0003$ \\
Discount factor ($\gamma$)       & 0.99 \\
Target network smoothing coefficient ($\tau$) & 0.005 \\
BC Coefficient ($\alpha$) & 10.0  \\
Flow Steps                & 10 \\
Actor Hidden Dimensions   & (512, 512, 512, 512) \\
Value Hidden Dimensions   & (512, 512, 512, 512) \\
\midrule
\ogantl & BC Coefficient ($\alpha$) = 10.0 \\
        
\midrule
\ogcube & BC Coefficient ($\alpha$) = 300.0 \\
\bottomrule
\end{longtable}
}

{\scriptsize
\begin{longtable}{l l}
  \caption[Hyperparameters for \tdthreebc across AntMaze/Maze/Kitchen]{%
    Hyperparameters for \tdthreebc \cite{fujimoto2018td3,td3bc_repo,zhyang2024dmbpcode} across \kitchc, \mazed, and \antmaze environments.%
  }\label{table:td3bc_full}\\
\toprule
\textbf{Hyperparameters} & \textbf{Value} \\
\midrule
\endfirsthead
\endfoot
\endlastfoot
Maximum Timesteps    & 1\,000\,000 \\
Exploration noise       & 0.1 \\
Batch Size       & 256 \\
Discount factor          & 0.99 \\
$\tau$               & 0.005 \\
Policy Noise     & 0.2 \\
Policy Noise Clipping       & 0.5 \\
Policy update frequency     & 2 \\
$\alpha$             & 2.5 \\
Normalize         & true \\
Optimizer & Adam\cite{kingma2015adam} \\
\bottomrule
\end{longtable}
}

{\scriptsize
\begin{longtable}{l l l l}
  \caption[Hyperparameters for \rorl across \mazed]{%
    Hyperparameters for \rorl
    \cite{yang2022rorl, yang2022_rorl_repo, zhyang2024dmbpcode} in \mazed environments.%
  }\label{table:rorl_full}\\
\toprule
\textbf{Hyperparameters}           &            &           \\
\midrule
\endfirsthead
$\gamma$                        & 0.99          &            &           \\
$\textbf{soft}_{\tau}$       & 0.005         &            &           \\
Q Learning Rate                       & $3\times10^{-4}$ &         &           \\
Policy Learning Rate                   & $3\times10^{-4}$ &         &           \\
$\alpha$                        & 1.0           &            &           \\
Auto-tune entropy         & true          &            &           \\
MaxQ Backup              & false         &            &           \\
Deterministic Backup        & false         &            &           \\
$\eta$                          & $-1$          &            &           \\
Batch Size                  & 256           &            &           \\
Hidden Size                & 256           &            &           \\
Target Update Interval     & 1             &            &           \\
$\tau$                          & 0.2           &            &           \\
Normalize                    & false         &            &           \\
n sample                      & 20                 &            &           \\
$\beta_Q$                        & 0.0001            &            &           \\
$\beta_P$                        & 1.0                &            &           \\
$\eps_{ood}$                       & 0.01              &            &           \\
Maximum Timesteps                & 3\,000\,000   &            &           \\
Optimizer                & Adam\cite{kingma2015adam}               &          &         \\
\midrule       
  & \textbf{\mazed}    \\
$\beta_{OOD}$                                & 0.5                    \\
$\eps_Q$                                & 0.01                 \\
$\eps_P$                                  & 0.03                  \\
$\lambda_{max}$                             & 1.0                  \\
$\lambda_{min}$                            & 0.5                     \\
$\lambda_{decay}$                    & $10^{-6}$              \\
\bottomrule
\end{longtable}
}

{\scriptsize
\setlength{\tabcolsep}{1pt} 
\begin{longtable}{@{}lccccc@{}} 
  \caption[Hyperparameters for \forl across AntMaze/Maze/Kitchen/Cube/LargeNavigate]{%
    Hyperparameters for \forl across \kitchc, \mazed, \antmaze, \ogantl, \ogcube environments.%
  }\label{tab:combined_hyperparameters}\\
\toprule
\textbf{Hyperparameters} &  &  & &  &  \\
\midrule
\endfirsthead
Batch Size             & 128                 &          &       &  &  \\
Hidden Size              & 128                 &          &      &  &  \\
\# denoiser samples & 50  &          &        &  &  \\
Optimizer                & Adam\cite{kingma2015adam}               &          &   &  &  \\
Maximum Timesteps             & 300\,000         &          &        &  &  \\
\midrule
& \textbf{\kitchc}   & \textbf{\antmaze}  & \textbf{\mazed} & \textbf{\ogantl}  & \textbf{\ogcube}    \\
Embedding Dimension        & 128  & 64 & 64 &      64    &    128     \\
$w$ & 32 & 256 & 128   &      256    &  64       \\
Learning rate &  $4\times10^{-4}$& $4\times10^{-4}$  &  $9\times10^{-4}$   &  $4\times10^{-4}$        &   $9\times10^{-4}$       \\
Observation Scale  & 1  & 1 & 100 &      1    &     1    \\
Time Concatenation & true  & true & false &  true        &   true      \\
\# middle hidden layers & 1 & 1 &  \parbox[t]{1.5cm}{\texttt{large: 1}\\\texttt{medium: 3}} &    1      &   1      \\
N & 10 & 10 &  \parbox[t]{1.5cm}{\texttt{large: 10}\\\texttt{medium: 20}} &   20       &     20    \\
\bottomrule
\end{longtable}
}